\pdfoutput=1
\documentclass{applemlr}

\usepackage{amsmath}
\usepackage{enumerate}
\usepackage{algorithm}
\usepackage{algpseudocode}
\usepackage{amsfonts}
\usepackage{amsthm}
\usepackage{cleveref}
\usepackage{diagbox}
\usepackage{colortbl}
\usepackage{amssymb}
\usepackage{xspace}
\usepackage{wrapfig}
\usepackage{adjustbox}
\usepackage{tabularx}
\usepackage{booktabs}
\usepackage{mathtools}
\usepackage{tikz}
\usepackage{enumitem}
\usepackage{silence}
\usepackage{dsfont}
\usepackage[table]{xcolor}
\usepackage[dvipsnames]{xcolor}
\usepackage{multirow}
\usepackage{makecell}
\usepackage{xfakebold}

\usepackage{amsmath,amsfonts,bm}

\def\eqref#1{equation~\ref{#1}}

\def\1{\bm{1}}

\def\allw{\mathbf{\Rvw}}
\def\allb{\mathbf{\Rvb}}

\def\vb{{\bm{b}}}

\def\vo{{\bm{o}}}

\def\vw{{\bm{w}}}
\def\vx{{\bm{x}}}
\def\vy{{\bm{y}}}

\def\vxi{\bm{\xi}}
\def\veta{\bm{\eta}}

\DeclareMathAlphabet{\mathsfit}{\encodingdefault}{\sfdefault}{m}{sl}
\SetMathAlphabet{\mathsfit}{bold}{\encodingdefault}{\sfdefault}{bx}{n}

\def\sR{{\mathbb{R}}}

\DeclareMathOperator*{\sort}{sort}
\DeclareMathOperator*{\E}{\mathbb{E}}

\DeclareMathOperator{\sign}{sign}

\DeclareMathOperator{\ST}{ST}
\DeclareMathOperator{\Prox}{Prox}

\definecolor{textgray}{HTML}{6E6E73}
\usetikzlibrary{positioning, calc}
\usetikzlibrary{decorations.pathmorphing}

\makeatletter
\patchcmd{\wrong@fontshape}{\@gobbletwo}{}{}{}
\makeatother
\numberwithin{equation}{section}
\makeatletter
\AtBeginDocument{
  \urlstyle{sf}
  
}
\makeatother

\definecolor{light}{RGB}{125, 125, 125}
\crefname{tcb@cnt@pbox}{code}{code}
\Crefname{tcb@cnt@pbox}{Code}{Code}
\crefname{assumption}{assumption}{assumption}
\Crefname{assumption}{Assumption}{Assumptions}

\newtcolorbox[auto counter]{pbox}[2][]{
  colback=white,
  title=Code~\thetcbcounter: #2,
  #1,fonttitle=\sffamily,
  fontupper=\sffamily,
  arc=2pt,
  colframe=bgcolor,
  coltitle=fgcolor,
  colbacktitle=bgcolor,
  toptitle=0.25cm,
  bottomtitle=0.125cm
}

\makeatletter
\newcommand\applefootnote[1]{%
  \begingroup
  \renewcommand\thefootnote{}%
  \renewcommand\@makefntext[1]{\noindent##1}%
  \footnote{#1}%
  \addtocounter{footnote}{-1}%
  \endgroup
}
\makeatother

\definecolor{cverbbg}{gray}{0.90}

\usepackage[utf8]{inputenc}
\usepackage[T1]{fontenc}
\usepackage{url}
\usepackage{nicefrac}
\usepackage{microtype}
\usepackage{nicematrix}
\usepackage{float}
\usepackage{needspace}
\usepackage{arydshln}
\usepackage{xparse}
\usepackage[disable,textsize=tiny]{todonotes}
\usepackage[textsize=tiny]{todonotes}

\PassOptionsToPackage{numbers, compress}{natbib}

\newcommand{\blue}[1]{\textcolor{blue}{#1}}
\newcommand{\red}[1]{\textcolor{red}{#1}}

\newcommand{\RT}[2][]{\red{T_{#2}^{#1}}}
\newcommand{\Bf}[2][]{\blue{f_{#2}^{#1}}}
\def\Rvw{\red{\vw}}
\def\Rvb{\red{\vb}}

\newcommand{\rebuttal}[1]{#1}

\newcommand{\reg}{\gamma}

\newcommand{\ROmega}[2][]{\red{\omega_{#2}^{#1}}}
\newcommand{\Rb}[2][]{\red{b_{#2}^{#1}}}

\newcommand{\RTC}{{RTC}\xspace}

\newcommand{\toxclsrtp}{Tox$_{\text{\RTC}}^{\text{RTP}}$\xspace}
\newcommand{\toxclstet}{Tox$_{\text{\RTC}}^{\text{TET}}$\xspace}

\newcommand{\pplwik}{PPL$_{\text{WIK}}$\xspace}
\newcommand{\imgscore}{IMGScore\xspace}
\newcommand{\clipscore}{CLIPScore\xspace}

\newcommand{\method}{LinEAS\xspace}
\newcommand{\prompt}{{Prompt}\xspace}
\newcommand{\caa}{{CAA}\xspace}
\newcommand{\reft}{{ReFT}\xspace}
\newcommand{\linearact}{Lin-\textsc{AcT}\xspace}
\newcommand{\lorartc}{L\RTC-RL\xspace}
\newcommand{\lofitrl}{LoFIT-RL\xspace}

\newcommand{\qwensevenbinstr}{Qwen2.5-7B-Instruct\xspace}
\newcommand{\qwensevenb}{Qwen2.5-7B\xspace}
\newcommand{\qwenonefiveb}{Qwen2.5-1.5B\xspace}
\newcommand{\gemmatwob}{Gemma2-2B\xspace}
\newcommand{\gemmatwosevenb}{Gemma2-27B\xspace}
\newcommand{\deepseeksevenb}{DeepSeek-7B\xspace}

\theoremstyle{plain}

\theoremstyle{definition}

\theoremstyle{remark}

\usepackage{hyperref}
\usepackage{url}
\usepackage{xspace}
\usepackage{cleveref}
\usepackage{graphicx}
\usepackage{stmaryrd}
\usepackage{subcaption}
\usepackage{booktabs}

\usepackage{algorithm}
\usepackage{algpseudocode}

\usepackage{tcolorbox}%
\usepackage{listings}
	
\usepackage{color, colortbl}
\usepackage{placeins}
\usepackage{hyperref}
\usepackage{multirow} %
\usepackage{nicematrix}
\usepackage{dsfont}
\usepackage{rotating}
\usepackage{tikz}
\usetikzlibrary{decorations.pathreplacing,positioning,calc,shapes.geometric,arrows,bending}
\usetikzlibrary[bending]

\definecolor{TableRow}{rgb}{0.9, 0.9, 0.92}

\DeclareFontEncoding{LS1}{}{}
\DeclareFontSubstitution{LS1}{stix}{m}{n}
\DeclareSymbolFont{stixletters}{LS1}{stix}{m}{it}
\DeclareMathAccent{\cev}{\mathord}{stixletters}{"91}
\DeclareMathAccent{\vec}{\mathord}{stixletters}{"92}
\DeclareMathAccent{\vecev}{\mathord}{stixletters}{"95}

\newcommand{\aura}{\textsc{AurA}\xspace}
\newcommand{\iti}{\textsc{ITI-c}\xspace}

\newcommand{\actadd}{\textsc{ActAdd}\xspace}
\newcommand{\repe}{\textsc{RepE}\xspace}

\newcommand{\eg}{\textit{e.g.,}\xspace}
\newcommand{\ie}{\textit{i.e.,}\xspace}

\definecolor{grayorange}{HTML}{967653}
\definecolor{takeawaycolor}{rgb}{0.1, 0.1, 0.12}
\definecolor{improvementcolor}{rgb}{0.122, 0.8, 0.22}
     
\newtcbox{\mybox}{
  colframe=takeawaycolor,colback=takeawaycolor!5!white,fontupper=\rmfamily,
  boxrule=0.0pt,arc=0pt,boxsep=0pt,left=2pt,right=2pt,top=1.9pt,bottom=1pt,nobeforeafter,tcbox raise base,after={\hspace{2pt}}}

\newtcbox{\myimprovbox}{
  colframe=improvementcolor,colback=improvementcolor!5!white,fontupper=\rmfamily,
  boxrule=0.0pt,arc=2pt,boxsep=0pt,left=2pt,right=2pt,top=1.9pt,bottom=1pt,nobeforeafter,tcbox raise base,after={\hspace{2pt}}}

\newcommand{\once}{{\textcolor{gray}{Once upon a time}}}

\newcommand{\vflip}[1]{\raisebox{1.5ex}{\scalebox{1}[-1]{#1}}}

\title{LinEAS: End-to-end Learning of Activation Steering with a Distributional Loss}

\newcommand{\talldiamond}{\raisebox{-0.1ex}{\scalebox{0.6}[1.1]{$\,\diamond$}}}

\author[*,1]{Pau Rodríguez}
\author[1]{Michal Klein}
\author[1]{Eleonora Gualdoni}
\author[1,2,\talldiamond]{Valentino Maiorca}
\author[1]{Arno Blaas}
\author[1]{Luca Zappella}
\author[1]{\\Marco Cuturi}
\author[*,1]{Xavier Suau}

\affiliation{\textsuperscript{1}Apple, \textsuperscript{2}Sapienza, \textsuperscript{\talldiamond}\textit{Work done as intern at Apple}}

\abstract{

The growing use of generative models in daily life calls for efficient mechanisms to control their generation, to \eg produce safe content or provide users with tools to explore style changes.
Ideally, such mechanisms should require low volume of unpaired data (\ie without explicit preference), and should be cheap, both at train and inference time, while preserving output quality.
Recent research has shown that such mechanisms can be obtained by intervening exclusively on model \textit{activations}, with the goal of correcting \textit{distributional} differences between activations seen when using prompts from a source vs. a target set (\eg toxic and non-toxic sentences).
While cheap, these fast methods are inherently crude: their maps are tuned locally, not accounting for their impact on downstream layers, resulting in interventions that cause unintended shifts when used out-of-sample.
We propose in this work linear end-to-end activation steering (\method), an approach trained with a global loss that accounts simultaneously for all layer-wise distributional shifts.
In addition to being more robust, the loss used to train \method can be regularized with sparsifying norms, which can automatically carry out neuron selection. 
\method only requires a handful of unpaired samples to be effective, and beats similar baselines on toxicity mitigation in language models, becoming competitive with oracle-dependent methods that have access to strong supervision. \method is modality-agnostic and we empirically find that it outperforms existing activation steering methods at mitigating and including new concepts at the output of single-step text-to-image generation models.

}

\metadata[Code]{\url{https://github.com/apple/ml-lineas}}
\metadata[Correspondence]{\sffamily pau.rodriguez@apple.com}
\date{\sffamily\today}

\begin{document}

\maketitle

\begin{figure}[ht!]
    \vskip-.3cm
    \centering
    \includegraphics[width=1.0\linewidth]{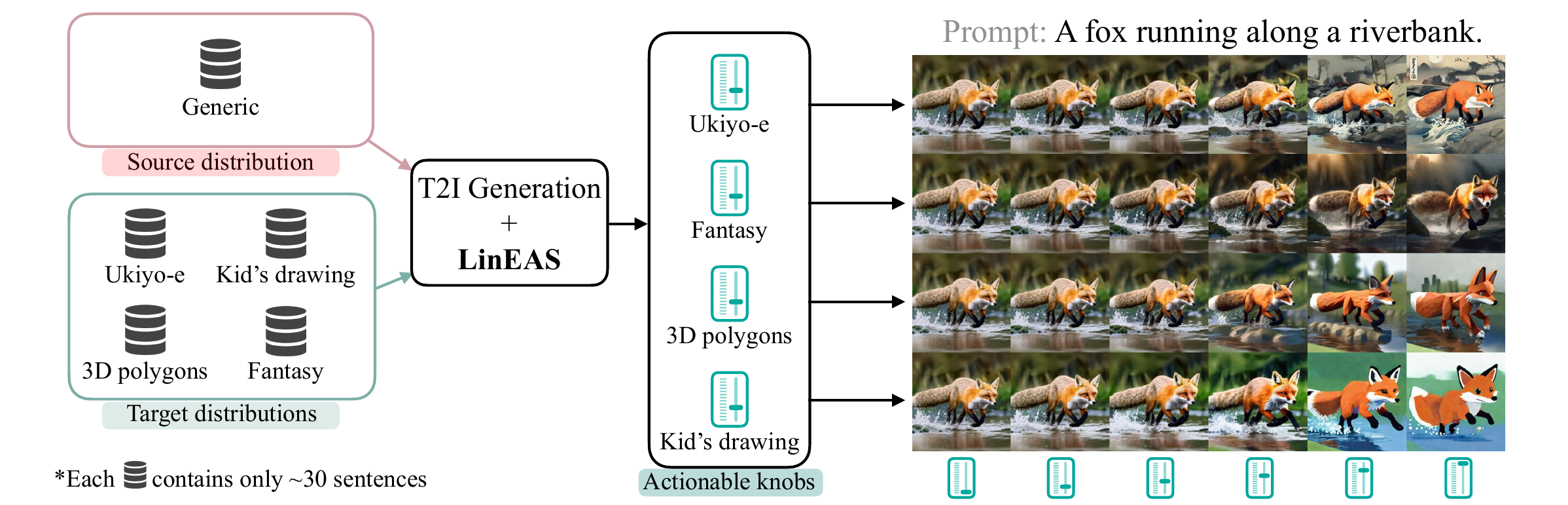}
    \caption{\method learns lightweight maps to steer pretrained model activations. With \method, we gain fine-grained control on text-to-image generation to induce  precise styles (in the figure) or remove objects (\eg \Cref{sec:results_diffusion}). The same procedure also allows controlling LLMs (\eg \Cref{sec:tox}). }
    \label{fig:fig_intro}
\end{figure}

\section{Introduction}  

Modern generative models are typically trained in two distinct phases. The first phase, known as pre-training, involves learning from a large corpus of data using tasks like next-token prediction or text-to-image generation. 
This is followed by an alignment phase designed to adjust the model towards a more specific, desired behavior. This alignment can be achieved through instruction fine-tuning~\citep{weifinetuned}, reinforcement learning from human feedback (RLHF)~\citep{ouyang2022training} for LLMs, or guidance~\citep{ho2022classifier} and LoRA adapters~\citep{luo2024stylus} in text-to-image diffusion.
Many of these approaches propose to modify the model's internal mechanisms, realigning its parameters by leveraging new data with, ideally, a minimal impact on the utility of the model.%

The rapid growth of model sizes, coupled with the potentially infinite combination of alignment goals, calls for alignment mechanisms that readily adapt to new and evolving user needs. Ideally, adaptable alignment methods should adhere to the following desiderata: low training cost (potentially on device), memory-efficiency (small set of parameters), low inference time overhead, data-efficiency (few annotated samples), and fine-grained control. Working in low-data regimes makes data collection simple or even unmanned, and makes training potentially faster while fine-grained control makes alignment methods more customizable~\citep{baumann2024continuous}. Such advantages together enable an agile control of generative models (see example in \Cref{fig:fig_intro}), giving tools to users to customize their experience, and to administrators to quickly intervene to prevent harmful model behaviors.

A body of work known as \textit{activation steering}~\citep{suau2022self,rimsky2023steering,zou2023representation,li2024inference,rodriguez2024controllinglanguagediffusionmodels,yin2024lofit} has proven to be effective at conditioning models while being memory and compute efficient, but still falls short in other desiderata.
Some methods require data points to be \textit{paired} with their corresponding counterfactuals~\citep{wu2024reft}, while others require a reward model (or a human) to indicate preference among generated outputs~\citep{ouyang2022training}. While this is a form of \textit{strong supervision} that provides a robust signal to alignment methods, it is costly or even impossible to obtain in many scenarios. %
A weaker and more flexible form of supervision consists of dealing with two sets of examples -- one for the target and one for the source behavior (\eg non-toxic and toxic sentences). In this setting, sentence pairs are not required, and no additional supervision is needed. Methods using this weaker signal typically analyze the distribution of activations from each set, thus we refer to them as a distributional approaches.

In this work, we propose Linear End-to-end Activation Steering (\method), a \textbf{low-data} and \textbf{weakly-supervised} (unpaired data, no reward model) method that is trained with a \textbf{global distributional cost} grounded in optimal transport theory as signal for steering. 
Our hypothesis, which we validate empirically, is that our end-to-end learning accounts for the interactions between maps at different layers, leading to improved results over other steering methods. An additional advantage of optimizing a global cost is that it allows us to introduce additional objectives such as a regularization coefficient, which results in more targeted interventions, preserving the utility of the model.

\begin{figure*}[t]
    \includegraphics[width=\textwidth]{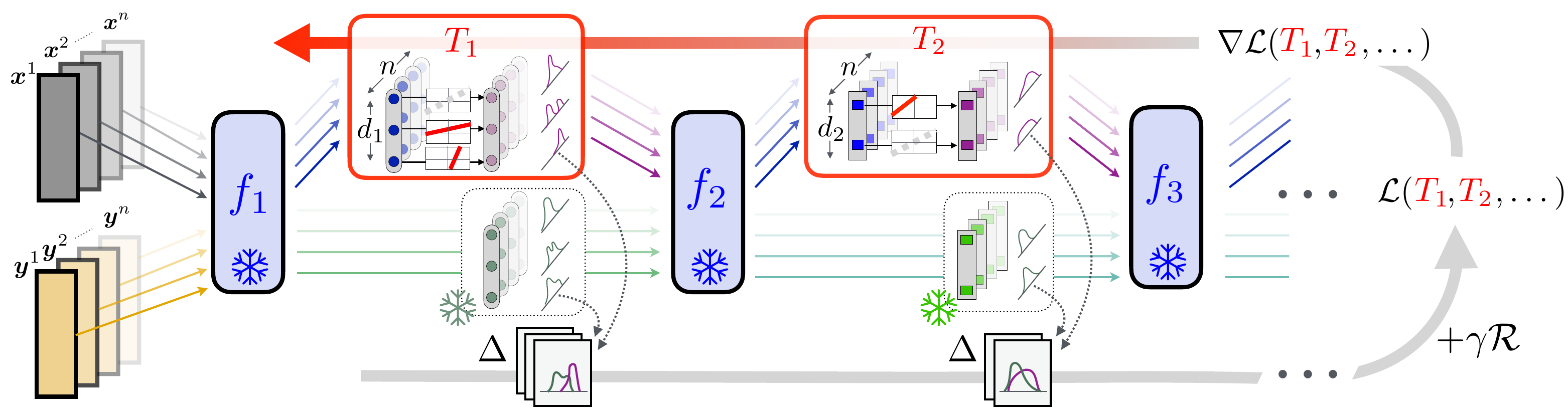}
    \caption{Given a frozen computational graph (\blue{blue}) of $L+1$ layers of interest, we interlace it with $L$ transport blocks (\red{red}). Each transport is defined as a collection of \textit{coordinate-wise} affine transformations, as displayed in the 3 and 2 boxes for maps $\RT{1}$ and $\RT{2}$ respectively. All transport maps are jointly trained to minimize a sum of distributional losses $\Delta$ between the neural activation distributions collected from samples $\vx_1,\dots,\vx_n \sim p$ (one shade of grey per sample) and $\vy_1,\dots,\vy_n\sim q$ (resp. yellow). 
    We learn the parameters of these maps \textit{jointly} by minimizing the penalized sum of $\Delta$ terms, where $\Delta$ is a 1D Wasserstein distances evaluated on the $d_\ell$ activations of layer $\ell$. Using a global optimization, we can consider sparsifying regularizers (included in $\mathcal{R}$), to, \eg select a sparse subset of activations that require interventions. 
    For instance, when adding a regularizer that promotes sparsity, both $\RT{1}$ and $\RT{2}$ do not intervene on one neuron, the first and the second, respectively.
    }
    \label{fig:fig_schema}
\end{figure*}

Our main contributions are:
\begin{itemize}[leftmargin=.2cm,itemsep=.0cm,topsep=0cm,parsep=2pt]

    \item We propose \method, a novel framework to steer activations based on affine optimal transport maps between activations trained jointly across layers (\eg $\RT{1}$, $\RT{2}$ in~\cref{fig:fig_schema}) with a global loss that enforces a global distributional alignment across all layers. Our method provides low-budget conditioning, with a continuous and theory-grounded application strength $\in [0, 1]$ (\eg \cref{fig:fig_intro}).

    \item We show how \method can be coupled with a \textit{sparse lasso} regularizer \cite{simon2013sparse} that can detect a small subset of activations that matter for a steering goal, reducing the intervened support by $100\times$, resulting in improved model utility.

    \item We show that \method learns effective interventions with as few as $32$ source and $32$ target unpaired data points, showing state-of-the-art toxicity mitigation among steering methods. 

    \item We validate fine-grained control of text-to-image (T2I) models, showing superior coherence with the prompt semantics while achieving the desired alignment goal.

\end{itemize}

\section{Related Work}
\label{sec:related_work}

We classify steering methods based on whether they require strong (paired data or use of a human supervision / oracle) or weak supervision (unpaired data and no further supervision).

\textbf{Strong supervision}. \caa \citep{rimsky2023steering} calculates the steering vector as the mean of differences between two sets of paired prompts. ReFT \citep{wuandarora2024reft} optimizes low-rank projections of representations. LoFiT \citep{yin2024lofit} learns a bias added to the representations of pre-selected attention heads. LoFiT training is similar to ReFT (although simpler, since only biases are trained). BitFit \citep{ben-zaken-etal-2022-bitfit} directly finetunes the bias terms of pre-selected linear layers. Most of these methods can be viewed as a simplified version of LoRA \citep{hulora}. Similarly to LoRA they also require either paired data or a human /oracle supervision during training.

\textbf{Weak supervision}. 
Some methods propose to add a vector to the activations. For example, \actadd \citep{turner2023activation} uses the difference between 2 prompts and Mean-AcT the difference in means~\cite{rodriguez2024controllinglanguagediffusionmodels}.
\iti \citep{li2024inference} uses a steering vector orthogonal to the hyperplane learned by a binary linear classifier on the activations from two sets of sentences. %
With a different approach, \aura by \citet{suau2024whispering} dampens activations proportionally to each neuron's ability to classify toxic and non-toxic sentences, effectively mitigating toxicity.
\repe, by \citet{zou2023representation}, does require paired data, however, we place this algorithm in the weak supervision family since it computes steering vectors based on a single prompt pair.
Closest to our work, \linearact \citep{rodriguez2024controllinglanguagediffusionmodels} uses an affine map to steer activations. In \Cref{sec:loss} we discuss in more detail the differences between our proposed \method with \linearact and other methods.

\section{End-to-end Learning of Steering Maps}\label{subsec:end2end}

We propose \method, a method to optimize activation-specific interventions in a joint manner. Our hypothesis is that a global estimation that exploits causal interdependencies between activations across layers is needed to maintain the model's utility while achieving the steering goal.

\subsection{Interventions Setup and Distributional Loss}
\label{sec:loss}
We consider a generative model and target a set of $L$ intermediate activation layers. We describe the model as a composition of $L+1$ abstract pretrained functions, where the output, given an input prompt $\vx\in\mathcal{S}$, can be written as 
$\vo = \Bf{L+1}\circ \Bf{L} \circ \dots \circ \Bf{1}(\vx).$
For convenience, we denote $d_\ell$ the dimension of the activations obtained at layer $\ell$, \ie the size of the activation vector $\Bf{\ell}\circ \dots\circ\Bf{1}(\vx)$. We interlace this pretrained computational graph of $L+1$ intermediate frozen layers with $L$ comparatively much simpler vector-to-vector maps:
$\vo = \Bf{L+1} \circ \RT{L} \circ \Bf{L} \circ \dots \circ \RT{2} \circ \Bf{2} \circ \RT{1} \circ \Bf{1}(\vx),$
where for $1\leq \ell\leq L$, the map $\RT{\ell}:\sR^{d_\ell}\rightarrow \sR^{d_\ell}$ acts on the intermediate activations observed after layer $\ell$, by altering its coordinates and outputting a vector of the same size.

\textbf{Layerwise Distributional Losses.}
We consider two distinct probability distributions over prompts, the source distribution $p$ and the desired target distribution $q$. For example, $p$ and $q$ could be such that a sample from the source distribution $p$ corresponds to toxic sentence, while a sample from the target distribution $q$ corresponds to a non-toxic sentence (see \cref{sec:tox}). We then view each sampled prompt through the lens of their sequence of L activations.
In practice, this means having  access to samples $\vx^1, \dots, \vx^n\sim p$ and $\vy^1, \dots, \vy^n\sim q$, tracking their execution trace of their $\ell\leq L+1$ activations, either modified through interleaved transports for samples from $p$:
\begin{equation}\label{eq:vx}
\vxi^i_\ell:=\RT{\ell}\circ \Bf{\ell} \circ \RT{\ell-1} \circ\dots \circ \RT{1} \circ \Bf{1}(\vx^i),\quad \ell\leq L,
\end{equation}
or ran through the original network for samples of $q$:
\begin{equation}\label{eq:vy}
\veta^j_\ell:=\Bf{\ell} \circ \dots \circ \Bf{1}(\vy^j),\quad \ell\leq L.
\end{equation}
Our goal is to learn \textit{jointly} all $L$ transport maps so that, for each $\ell\leq L$, the families of vectors $(\vxi^i_\ell)_i$ and $(\veta^j_\ell)_j$ are \textit{similar} with respect to a distributional metric $\Delta$ between probability measures, making the cost below small:
\begin{equation}\label{eq:cost}
\mathcal{C}(\RT{1}, \dots, \RT{\ell};(\vx^i)_i, (\vy^j)_j) = \sum_{\ell\leq L} \Delta((\vxi^i_\ell)_i, (\veta^j_\ell)_j).
\end{equation}
\textbf{Sliced Wasserstein Losses.}
To define $\Delta$ at each layer, we adopt the approach of~\citet{rodriguez2024controllinglanguagediffusionmodels} and sum $d_{\ell}$ univariate Wasserstein distances between the $d_{\ell}$ marginal distributions at each layer $\ell$. This choice is motivated by the fact that in the typical setting targeted in this work, we must deal with a high-dimensionality / low sample regime, $d_{\ell} \gg N$, that would hinder the use of more complex multivariate distributional losses that account more closely for cross-variable effects. We observe that adding univariate quantities yields a more robust loss estimation that translates to better downstream tasks than considering, e.g., Sinkhorn divergences~\citep{genevay2018learning}.

 To define $\Delta$ at each layer, we adopt the approach of~\citet{rodriguez2024controllinglanguagediffusionmodels} and use $d$ univariate Wasserstein distances between the $d$ marginal distributions. The activations can be arranged as matrices $U:=[\vxi^1_\ell, \dots, \vxi^n_\ell]$ and $V=[\veta^1_\ell, \dots, \veta^n_\ell]$, both in $\mathbb{R}^{n\times d_l}$. To compute their 1D-Wasserstein distance~\citep[Chap. 2]{santambrogio2015optimal}, these activations must be first sorted in increasing order along the feature axis:
$$\tilde{U} = \sort(U, \textrm{axis}=-1), \tilde{V} = \sort(V, \textrm{axis}=-1)$$
to define the sliced Wasserstein distance~\citep{rabin2012wasserstein} computed only on the \textit{canonical directions}, namely:
\begin{equation}\label{eq:sw}
\Delta(U, V):= \sum_{j=1}^d W_2^2(U_{\cdot j}, V_{\cdot j}) = \frac1n\sum_{j=1}^d \|\tilde{U}_{\cdot j} - \tilde{V}_{\cdot j}\|^2.
\end{equation}

\textbf{Differences with \iti, \linearact and \reft.}
Both \iti~\citep{li2024inference} and \linearact~\citep{rodriguez2024controllinglanguagediffusionmodels} optimize each $\RT{\ell}$ \textit{independently} across layers, minimizing a single distributional difference in closed form (\linearact) or learning a linear classifier (\iti), and assuming all other layers are frozen. While arguably much faster, this also generates causal inconsistencies, which can be partially resolved as in \linearact by using a sequential approach: when $\RT{\ell-1}$ is trained, $\RT{\ell-1}$ is reused to recompute all activation distributions used for $\RT{\ell}$. We claim that this suboptimality is to blame for poor generalization. This independent approach also precludes trade-offs when choosing which activations to turn on/off across layers, which we can easily be surfaced using sparsity regularizers. One fundamental difference with \reft~\citep{wu2024reft} is that \method does not need paired data, \ie sample $\vy^i$ does not need to be a counterfactual of $\vx^i$, which is the case for \reft. While counterfactual data is important for applications like translation, there is a vast amount of applications (\eg toxicity) where paired data is not available.

\subsection{Parameterization and Regularization}
Building on the approach outlined by~\citet{rodriguez2024controllinglanguagediffusionmodels}, we propose to parameterize each map $\RT{\ell}$ as an affine map for each activation $\ell\leq L$, namely for $z\in\sR^{d_\ell}$, one has
\begin{equation}\label{eq:map}
\RT{\ell}(z) := \ROmega{\ell} \odot z + \Rb{\ell}, \quad \ROmega{\ell}, \Rb{\ell}\in\sR^{d_\ell},
\end{equation}
where $\odot$ is the element-wise product. We write $\allw:=(\ROmega{1}, \dots, \ROmega{L})$ and $\allb:=(\Rb{1}, \dots, \Rb{L})$ for the collections of all scale and intercept parameters.
In what follows since each map ${\RT{\ell}}$ is entirely parameterized through its scale/intercept parameters $\allw, \allb$, we overload notations to define
$$
\mathcal{C}(\allw, \allb ;(\vx^i)_i, (\vy^j)_j) := \mathcal{C}(\RT{1}, \dots, \RT{\ell-1};(\vx^i)_i, (\vy^j)_j).$$

\textbf{Sparse Regularization.}
We propose to use a sparsity regularizer that will carry out both \textit{layer} and \textit{within-layer} selection of activations. This can be achieved by using structured regularization, using either 1-norms or 2-norms:
$$
\begin{aligned}
\mathcal{R}_1(\allw, \allb)&:= \sum_\ell \|\ROmega{\ell} - \mathbf{1}\|_1 + \|\Rb{\ell}\|_1\quad\text{and}\quad
\mathcal{R}_{\textrm{G}}(\allw, \allb)&:=\sum_\ell\sqrt{d_\ell}\left(\|\ROmega{\ell} - \mathbf{1}\|_2 + \|\Rb{\ell}\|_2\right).
\end{aligned}
$$
resulting in a \textit{sparse group lasso} regularizer~\citep{simon2013sparse,pmlr-v70-yoon17a}, $
\mathcal{R}:= \lambda_1 \mathcal{R}_{1} + \lambda_{\textrm{G}} \mathcal{R}_{\textrm{G}}$, which can be added to the cost to result in:
\begin{equation}
\label{eq:objective}
    \mathcal{L}(\allw, \allb) :=\!\!\!\! \E_{\substack{(\vx^i)_i \sim p,\\ (\vy^j)_j \sim q}}\!\!\!\!\!\!\!\left[ \mathcal{C}(\allw, \allb; (\vx^i)_i, (\vy^j)_j)\right] + \reg \mathcal{R}(\allw, \allb),
\end{equation}
where $\reg$ controls the amount of sparsity, \ie larger $\reg$ will result in fewer activations and/or layers being intervened on, with solution such that $\ROmega{} \approx \mathbf{1}$ and $\Rb{} \approx\mathbf{0}$. 

\textbf{On the choice of sparsity hyperparameters.}\quad
We only consider two hyperparameters at the moment, $\lambda_{1}$ and $\lambda_{\textrm{G}}$, which is relatively small since \method intervention models have none.
Empirically, we found that tuning $\gamma\in[0, 1]$ with $\lambda_{1} = \lambda_{\textrm{G}} = 1$  already provides interesting trade-offs between conditioning and utility as reported in \Cref{fig:tox_sparsity_data32,fig:tox_sparsity_data1024}. This also reduces the number of additional hyper-parameters to tune to just one ($\gamma$) and enables automatic layer selection (\Cref{fig:interpretability_app}).

\subsection{Optimization}
\textbf{Proximal SGD.}\quad
We optimize $\mathcal{L}$ in (\ref{eq:objective}) with proximal stochastic gradient descent (PSGD) with a learning rate of 0.1 and cosine decay. %
We assume access to 2 sets of unpaired $N$ prompts $(\vx^i)_{i=1}^N$ and $(\vy^i)_{i=1}^N$ and run PSGD on minibatches of activations $(\vxi^i_\ell)_i$ and $(\veta^i_\ell)_i$ of size $n$. Note that activations originating from $\vy^i$ use the untouched network, and can be pre-computed beforehand. 
In \Cref{app:optimization} we provide details on the algorithm and proximal operators.

\section{Experimental Results}
\label{sec:results}

\subsection{Toxicity Mitigation in LLMs}
\label{sec:tox}

We analyze the effectiveness of \method at the important task of toxicity mitigation. To that end, we compare with prompting, \caa~\citep{rimsky2023steering}, \reft~\citep{wu2024reft}, \iti~\citep{li2024inference} and \linearact~\citep{rodriguez2024controllinglanguagediffusionmodels} on three LLMs ranging from 1.5B to 7B parameters, by aligning the activations of $N=32$ toxic to $32$ non-toxic sentences sampled from the Jigsaw dataset~\citep{jigsaw-toxic-comment-classification-challenge}.

\textbf{Toxicity Metrics.} We evaluate toxicity mitigation on the \textit{RealToxicityPrompts} (RTP) dataset~\citep{gehman2020realtoxicityprompts} and the \textit{Thoroughly Engineered Toxicity} (TET) dataset~\citep{tet_dataset}. For RTP, we follow \citet{rodriguez2024controllinglanguagediffusionmodels} 
by sampling 1000 prompts from the dataset and let the model (intervened or not) complete them. For TET, we use the 2546 prompts provided. Then, we score the generations using the open-source Roberta toxicity classifier (\RTC)~\citep{logacheva-etal-2022-paradetox}. We report \toxclsrtp and \toxclstet, the respective percentage of generations flagged as toxic on RTP and TET.

\textbf{Utility Metrics.} To measure whether the utility of the model is affected by these interventions, we report \pplwik, the perplexity obtained on a fixed set of 20k Wikipedia sentences~\citep{wikidump}, as well as the overall 5-shot accuracy on the MMLU compendium~\citep{hendryckstest2021}.

\textbf{Oracle Baselines.}
In addition, we introduce two baselines that require a strong supervision signal directly from the \RTC~oracle, \ie the classifier used to compute \textit{test-time metrics}, giving them a significant advantage. More precisely, we train a LoRA adapter and LoFIT~\citep{yin2024lofit} using the RTC labels, yielding the oracle methods \lorartc~and \lofitrl. We report in \Cref{tab:toxicity_main} only the results of \lofitrl~as it performed better than \lorartc~(available in ~\cref{tab:toxicity_full}).
See~\cref{app:rlhf} for details on the oracle training protocol.

\textbf{Setup.} All methods have access to only 32 toxic and 32 non-toxic sentences (unpaired). We optimize \method for 1K steps using SGD and a learning rate of 0.1. As we focus on the benefit of using an end-to-end loss, we use $\gamma=0$ for \method (see Section~\ref{sec:tox_sparsity} for $\gamma$'s impact).
For \linearact and \caa, we use their default settings, and set intervention strength to $\lambda=1$. For \reft, we train for 10 epochs (selected with an epoch sweep). 
For \iti we use $\lambda=0.5$, obtained through grid search. We evaluate toxicity mitigation by intervening upon different layers, and report the best overall layer type per method, namely {\texttt{.*post.*layernorm}} for \method and \texttt{.*o\_proj} for \iti and \linearact, according to the Huggingface implementation of the models. Both \caa and \reft intervene upon the residuals of the Transformer blocks, as suggested in their original works. For \caa we run two baselines, intervening on all layers (reported in \Cref{tab:toxicity_main}), and intervening on the middle layer of each model (as in the original paper, reported in ~\cref{tab:toxicity_full} given its worse performance). Additionally, we include prompting as a conditioning strategy, where the model is preprompted with \textit{``Continue the text in a non-toxic way:''}.

\definecolor{Gray}{gray}{0.95}
\begin{table}[th]
    \centering
    \resizebox{0.83\columnwidth}{!}{
\begin{tabular}{llllrrrr}
\toprule
Model & Method & \#Params & \toxclsrtp $(\downarrow)$ & \toxclstet $(\downarrow)$ & \pplwik $(\downarrow)$ & MMLU $(\uparrow)$ \\

\arrayrulecolor{black} \midrule
\rowcolor{Gray} \multirow{9}{*}{\cellcolor{white}\qwenonefiveb} &  None & - &  $3.00_{\color{gray}{\,0.54}}$ & $23.09_{\color{gray}{\,0.67}}$ & $13.67_{\color{gray}{\,0.00}}$ & $60.95_{\color{gray}{\,0.00}}$ \\
\rowcolor{Gray}\cellcolor{white} & LoFIT-RL & 0.86M & $0.37_{\color{gray}{\,0.06}}$ & $4.36_{\color{gray}{\,0.00}}$ & $14.12_{\color{gray}{\,0.07}}$ & $59.74_{\color{gray}{\,0.14}}$ \\
\arrayrulecolor{gray!50} \cmidrule{2-7}

  &  \prompt & - & $4.07_{\color{gray}{\,0.38}}$ & $21.02_{\color{gray}{\,1.44}}$ & $13.65_{\color{gray}{\,0.00}}$ & $60.96_{\color{gray}{\,0.00}}$ \\
  &  \caa\hspace{-1mm}${}^\star$ & 0.043M &  $1.15_{\color{gray}{\,0.37}}$ & $5.77_{\color{gray}{\,2.14}}$ & $\red{19.30}_{\color{gray}{\,2.76}}$ & $\red{37.67}_{\color{gray}{\,6.95}}$ \\
 &  \reft\hspace{-1mm}${}^\star$ & 0.39M &  $2.57_{\color{gray}{\,0.60}}$ & $18.17_{\color{gray}{\,3.04}}$ & $15.58_{\color{gray}{\,0.52}}$ & $58.84_{\color{gray}{\,0.23}}$ \\

 &  \iti & 0.043M &  $1.87_{\color{gray}{\,0.21}}$ & $18.16_{\color{gray}{\,0.62}}$ & $12.39_{\color{gray}{\,0.09}}$ & $60.88_{\color{gray}{\,0.08}}$ \\
 &  \linearact & 0.086M &  $1.50_{\color{gray}{\,0.35}}$ & $13.88_{\color{gray}{\,1.72}}$ & $13.88_{\color{gray}{\,0.16}}$ & $60.09_{\color{gray}{\,0.25}}$ \\
 &  \method & 0.086M &  $\textbf{1.07}_{\color{gray}{\,0.46}}$ & $\textbf{12.70}_{\color{gray}{\,0.74}}$ & $14.10_{\color{gray}{\,0.07}}$ & $59.97_{\color{gray}{\,0.16}}$ \\
\arrayrulecolor{black} \midrule

\rowcolor{Gray} \multirow{9}{*}{\cellcolor{white}\gemmatwob} & None & - &  $4.00_{\color{gray}{\,0.45}}$ & $13.39_{\color{gray}{\,1.42}}$ & $14.79_{\color{gray}{\,0.00}}$ & $53.03_{\color{gray}{\,0.00}}$ \\
\rowcolor{Gray}\cellcolor{white} & LoFIT-RL & 0.11M & $0.40_{\color{gray}{\,0.20}}$ & $1.76_{\color{gray}{\,0.00}}$ & $15.43_{\color{gray}{\,0.08}}$ & $52.17_{\color{gray}{\,0.17}}$ \\
\arrayrulecolor{gray!50} \cmidrule{2-7}
 
  &  \prompt & - &  $4.60_{\color{gray}{\,0.36}}$ & $12.32_{\color{gray}{\,0.67}}$ & $14.81_{\color{gray}{\,0.00}}$ & $53.18_{\color{gray}{\,0.00}}$ \\
  &  \caa\hspace{-1mm}${}^\star$ & 0.06M &  $0.80_{\color{gray}{\,0.00}}$ & $2.44_{\color{gray}{\,1.99}}$ & $\red{23.52}_{\color{gray}{\,2.67}}$ & $\red{26.86}_{\color{gray}{\,0.08}}$ \\
 &  \reft\hspace{-1mm}${}^\star$ & 0.54M &  $2.85_{\color{gray}{\,0.49}}$ & $11.15_{\color{gray}{\,1.91}}$ & $\red{19.93}_{\color{gray}{\,0.30}}$ & $48.99_{\color{gray}{\,1.34}}$ \\

&  \iti & 0.06M &  $1.17_{\color{gray}{\,0.60}}$ & $7.15_{\color{gray}{\,0.92}}$ & $14.00_{\color{gray}{\,0.11}}$ & $52.78_{\color{gray}{\,0.23}}$ \\
&  \linearact & 0.12M &  $1.60_{\color{gray}{\,0.32}}$ & $7.76_{\color{gray}{\,0.39}}$ & $14.78_{\color{gray}{\,0.12}}$ & $52.43_{\color{gray}{\,0.57}}$ \\
&  \method & 0.24M &  $\textbf{0.73}_{\color{gray}{\,0.10}}$ & $\textbf{4.02}_{\color{gray}{\,0.68}}$ & $15.46_{\color{gray}{\,0.21}}$ & $52.22_{\color{gray}{\,0.40}}$ \\

\arrayrulecolor{black} \midrule
\rowcolor{Gray} \multirow{9}{*}{\cellcolor{white}\qwensevenb} & None & - &  $3.92_{\color{gray}{\,0.59}}$ & $25.16_{\color{gray}{\,0.92}}$ & $10.67_{\color{gray}{\,0.00}}$ & $74.26_{\color{gray}{\,0.00}}$ \\
\rowcolor{Gray}\cellcolor{white} & LoFIT-RL & 0.10M & $1.10_{\color{gray}{\,0.38}}$ & $7.11_{\color{gray}{\,0.30}}$ & $10.91_{\color{gray}{\,0.16}}$ & $73.87_{\color{gray}{\,0.17}}$ \\
\arrayrulecolor{gray!50} \cmidrule{2-7}

  &  \prompt & - &  $6.80_{\color{gray}{\,0.00}}$ & $21.22_{\color{gray}{\,0.21}}$ & $10.65_{\color{gray}{\,0.00}}$ & $74.23_{\color{gray}{\,0.00}}$ \\
  &  \caa\hspace{-1mm}${}^\star$ & 0.10M &  $1.20_{\color{gray}{\,0.00}}$ & $9.25_{\color{gray}{\,3.07}}$ & $12.83_{\color{gray}{\,0.00}}$ & $\red{48.58}_{\color{gray}{\,0.00}}$ \\
 &  \reft\hspace{-1mm}${}^\star$ & 0.90M &  $3.33_{\color{gray}{\,0.96}}$ & $20.38_{\color{gray}{\,2.37}}$ & $13.80_{\color{gray}{\,1.20}}$ & $70.43_{\color{gray}{\,0.60}}$ \\

 &  \iti & 0.10M &  $2.63_{\color{gray}{\,0.44}}$ & $19.98_{\color{gray}{\,1.24}}$ & $9.63_{\color{gray}{\,0.03}}$ & $74.08_{\color{gray}{\,0.05}}$ \\
 &  \linearact & 0.20M &  $2.72_{\color{gray}{\,0.46}}$ & $21.64_{\color{gray}{\,2.00}}$ & $11.42_{\color{gray}{\,0.34}}$ & $72.18_{\color{gray}{\,0.16}}$ \\
 &  \method & 0.20M &  $\textbf{1.95}_{\color{gray}{\,0.48}}$ & $\textbf{14.95}_{\color{gray}{\,0.92}}$ & $10.91_{\color{gray}{\,0.35}}$ & $73.67_{\color{gray}{\,0.05}}$ \\
\arrayrulecolor{black}\bottomrule
\end{tabular}}
\vskip 3mm
\caption{Toxicity mitigation on the RTP and TET datasets using three different models, \qwenonefiveb, \gemmatwob, and \qwensevenb. Strongly degraded utility is marked in \red{red}. We report results at low data regime ($N=32$ sentences to estimate the interventions). See \cref{app:tox} for more models, baselines and an ablation with larger training size and different intervention layers. Results for \method improve significantly on \iti and \linearact with similar impact on utility metrics. The quality of these interventions is often on par with the oracle baseline \lofitrl in terms of utility, although the strong oracle supervision yields better mitigation. \textit{$^\star$\caa and \reft are designed to use paired data, which is not available in the toxicity setting.}}
    \label{tab:toxicity_main}
\vspace{-1em}
\end{table}

\begin{figure*}[t!]
    \centering
    \includegraphics[trim={0 0 0 0},clip,width=\textwidth]{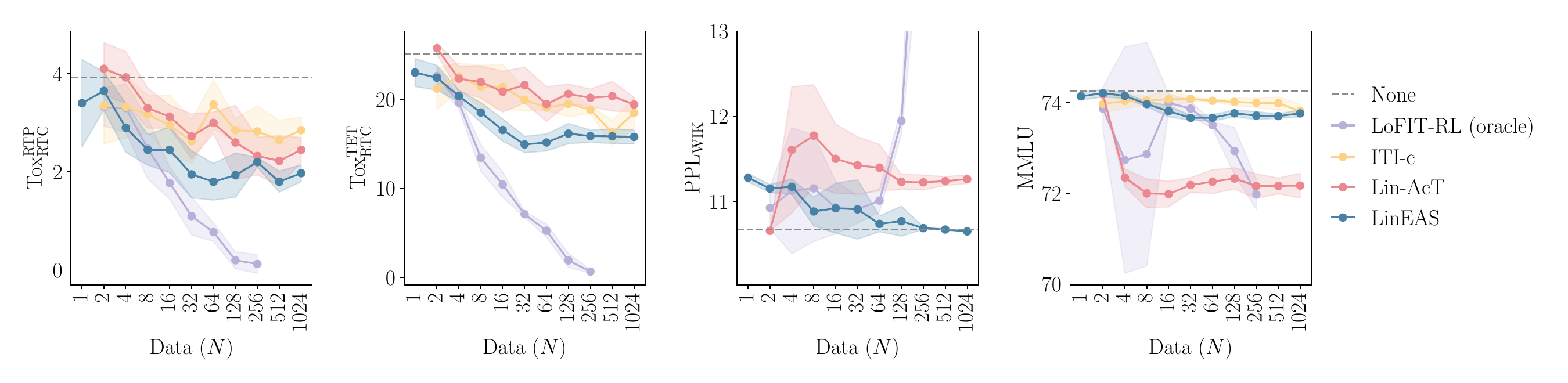}
    \vskip -1mm
    \caption{\textbf{\method is effective at low data regime.} We study toxicity mitigation (two left-most plots) and utility (two right-most plots) as a function of the amount of data available to learn interventions. \method shows better performance (low toxicity and utility close to original dashed lines) for low data, and stable performance for $N\geq 32$.}
    \label{fig:data_ablation}
\end{figure*}

\textbf{Results.} \Cref{tab:toxicity_main} summarizes the toxicity mitigation experiments averaged over 4 generation seeds (and RTP samplings). \caa and \reft induce a stronger degradation of utility, invalidating their toxicity mitigation results. Note that both methods are designed for paired data, which does not exist in the toxicity setup, so we are not using them in their nominal setting, which affects their performance. 
Prompting is not effective for the models tested, and even increases \toxclsrtp. 
\method achieves a consistent toxicity mitigation, outperforming all steering methods at this low data regime. For example, \method reduces \gemmatwob toxicity by $5.5\times$, getting closer to the mitigation obtained with the oracle \lofitrl. 
In terms of utility, \method shows a minimal degradation, with values very similar to the utility incurred by the oracle \lofitrl. In absolute terms, \method reduces MMLU by less than 1 point and increases \pplwik by less than 0.6.
In \Cref{tab:toxicity_layer} (\Cref{app:tox}) we  show that \method is much more robust to the choice of layer than \iti and \linearact. Additionally, we also provide in \Cref{fig:data_ablation} (and \Cref{tab:toxicity_large}) an analysis on a higher data regimes, up to $N=1024$ sentences, showing that \method is also reliable in such scenario.
Beyond the differences in parameter sizes between \iti, \linearact and \method on the one hand, and the \lorartc approach on the other, we also note that the compute needed to train these methods is significantly different:~in the low data regime $N=32$, estimating each method on a Nvidia A100-80Gb GPU and \qwensevenb takes 37s (\iti), 30s (\linearact) leveraging closed forms, 500s (\method for 1K steps) and 27300s for \lofitrl, see~\cref{app:hardware} for a deeper analysis.

\noindent \textbf{User Study.} We complement the quantitative results with a user preference study to evaluate the perceived quality of continuations generated by different intervention methods. Our findings indicate a strong preference for \method over three other alternatives: \linearact, \iti, and no intervention (identity). Specifically, users preferred LinEAS in $57.70\%$ of cases compared to $18.43\%$ for \linearact and $11.67\%$ for \iti. We provide more details in~\cref{app:user-study}.

\subsection{Effect of Data  on Toxicity Mitigation}
\label{sec:tox_data}

With the same toxicity setting as in \Cref{sec:tox}, we ablate the amount of data used to estimate interventions using \qwensevenb (best model studied in terms of MMLU). We sweep $N$ from 1 to 1024, meaning we have access to $N$ toxic and $N$ non-toxic sentences. Note that the results in \Cref{tab:toxicity_main}  correspond to $N=32$. In \Cref{fig:data_ablation} we plot the evolution of toxicity (\toxclsrtp and \toxclstet) and utility (\pplwik and MMLU) as a function of $N$, averaged over 4 random sweeps (standard deviation as shaded areas). \method achieves superior toxicity mitigation even at very low data regimes, while maintaining utility close to the original model (horizontal dashed lines) and the \lofitrl oracle. Moreover, \method's performance is more stable for a large range of $N$ (32 to 1024), even more stable than the oracle which diverges for $N>128$. Note that we fix the hyper-parameters for all the methods, including the oracle, which shows to be more sensitive to the training setting.

\begin{figure*}[t!]
    \centering
    \includegraphics[trim={0 0 0 0},clip,width=\textwidth]{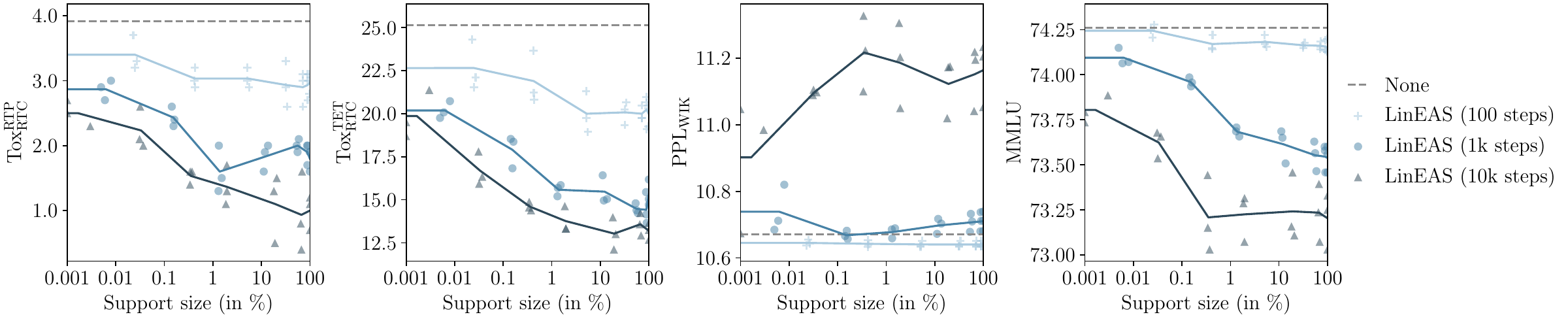}
    \vskip -1mm
    \caption{\textbf{Sparsity improves utility while mitigating toxicity}. Toxicity results on \qwensevenb using only 32 sentences, at different levels of sparsity $\reg$ that result in different support sizes (x axis). At 1K optimization steps, with a support of about 1\% we maintain similar toxicity (left, center-left) while \pplwik decreases (center-right) and MMLU increases (right). Note that too long optimizations (10k steps) might harm utility, due to overfitting. Similarly, short optimizations (\eg 100 steps) and strong sparsity leads to low conditioning (mild toxicity mitigation).}
    \label{fig:tox_sparsity_data32}
\end{figure*}

\subsection{Effect of Sparsity  on Toxicity Mitigation}
\label{sec:tox_sparsity}

Intuitively one should only steer the smallest set of activations needed to achieve a desired goal in order to preserve utility and keep most of the model's inference graph untouched. In this section, we explore how sparsity affects toxicity mitigation in the setup of \Cref{sec:tox} as we increase $\reg$ from $0$ to $0.1$. Increasing $\reg$ results in less activations being \textit{transported}, which we measure as $\mathrm{support} = \|(\allw \neq \mathbf{1}) + (\allb \neq \mathbf{0})\|_0$, \ie all activations transported either by rescaling or shifts. 

In \Cref{fig:tox_sparsity_data32} we show how \toxclsrtp and \toxclstet, as well as the utility \pplwik and MMLU, evolve as the sparse support decreases (x axis), for \qwensevenb in the low ($N=32$) data regime. We show the results of 3 sweeps of $\reg$ with different random seeds (markers), and plot the average at each $\reg$ level (line). Note that at $\reg=0$ the support is approximately 100\%. In the case of 1K steps (reported in \Cref{tab:toxicity_main}), one can afford reducing the support to about $1\%$ and still maintain the toxicity mitigation values at 100\% support. Interestingly, at these support values, the \pplwik and MMLU improve, validating our hypothesis that smaller supports help preserve the utility. We also observe that short optimizations (\eg 100 steps) lead to mild conditioning (poor toxicity mitigation) while long optimizations (\eg 10k steps) lead to a gradual degradation in utility. In \Cref{fig:tox_sparsity_data1024} (\Cref{app:tox_sparsity}) we show the same plot for the high data regime, with similar conclusions. Additionally, in \Cref{app:men_similarity} we study how the similarity of \method interventions is correlated with human judgment, on pairs of concepts from the MEN dataset \citep{men_dataset}, showing strong correlation when using sparsity. We provide a similar analysis for T2I generation in~\cref{app:t2i-sparsity}.

\begin{figure*}[htb]
    \centering
    \includegraphics[width=\linewidth]{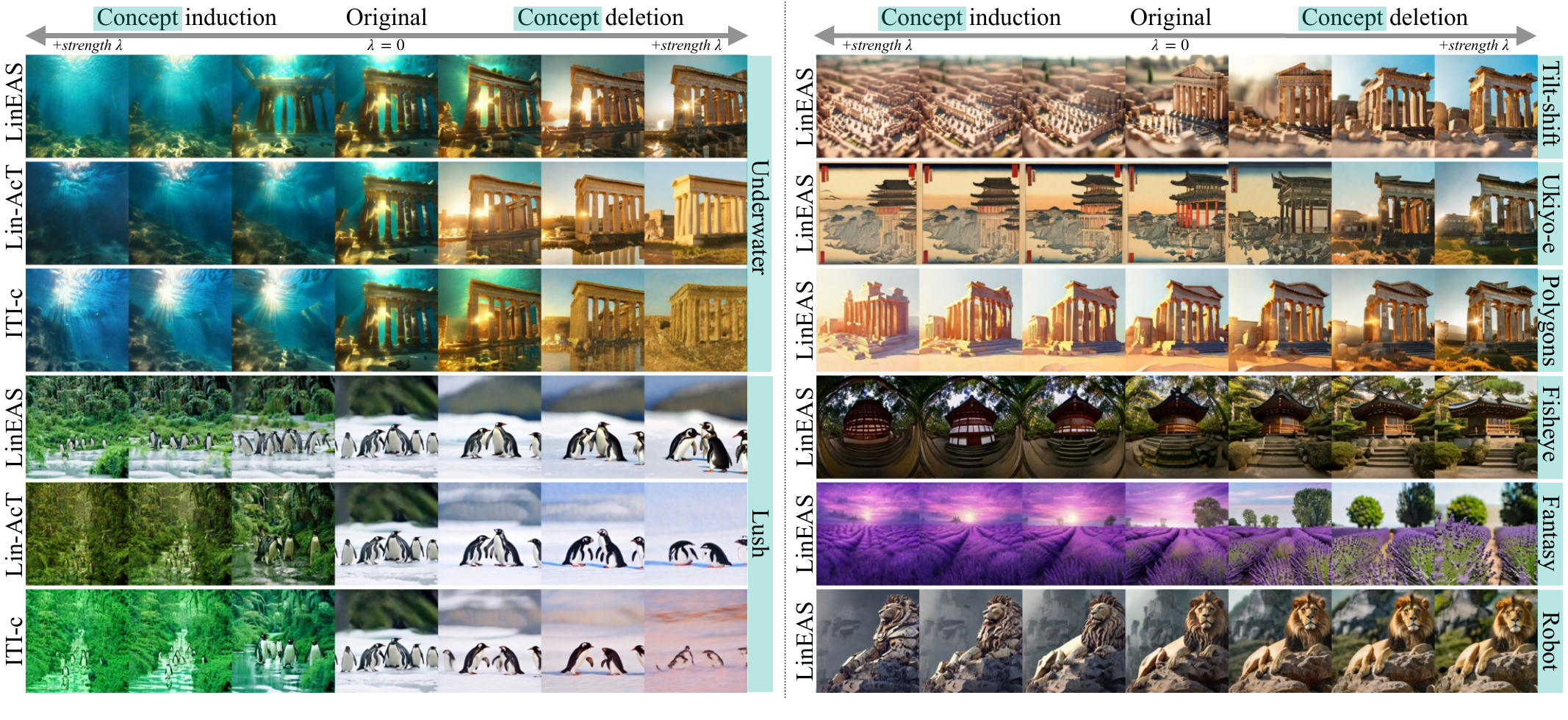}
    \vskip -1mm
    \caption{Generations using DMD2~\citep{yin2024dmd2}. (Left) Concept deletion using \iti, \linearact and \method for two different concepts, starting from prompts that contain the concept. \method shows a more gradual deletion ($\lambda=0.4, 0.7, 1$), and better preservation of the original image ($\lambda=0)$. (Right) Qualitative examples of \method on 6 more concepts. We also show that inverting the steering maps surprisingly results in concept induction, probably due to strong structure in activation space. \method also outperforms the other methods under this setting.}
    \label{fig:diffusion_comparison}
\end{figure*}
\begin{figure}[t]
\begin{minipage}{0.49\linewidth}
\centering
\begin{tabular}{l@{\hskip 3pt}c@{\hskip 3pt}c@{\hskip 3pt}c}
\toprule
Method & $\uparrow$ User Pref. & $\uparrow$ IMGSc. 
 & $\downarrow$ CLIPSc. \\
\midrule
\iti & $12.4_{\color{gray}{\,5.5}}\%$ & $0.24_{\color{gray}{\,0.19}}$ & $0.19_{\color{gray}{\,0.02}}$ \\
\linearact & $24.4_{\color{gray}{\,7.0}}\%$ & $0.45_{\color{gray}{\,0.21}}$ & $0.18_{\color{gray}{\,0.03}}$ \\
\method & $\mathbf{63.3}_{\color{gray}{\,6.6}}\%$  & $\mathbf{0.66}_{\color{gray}{\,0.19}}$ & $0.18_{\color{gray}{\,0.03}}$ \\
\bottomrule
\end{tabular}
\captionof{table}{\method mitigates concepts on DMD2~\citep{yin2024dmd2} while staying perceptually similar to the original image. Users prefer \method 63.3\% of the time (left)  since it maintains a higher fidelity to the non-intervened original model when using the same prompt (center), and matches other methods at concept removal (right). Results were obtained with $\lambda=1$ and they are aggregated across all concepts.}
\label{tab:t2i-quantitative}
\end{minipage}\hfill%
\begin{minipage}{0.49\linewidth}
\centering
\includegraphics[trim={0 0 0 0},clip,width=0.8\linewidth]{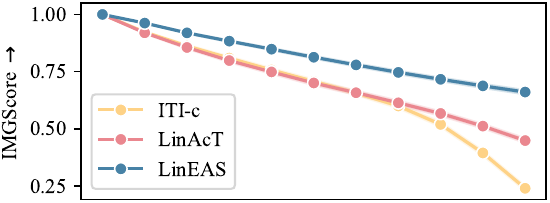}
\includegraphics[trim={0 0 0 0},clip,width=0.8\linewidth]{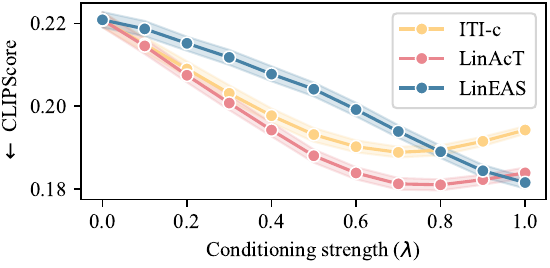}
\vskip -2mm
\captionof{figure}{\imgscore and \clipscore at multiple conditioning strengths ($\lambda$). (top) \method is more faithful to DMD2 with $\lambda=0$. (bottom) \method is more consistent and predictable at mitigation.}
\vskip -4mm
\label{fig:t2i-curves}
\end{minipage}
\end{figure}
\subsection{Steering Text-to-Image Generation Models}
\label{sec:results_diffusion}

In this section, we explore how \method can be used to remove the presence of concepts in text-to-image (T2I) generation: a task that plays a key role in generation alignment, similar to toxicity mitigation in LLMs. We preserve the focus on low data and compute budget,
so we apply \method to DMD2~\citep{yin2024dmd2}, a recent single-step text-to-image generation model distilled from SDXL~\citep{podellsdxl} with a GAN~\citep{goodfellow2014generative} loss. We note that there are diffusion guidance methods available to condition these models; however, they typically require multiple denoising steps (even up to 200)~\citep{Ye2024-TFG, Guo2024-GGDM} and, at times, trajectory resampling and evaluation strategies \citep{Singhal2025-FK}. These additional overheads make them suitable for settings with higher compute budget, which differs from the low-compute setting we are interested in. 

\textbf{Setup.} As in \citep{rodriguez2024controllinglanguagediffusionmodels}, we modulate the strength of \method applied to all \texttt{layernorm} layers by introducing a scale $0\leq \lambda\leq 1$ when applying interventions. Intuitively $\lambda=0$ results in no intervention, while $\lambda=1$ carries out a full \method transport, any value in between reflecting a gradual change. Following~\Cref{sec:tox}, we focus on concept mitigation/removal and we use $32$ samples for each the source and the target distribution. We train \method for 1000 steps with batch size 4, AdamW, learning rate of $1e^{-4}$ and $\gamma=0$. Find additional implementation details in~\Cref{app:t2i}. 
We compare \method with \iti and \linearact, also weakly supervised methods that work with \textit{unpaired data}.

\textbf{Data.} We query an open-source LLM for a diverse set of prompts covering 3 different conditioning categories and 5 different concepts per category with 32 prompts per concept, totalling 480 prompts. \textbf{Styles}: \texttt{vaporwave, lush, low-poly, ukiyo-e, fantasy}; \textbf{objects}: \texttt{robots, axolotl, book, car, hourglass}; and \textbf{perspective}: \texttt{macro, fisheye, bokeh, underwater, and tilt-shift}. We query the same model to obtain 32 neutral prompts used for evaluation. We provide a complete description and the prompts themselves in \cref{app:t2i-prompts}.

\textbf{Metrics.} We measure (1) \textbf{\clipscore}~\citep{hessel2021clipscore}, the cosine similarity between CLIP embeddings of the generated images and the concept description.
(2) \textbf{\imgscore}, the cosine similarity between DINOv2-small~\citep{oquab2024dinov} image embeddings generated with $\lambda>0$ and images generated with $\lambda=0$ using the same prompts. 
\clipscore assesses whether the generations are truthful to the desired style, and \imgscore whether they are perceptually similar to those generated without intervention.

\textbf{User study.} Aiming at complementing the quantitative analysis, we also run a user study. We consider 20 prompts $\times$ 5 conditioning concepts, yielding a total of 100 pairs. We generate an original image and steer it using \iti, \linearact, or \method on 5 concepts. We ask a pool of 10 participants to select their preferred output (a total of 1000 evaluations), showing a strong preference for \method, as reported in \Cref{tab:t2i-quantitative} (left). The actual question asked to the participants is \textit{"Which image blends best prompt and style, while remaining faithful to the untouched output?"}.

\textbf{Results.} \Cref{tab:t2i-quantitative} summarizes the results for the text to image evaluation. We found that 63\% of the users prefer \method (24.4\% for \linearact, and 12.4\% for \iti). These results are in agreement with the automated metrics: \method is \textit{significantly} more faithful to the images produced by DMD2 with $\lambda=0$ for the same prompts, with an \imgscore of 0.66 compared to 0.45 for \linearact and 0.24 for \iti while all methods achieve a similar \clipscore. We report more granular per-concept scores in~\Cref{app:t2i-quantitative-detail}. \Cref{fig:t2i-curves} explores how \imgscore and \clipscore change with $\lambda$. We find that \method is consistently more faithful to DMD2 with $\lambda=0$ in terms of \imgscore (top) while showing a strong linear correlation with \clipscore, making \method more consistent and predictable.

\textbf{\vflip{LinEAS}}. Surprisingly, inverting the affine maps in \method: $\RT{\ell}^{-1}(z) := (z - \Rb{\ell}) \odot \frac{1}{\ROmega{\ell}}$ tends to negate the conditioning thus switching from mitigation to induction and vice-versa (see ``concept induction'' in \cref{fig:diffusion_comparison}). We speculate that this behavior is due to a strong structure in the activation space. Quantitative results using the inverse maps can be found in~\Cref{app:t2i-quantitative-detail}.

\subsection{Layer Selection Analysis}
\label{sec:layer_ablation}

To evaluate the robustness of \method with respect to the choice of intervened layers, we conducted a sweep over different layer types within the DMD2 UNet. The results, averaged over 15 concepts, are presented in Table~\ref{tab:layer_ablation}. We measure image consistency using IMGScore (where higher is better) and concept removal using CLIPScore (where lower is better).

Our findings indicate that while intervening on all layer normalization modules --- the setting used in our main experiments --- provides the best trade-off between the two metrics, \method demonstrates robust performance across all tested layer configurations. This suggests that activation steering is not overly sensitive to the specific layer choice in UNets.

\begin{table}[h!]
\centering
\resizebox{0.83\columnwidth}{!}{\begin{tabular}{l c c c}
\toprule
\textbf{Intervened Layers} & \textbf{\# Modules} & \textbf{IMGScore} $\uparrow$ & \textbf{CLIPScore} $\downarrow$ \\
\midrule
\texttt{unet} - all layer norms & 256 & $0.714 \pm 0.054$ & $0.131 \pm 0.033$ \\
\texttt{unet.transformer} - all FeedForward & 70 & $0.780 \pm 0.053$ & $0.137 \pm 0.031$ \\
\texttt{unet} attentions - K and Q projections & 280 & $0.935 \pm 0.016$ & $0.157 \pm 0.028$ \\
\texttt{unet} attentions - V projections & 140 & $0.818 \pm 0.051$ & $0.140 \pm 0.031$ \\
\texttt{unet} attentions - in projections & 11 & $0.923 \pm 0.017$ & $0.155 \pm 0.028$ \\
\texttt{unet.resnet} - all layer norms & 34 & $0.896 \pm 0.023$ & $0.156 \pm 0.027$ \\
\bottomrule
\end{tabular}}

\caption{Study on the choice of intervened layers for LinEAS in the DMD2 UNet. We report image consistency (IMGScore $\uparrow$) and concept removal (CLIPScore $\downarrow$), averaged over 15 concepts. The method shows robustness, with the default setting (all layer norms) offering the best trade-off.}
\label{tab:layer_ablation}
\end{table}

\section{Limitations and Open Problems}
\label{sec:limitations}
While the field of activation steering has gained considerable momentum there are some limitations that affect practically every method, including ours.

\textbf{Compositionality.} One such limitation is the ability to compose multiple interventions (learnt separately) so that multiple steering objectives are satisfied at the same time. We refer to this ability as \textit{compositionality}, and we argue its difficulty lies in the fact that multiple interventions can interfere with one another and produce unexpected results. Our initial hypothesis was that sparsity could help mitigate such unwanted interference, since interventions for different concepts would act on (almost) disjoint sets of neurons. In \cref{app:composition} we present an analysis where we intervene for two concepts simultaneously, with different sparsity $\gamma$s. 
We find that \method outperforms even prompting, which reinforces the value of steering for compositionality. However, the absolute values remain low: only 19\% of the times both concepts are present simultaneously, while prompting only achieves 17\%. We observe that most of the gain comes from the end-to-end optimization, which reaches 16\% probability without sparsity, which is a $15\times$ improvement over \linearact (no end-to-end optimization).  Adding group lasso regularization improves results of \method by an additional 3\%.
These results show, on the one hand that compositionality is still a challenging task, and on the other hand that there is room to investigate more suitable sparse regularizers in order to overcome the current limitations.

\textbf{Intervention selectivity.} Another common limitation of steering mechanisms that also affects \method the intervention is applied to all tokens, usually with the goal of keeping inference cost in a budget. Finding a way to selectively apply the intervention while avoiding adding inference cost (\eg avoid using a classifier to decide on which tokens to intervene or not) remains an open problem.

\section{Conclusion}
We propose \method, a novel framework to learn lightweight interventions on activations to steer model generation towards a desired property. \method achieves state-of-the-art performance among steering methods on safety applications, such as avoiding toxic outputs, or style changes, both for LLMs (\gemmatwob, \qwenonefiveb and \qwensevenb) and text-to-image generation (DMD2). Our approach learns a set of univariate maps that reshape a source to a target activation distribution, with an improved loss that yields improved controllability and robustness.
Unlike previous methods, such as~\cite{rodriguez2024controllinglanguagediffusionmodels,li2024inference}, which require local adjustments and manual layer selection, our method optimizes the transformation globally in an end-to-end fashion.
We find that this global optimization makes \method more precise than layer-wise training, where errors accumulate layer after layer. This makes \method more effective with very low data (32 exemplar sentences from both source and target sets) while its distributional nature provides an intervention strength parameter that is continuous and bounded, making it more intuitive to apply.
This approach also allows for the incorporation of flexible regularizers, such as group-sparse and sparse, allowing for automatic selection of layers and/or neurons, leading to improved utility.
The code will be publicly available shortly.

\newpage

\bibliographystyle{unsrtnat}
\bibliography{biblio}

\begin{thebibliography}{44}
\providecommand{\natexlab}[1]{#1}
\providecommand{\url}[1]{\texttt{#1}}
\expandafter\ifx\csname urlstyle\endcsname\relax
  \providecommand{\doi}[1]{doi: #1}\else
  \providecommand{\doi}{doi: \begingroup \urlstyle{rm}\Url}\fi

\bibitem[Wei et~al.(2022)Wei, Bosma, Zhao, Guu, Yu, Lester, Du, Dai, and Le]{weifinetuned}
Jason Wei, Maarten Bosma, Vincent Zhao, Kelvin Guu, Adams~Wei Yu, Brian Lester, Nan Du, Andrew~M Dai, and Quoc~V Le.
\newblock Finetuned language models are zero-shot learners.
\newblock In \emph{International Conference on Learning Representations}, 2022.

\bibitem[Ouyang et~al.(2022)Ouyang, Wu, Jiang, Almeida, Wainwright, Mishkin, Zhang, Agarwal, Slama, Ray, et~al.]{ouyang2022training}
Long Ouyang, Jeffrey Wu, Xu~Jiang, Diogo Almeida, Carroll Wainwright, Pamela Mishkin, Chong Zhang, Sandhini Agarwal, Katarina Slama, Alex Ray, et~al.
\newblock Training language models to follow instructions with human feedback.
\newblock \emph{Advances in Neural Information Processing Systems}, 35:\penalty0 27730--27744, 2022.

\bibitem[Ho and Salimans(2022)]{ho2022classifier}
Jonathan Ho and Tim Salimans.
\newblock Classifier-free diffusion guidance.
\newblock \emph{arXiv preprint arXiv:2207.12598}, 2022.

\bibitem[Luo et~al.(2024)Luo, Wong, Trabucco, Huang, Gonzalez, Chen, Salakhutdinov, and Stoica]{luo2024stylus}
Michael Luo, Justin Wong, Brandon Trabucco, Yanping Huang, Joseph~E. Gonzalez, Zhifeng Chen, Russ Salakhutdinov, and Ion Stoica.
\newblock Stylus: Automatic adapter selection for diffusion models.
\newblock In \emph{The Thirty-eighth Annual Conference on Neural Information Processing Systems}, 2024.
\newblock URL \url{https://openreview.net/forum?id=3Odq2tGSpp}.

\bibitem[Baumann et~al.(2024)Baumann, Krause, Neumayr, Stracke, Sevi, Hu, and Ommer]{baumann2024continuous}
Stefan~Andreas Baumann, Felix Krause, Michael Neumayr, Nick Stracke, Melvin Sevi, Vincent~Tao Hu, and Bj{\"o}rn Ommer.
\newblock Continuous, subject-specific attribute control in t2i models by identifying semantic directions.
\newblock 2024.

\bibitem[Suau et~al.(2022)Suau, Zappella, and Apostoloff]{suau2022self}
Xavier Suau, Luca Zappella, and Nicholas Apostoloff.
\newblock Self-conditioning pre-trained language models.
\newblock In \emph{International Conference on Machine Learning}, pages 4455--4473. PMLR, 2022.

\bibitem[Rimsky et~al.(2023)Rimsky, Gabrieli, Schulz, Tong, Hubinger, and Turner]{rimsky2023steering}
Nina Rimsky, Nick Gabrieli, Julian Schulz, Meg Tong, Evan Hubinger, and Alexander~Matt Turner.
\newblock Steering llama 2 via contrastive activation addition.
\newblock \emph{arXiv preprint arXiv:2312.06681}, 2023.

\bibitem[Zou et~al.(2023)Zou, Phan, Chen, Campbell, Guo, Ren, Pan, Yin, Mazeika, Dombrowski, et~al.]{zou2023representation}
Andy Zou, Long Phan, Sarah Chen, James Campbell, Phillip Guo, Richard Ren, Alexander Pan, Xuwang Yin, Mantas Mazeika, Ann-Kathrin Dombrowski, et~al.
\newblock Representation engineering: A top-down approach to ai transparency.
\newblock \emph{arXiv preprint arXiv:2310.01405}, 2023.

\bibitem[Li et~al.(2024)Li, Patel, Vi{\'e}gas, Pfister, and Wattenberg]{li2024inference}
Kenneth Li, Oam Patel, Fernanda Vi{\'e}gas, Hanspeter Pfister, and Martin Wattenberg.
\newblock Inference-time intervention: Eliciting truthful answers from a language model.
\newblock \emph{Advances in Neural Information Processing Systems}, 36, 2024.

\bibitem[Rodriguez et~al.(2025)Rodriguez, Blaas, Klein, Zappella, Apostoloff, Cuturi, and Suau]{rodriguez2024controllinglanguagediffusionmodels}
Pau Rodriguez, Arno Blaas, Michal Klein, Luca Zappella, Nicholas Apostoloff, Marco Cuturi, and Xavier Suau.
\newblock Controlling language and diffusion models by transporting activations, 2025.

\bibitem[Yin et~al.(2024{\natexlab{a}})Yin, Ye, and Durrett]{yin2024lofit}
Fangcong Yin, Xi~Ye, and Greg Durrett.
\newblock Lofit: Localized fine-tuning on llm representations.
\newblock \emph{NeurIPS}, 2024{\natexlab{a}}.

\bibitem[Wu et~al.(2024{\natexlab{a}})Wu, Arora, Wang, Geiger, Jurafsky, Manning, and Potts]{wu2024reft}
Zhengxuan Wu, Aryaman Arora, Zheng Wang, Atticus Geiger, Dan Jurafsky, Christopher~D Manning, and Christopher Potts.
\newblock Reft: Representation finetuning for language models.
\newblock \emph{arXiv preprint arXiv:2404.03592}, 2024{\natexlab{a}}.

\bibitem[Simon et~al.(2013)Simon, Friedman, Hastie, and Tibshirani]{simon2013sparse}
Noah Simon, Jerome Friedman, Trevor Hastie, and Robert Tibshirani.
\newblock A sparse-group lasso.
\newblock \emph{Journal of computational and graphical statistics}, 22\penalty0 (2):\penalty0 231--245, 2013.

\bibitem[Wu et~al.(2024{\natexlab{b}})Wu, Arora, Wang, Geiger, Jurafsky, Manning, and Potts]{wuandarora2024reft}
Zhengxuan Wu, Aryaman Arora, Zheng Wang, Atticus Geiger, Dan Jurafsky, Christopher~D. Manning, and Christopher Potts.
\newblock {ReFT}: Representation finetuning for language models.
\newblock 2024{\natexlab{b}}.
\newblock URL \url{arxiv.org/abs/2404.03592}.

\bibitem[Ben~Zaken et~al.(2022)Ben~Zaken, Goldberg, and Ravfogel]{ben-zaken-etal-2022-bitfit}
Elad Ben~Zaken, Yoav Goldberg, and Shauli Ravfogel.
\newblock {B}it{F}it: Simple parameter-efficient fine-tuning for transformer-based masked language-models.
\newblock In \emph{Proceedings of the 60th Annual Meeting of the Association for Computational Linguistics (Volume 2: Short Papers)}, 2022.

\bibitem[Hu et~al.(2022)Hu, yelong shen, Wallis, Allen-Zhu, Li, Wang, Wang, and Chen]{hulora}
Edward~J Hu, yelong shen, Phillip Wallis, Zeyuan Allen-Zhu, Yuanzhi Li, Shean Wang, Lu~Wang, and Weizhu Chen.
\newblock Lo{RA}: Low-rank adaptation of large language models.
\newblock In \emph{International Conference on Learning Representations}, 2022.
\newblock URL \url{https://openreview.net/forum?id=nZeVKeeFYf9}.

\bibitem[Turner et~al.(2023)Turner, Thiergart, Udell, Leech, Mini, and MacDiarmid]{turner2023activation}
Alex Turner, Lisa Thiergart, David Udell, Gavin Leech, Ulisse Mini, and Monte MacDiarmid.
\newblock Activation addition: Steering language models without optimization.
\newblock \emph{arXiv preprint arXiv:2308.10248}, 2023.

\bibitem[Suau et~al.(2024)Suau, Delobelle, Metcalf, Joulin, Apostoloff, Zappella, and Rodriguez]{suau2024whispering}
Xavier Suau, Pieter Delobelle, Katherine Metcalf, Armand Joulin, Nicholas Apostoloff, Luca Zappella, and Pau Rodriguez.
\newblock Whispering experts: Neural interventions for toxicity mitigation in language models.
\newblock In \emph{Forty-first International Conference on Machine Learning}, 2024.
\newblock URL \url{https://openreview.net/forum?id=2P6GVfSrfZ}.

\bibitem[Genevay et~al.(2018)Genevay, Peyr{\'e}, and Cuturi]{genevay2018learning}
Aude Genevay, Gabriel Peyr{\'e}, and Marco Cuturi.
\newblock Learning generative models with sinkhorn divergences.
\newblock In \emph{International Conference on Artificial Intelligence and Statistics}, pages 1608--1617. PMLR, 2018.

\bibitem[Santambrogio(2015)]{santambrogio2015optimal}
Filippo Santambrogio.
\newblock Optimal transport for applied mathematicians.
\newblock \emph{Birk{\"a}user, NY}, 55\penalty0 (58-63):\penalty0 94, 2015.

\bibitem[Rabin et~al.(2012)Rabin, Peyr{\'e}, Delon, and Bernot]{rabin2012wasserstein}
Julien Rabin, Gabriel Peyr{\'e}, Julie Delon, and Marc Bernot.
\newblock Wasserstein barycenter and its application to texture mixing.
\newblock In \emph{Scale Space and Variational Methods in Computer Vision: Third International Conference, SSVM 2011, Ein-Gedi, Israel, May 29--June 2, 2011, Revised Selected Papers 3}, pages 435--446. Springer, 2012.

\bibitem[Yoon and Hwang(2017)]{pmlr-v70-yoon17a}
Jaehong Yoon and Sung~Ju Hwang.
\newblock Combined group and exclusive sparsity for deep neural networks.
\newblock In Doina Precup and Yee~Whye Teh, editors, \emph{Proceedings of the 34th International Conference on Machine Learning}, volume~70 of \emph{Proceedings of Machine Learning Research}, pages 3958--3966. PMLR, 06--11 Aug 2017.
\newblock URL \url{https://proceedings.mlr.press/v70/yoon17a.html}.

\bibitem[Adams et~al.(2017)Adams, Sorensen, Elliott, Dixon, McDonald, nithum, and Cukierski]{jigsaw-toxic-comment-classification-challenge}
CJ~Adams, Jeffrey Sorensen, Julia Elliott, Lucas Dixon, Mark McDonald, nithum, and Will Cukierski.
\newblock Toxic comment classification challenge, 2017.
\newblock URL \url{https://kaggle.com}.

\bibitem[Gehman et~al.(2020)Gehman, Gururangan, Sap, Choi, and Smith]{gehman2020realtoxicityprompts}
Samuel Gehman, Suchin Gururangan, Maarten Sap, Yejin Choi, and Noah~A Smith.
\newblock Realtoxicityprompts: Evaluating neural toxic degeneration in language models.
\newblock \emph{arXiv preprint arXiv:2009.11462}, 2020.

\bibitem[Luong et~al.(2024)Luong, Le, Van, and Nguyen]{tet_dataset}
Tinh~Son Luong, Thanh-Thien Le, Linh~Ngo Van, and Thien~Huu Nguyen.
\newblock Realistic evaluation of toxicity in large language models, 2024.

\bibitem[Logacheva et~al.(2022)Logacheva, Dementieva, Ustyantsev, Moskovskiy, Dale, Krotova, Semenov, and Panchenko]{logacheva-etal-2022-paradetox}
Varvara Logacheva, Daryna Dementieva, Sergey Ustyantsev, Daniil Moskovskiy, David Dale, Irina Krotova, Nikita Semenov, and Alexander Panchenko.
\newblock {P}ara{D}etox: Detoxification with parallel data.
\newblock In \emph{Proceedings of the 60th Annual Meeting of the Association for Computational Linguistics (Volume 1: Long Papers)}, pages 6804--6818, Dublin, Ireland, May 2022. Association for Computational Linguistics.
\newblock URL \url{https://aclanthology.org/2022.acl-long.469}.

\bibitem[Wikimedia()]{wikidump}
Foundation Wikimedia.
\newblock Wikimedia downloads.
\newblock URL \url{https://dumps.wikimedia.org}.

\bibitem[Hendrycks et~al.(2021)Hendrycks, Burns, Basart, Zou, Mazeika, Song, and Steinhardt]{hendryckstest2021}
Dan Hendrycks, Collin Burns, Steven Basart, Andy Zou, Mantas Mazeika, Dawn Song, and Jacob Steinhardt.
\newblock Measuring massive multitask language understanding.
\newblock \emph{Proceedings of the International Conference on Learning Representations (ICLR)}, 2021.

\bibitem[Bruni et~al.(2014)Bruni, Tran, and Baroni]{men_dataset}
Elia Bruni, Nam~Khanh Tran, and Marco Baroni.
\newblock Multimodal distributional semantics.
\newblock \emph{J. Artif. Int. Res.}, 49\penalty0 (1):\penalty0 1–47, January 2014.
\newblock ISSN 1076-9757.

\bibitem[Yin et~al.(2024{\natexlab{b}})Yin, Gharbi, Park, Zhang, Shechtman, Durand, and Freeman]{yin2024dmd2}
Tianwei Yin, Micha{\"e}l Gharbi, Taesung Park, Richard Zhang, Eli Shechtman, Fredo Durand, and William~T Freeman.
\newblock Improved distribution matching distillation for fast image synthesis.
\newblock In \emph{NeurIPS}, 2024{\natexlab{b}}.

\bibitem[Podell et~al.()Podell, English, Lacey, Blattmann, Dockhorn, M{\"u}ller, Penna, and Rombach]{podellsdxl}
Dustin Podell, Zion English, Kyle Lacey, Andreas Blattmann, Tim Dockhorn, Jonas M{\"u}ller, Joe Penna, and Robin Rombach.
\newblock Sdxl: Improving latent diffusion models for high-resolution image synthesis.
\newblock In \emph{The Twelfth International Conference on Learning Representations}.

\bibitem[Goodfellow et~al.(2014)Goodfellow, Pouget-Abadie, Mirza, Xu, Warde-Farley, Ozair, Courville, and Bengio]{goodfellow2014generative}
Ian~J Goodfellow, Jean Pouget-Abadie, Mehdi Mirza, Bing Xu, David Warde-Farley, Sherjil Ozair, Aaron Courville, and Yoshua Bengio.
\newblock Generative adversarial nets.
\newblock \emph{Advances in neural information processing systems}, 27, 2014.

\bibitem[Ye et~al.(2024)Ye, Lin, Han, Xu, Liu, Liang, Ma, Zou, and Ermon]{Ye2024-TFG}
Haotian Ye, Haowei Lin, Jiaqi Han, Minkai Xu, Sheng Liu, Yitao Liang, Jianzhu Ma, James Zou, and Stefano Ermon.
\newblock Tfg: Unified training-free guidance for diffusion models.
\newblock In \emph{Advances in Neural Information Processing Systems}, volume~37, 2024.

\bibitem[Guo et~al.(2024)Guo, Yuan, Yang, and Chen]{Guo2024-GGDM}
Yingqing Guo, Hui Yuan, Yukang Yang, and Mengdi Chen, Minshuo an d~Wang.
\newblock Gradient guidance for diffusion models: An optimization perspective.
\newblock In \emph{Advances in Neural Information Processing Systems}, volume~37, 2024.

\bibitem[Singhal et~al.(2025)Singhal, Horvitz, Teehan, Ren, Yu, McKeown, and Ranganath]{Singhal2025-FK}
Raghav Singhal, Zachary Horvitz, Ryan Teehan, Mengye Ren, Zhou Yu, Kathleen McKeown, and Rajesh Ranganath.
\newblock A general framework for inference-time scaling and steering of diffusion models, 2025.

\bibitem[Hessel et~al.(2021)Hessel, Holtzman, Forbes, Le~Bras, and Choi]{hessel2021clipscore}
Jack Hessel, Ari Holtzman, Maxwell Forbes, Ronan Le~Bras, and Yejin Choi.
\newblock Clipscore: A reference-free evaluation metric for image captioning.
\newblock In \emph{Proceedings of the 2021 Conference on Empirical Methods in Natural Language Processing}, pages 7514--7528, 2021.

\bibitem[Oquab et~al.(2024)Oquab, Darcet, Moutakanni, Vo, Szafraniec, Khalidov, Fernandez, HAZIZA, Massa, El-Nouby, Assran, Ballas, Galuba, Howes, Huang, Li, Misra, Rabbat, Sharma, Synnaeve, Xu, Jegou, Mairal, Labatut, Joulin, and Bojanowski]{oquab2024dinov}
Maxime Oquab, Timoth{\'e}e Darcet, Th{\'e}o Moutakanni, Huy~V. Vo, Marc Szafraniec, Vasil Khalidov, Pierre Fernandez, Daniel HAZIZA, Francisco Massa, Alaaeldin El-Nouby, Mido Assran, Nicolas Ballas, Wojciech Galuba, Russell Howes, Po-Yao Huang, Shang-Wen Li, Ishan Misra, Michael Rabbat, Vasu Sharma, Gabriel Synnaeve, Hu~Xu, Herve Jegou, Julien Mairal, Patrick Labatut, Armand Joulin, and Piotr Bojanowski.
\newblock {DINO}v2: Learning robust visual features without supervision.
\newblock \emph{Transactions on Machine Learning Research}, 2024.
\newblock URL \url{https://openreview.net/forum?id=a68SUt6zFt}.

\bibitem[Belloni and Chernozhukov(2013)]{belloni2013least}
Alexandre Belloni and Victor Chernozhukov.
\newblock Least squares after model selection in high-dimensional sparse models.
\newblock 2013.

\bibitem[Chzen et~al.(2019)Chzen, Hebiri, and Salmon]{chzhen2019lasso}
E.~Chzen, M.~Hebiri, and J.~Salmon.
\newblock On lasso refitting strategies.
\newblock \emph{Bernoulli}, 25\penalty0 (4A):\penalty0 3175--3200, 2019.

\bibitem[Veit et~al.(2016)Veit, Wilber, and Belongie]{veit2016residual}
Andreas Veit, Michael~J Wilber, and Serge Belongie.
\newblock Residual networks behave like ensembles of relatively shallow networks.
\newblock \emph{Advances in neural information processing systems}, 29, 2016.

\bibitem[Jastrzebski et~al.(2018)Jastrzebski, Arpit, Ballas, Verma, Che, and Bengio]{jastrzebski2018residual}
Stanis{\l}aw Jastrzebski, Devansh Arpit, Nicolas Ballas, Vikas Verma, Tong Che, and Yoshua Bengio.
\newblock Residual connections encourage iterative inference.
\newblock In \emph{International Conference on Learning Representations}, 2018.

\bibitem[Schulman et~al.(2017)Schulman, Wolski, Dhariwal, Radford, and Klimov]{schulman2017proximal}
John Schulman, Filip Wolski, Prafulla Dhariwal, Alec Radford, and Oleg Klimov.
\newblock Proximal policy optimization algorithms.
\newblock \emph{arXiv preprint arXiv:1707.06347}, 2017.

\bibitem[Fedzechkina et~al.(2025)Fedzechkina, Gualdoni, Williamson, Metcalf, Seto, and Theobald]{fedzechkina2025analyze}
Masha Fedzechkina, Eleonora Gualdoni, Sinead Williamson, Katherine Metcalf, Skyler Seto, and Barry-John Theobald.
\newblock Analyze the neurons, not the embeddings: Understanding when and where llm representations align with humans, 2025.

\bibitem[Lin et~al.(2021)Lin, Hilton, and Evans]{lin2021truthfulqa}
Stephanie Lin, Jacob Hilton, and Owain Evans.
\newblock Truthfulqa: Measuring how models mimic human falsehoods.
\newblock \emph{arXiv preprint arXiv:2109.07958}, 2021.

\end{thebibliography}

\appendix
\FloatBarrier
\section{Broader Impact}
\label{app:impact}

This paper presents an algorithm that aims to advance the field of Machine Learning without a specific application in mind. 

Our objective has been to use our algorithm to condition models towards desired behaviors (\eg being less toxic), however, a malicious user with access to the model's activations could condition the model to behave in a negative way, \eg forcing the model to be more toxic.

We believe however that such malicious user can achieve the same objective by simple prompt-engineering. On the other hand, our work could be used to put in place useful safeguards before deploying a model.

\FloatBarrier
\section{Hardware and Compute Requirements}
\label{app:hardware}
The experiments in this work were computed on a single NVIDIA A100 GPU with 80GB RAM and they could also fit in an NVIDIA A100 with 40GB RAM.

\paragraph{Memory Consumption}
During training, \method leverages backpropagation to compute gradients for its diagonal affine interventions. This design ensures that only activations relevant to the intervened layers require storage during the forward pass, leading to a substantially reduced memory footprint compared to full parameter tuning.

\paragraph{Compute}
Compared to local methods, the computational cost of \method is primarily determined by the number of optimization steps required for intervention training. While this can result in slower estimation times than some local approaches, \method is notably an order of magnitude faster than the RL baseline. Furthermore, we demonstrate that \method can achieve competitive results with significantly reduced computational resources, specifically using 10x fewer optimization steps (e.g., 100 steps, approximately 50s for estimation) than the full configuration presented in the paper (Figure 4). This contrasts favorably with Lin-AcT's \textasciitilde30s and dramatically outperforms LoFIT-RL's 27300s (Table \ref{tab:estimation_times}). Overall, \method offers a positive trade-off between computational expenditure and conditioning performance.

\paragraph{Timing} The table below summarizes the estimation time for each method and the number of steps used in our submission.

\begin{table}[htbp]
    \centering
    \caption{Estimation times for various methods and models.}
    \label{tab:estimation_times}
    \begin{tabular}{l c c c c}
        \toprule
        \textbf{Method} & \textbf{\# Steps} & \textbf{Gemma2 (s)} & \textbf{Qwen1.5B (s)} & \textbf{Qwen7B (s)} \\
        \midrule
        LoFIT-RL & 100 & 7600 & 25900 & 27300 \\
        ReFT & 10 & 92 & 80 & 100 \\
        ITI & 1 & 29 & 30 & 37 \\
        Lin-AcT & 1 & 17 & 14 & 22 \\
        \method & 1000 & 430 & 340 & 500 \\
        \bottomrule
    \end{tabular}
\end{table}

We have also timed DMD2 for T2I generation. The table below contains the total training time of ITI, Lin-AcT, and \method on all normalization layers of DMD2's UNet. It is interesting to see that when conditioning many layers, the difference in training time between \method and other methods shrinks. This is because \method leverages PyTorch's backpropagation, which is optimized compared to ITI and Lin-Act's layer-wise estimation methods.

\begin{table}[htbp]
    \centering
    \caption{Total training time for methods on DMD2's UNet.}
    \label{tab:training_times_dmd2}
    \begin{tabular}{l c c}
        \toprule
        \textbf{Method} & \textbf{\# steps} & \textbf{DMD2} \\
        \midrule
        ITI & 1 & 26m 44s \\
        LinAcT & 1 & 25m 53s \\
        \method & 1000 & 29m 27s \\
        \bottomrule
    \end{tabular}
\end{table}

\subsection{Detailed Complexity Analysis}

\noindent
\textbf{Variables:}
\begin{itemize}
    \item $B$: batch size
    \item $B \log B$: sorting cost for optimal transport
    \item $F$: cost of forward (backward) pass
    \item $N$: Number of SGD steps
    \item $T$: Number of logistic regression L-BFGS steps
    \item $I$: Number of intervened layers
\end{itemize}

\noindent
\textbf{Computational Costs:}
\begin{itemize}
    \item The computational cost of \method is dominated by $O(N \cdot (2F + I \cdot B \log B))$
    \item The computational cost of ITI is $O(F + I \cdot T)$ where $T$ is the number of logistic regression steps
    \item The computational cost of Lin-AcT is $O(2F + I \cdot B \log B)$
\end{itemize}

\noindent
\textbf{Memory required during training:}
\begin{itemize}
    \item $M$: Memory required by the model activations during the forward pass
    \item $L$: Memory required to compute the forward pass on one layer
    \item $D$: Affine parameter weight matrix size
\end{itemize}

\noindent
\textbf{Memory Requirements per Method:}
\begin{itemize}
    \item \method: $M + ID$
    \item LinAcT: $L + D$
    \item ITI: $L + D$
\end{itemize}

\FloatBarrier
\section{User Study on LMs}
\label{app:user-study}
We conducted a user preference study to evaluate the perceived quality of continuations generated by different intervention methods. The study involved a pool of 20 volunteers. Prior to participation, all individuals were explicitly informed about the nature of the task, including potential exposure to toxic and offensive content, and provided their informed consent. Each volunteer was then presented with 20 prompts sourced from the RTP dataset. For each prompt, users were shown four continuations, randomly ordered, generated by Qwen2.5-7B using the following methods: no intervention (identity), \iti, \linearact, and \method. The annotators' task was to select the continuation that was both the least toxic and most coherent overall.

The aggregated results of the preference study are summarized in the table below, showing the percentage of times each method was preferred, along with its standard deviation. Our findings indicate a strong preference for LinEAS over three other alternatives: Linear-ACT, ITI-C, and no intervention (identity). Specifically, users preferred LinEAS in 57.70\% of cases.

\begin{table}[h!]
    \centering
    \caption{Human Preference Study Results}
    \label{tab:preference_results}
    \begin{tabular}{l c c c c}
        \toprule
        Method & Identity (\%) & ITI (\%) & Lin-AcT (\%) & LinEAS (\%) \\
        \midrule
        Preference & $12.19 \pm 5.5$ & $11.67 \pm 8.7$ & $18.43 \pm 11.49$ & $57.70 \pm 14.64$ \\
        \bottomrule
    \end{tabular}
\end{table}

\FloatBarrier
\section{Optimization details for \method}
\label{app:optimization}

See Algo. \ref{algo:e2ealgo} for the description of a single \method optimization step. 
Recall that for a vector $z\in\mathbb{R}^d$, the proximal operators of the 1-norm (a.k.a. soft-thresholding) and 2-norm are given by:
\begin{equation}
\begin{aligned}
    \ST_\tau(z) & := \sign(z) \odot \max(|z|-\tau,\mathbf{0}) \qquad \text{and} \qquad
    \Prox_{\tau\|\cdot\|_2}(z) & :=\left(1 - \frac{\tau}{\|z\|_2}\right)_+ z\,. \label{eq:GLasso}
\end{aligned}
\end{equation}

\begin{algorithm}[h]
\caption{Proximal E2E Training Step.}
\label{algo:e2ealgo}
\begin{algorithmic}[1]
\State \textbf{Require:} prompts $(\vx^i)_i\sim p$, $(\vy^j)_j \sim q$, LR $\rho$.
\State (pre-) compute activations $\veta^i_\ell, i\leq n, \ell\leq L$ \Comment{Eq.(\ref{eq:vy})}
\State compute activations lists $\vxi^i_\ell, i\leq n, \ell\leq L$. \Comment{Eq.(\ref{eq:vx})}
\State set loss to $\mathcal{C}=0$
\For{$\ell\leq L$}\Comment{Forward}
    \State \rebuttal{$Z :=  [\vxi^1_\ell, \dots, \vxi^n_\ell] \in \mathbb{R}^{n\times d_\ell}$ \Comment{Eq. (\ref{eq:vx})}}
    \State \rebuttal{$V :=[\veta^1_\ell, \dots, \veta^n_\ell] \in \mathbb{R}^{n\times d_\ell}$ \Comment{Eq. (\ref{eq:vy})}}
    \State $\mathcal{C} \leftarrow \mathcal{C} + \Delta(Z, V)$ \Comment{$\ell$-layer loss, Eq. (\ref{eq:sw})}
\EndFor
\For{$\ell\leq L$}%
    \State $g_\omega, g_{b}\leftarrow \nabla_{\ROmega{\ell},\Rb{\ell}} \mathcal{C}$ \Comment{Backpropagation}
    \State $\ROmega{\ell},\Rb{\ell} \leftarrow \ROmega{\ell} - \rho\, g_\omega, \Rb{\ell} - \rho\,g_{b}$ \Comment{Updates}
    \State $\ROmega{\ell}\leftarrow \Prox_{\reg\lambda_G\|\cdot\|_2}\!\circ \ST_{\reg\lambda_1}(\ROmega{\ell} -\mathbf{1}) + \!\mathbf{1}$ \Comment{Eq. (\ref{eq:GLasso})}
    \State $\Rb{\ell}\leftarrow \Prox_{\reg\lambda_G\|\cdot\|_2}\!\circ \ST_{\reg\lambda_1}(\Rb{\ell})$ \Comment{Eq. (\ref{eq:GLasso})}
\EndFor
\end{algorithmic}
\end{algorithm}

\section{Toxicity Mitigation (extended results)}
\label{app:tox}

\Cref{tab:toxicity_full} is an extension of \cref{tab:toxicity_main} in which we include one more model, \deepseeksevenb, and more baselines. Specifically we include here another oracle baseline, a LoRA adapter, that we named \lorartc, and that is trained similarly to \lofitrl as explained in the main paper. More details on the training of these strongly supervised baselines can be found in \cref{app:rlhf}.

Additionally we include here \caa as proposed in the original work (\ie intervening only on the middle layer rather than on all layers as shown in the main paper).

\definecolor{Gray}{gray}{0.95}
\begin{table}[tb]
    \centering
    \resizebox{0.8\columnwidth}{!}{
\begin{tabular}{lllrrrr}
\toprule
Model & Method & \#Params & \toxclsrtp $(\downarrow)$ & \toxclstet $(\downarrow)$ & \pplwik $(\downarrow)$ & MMLU $(\uparrow)$ \\

\arrayrulecolor{black} \midrule
\rowcolor{Gray} \multirow{9}{*}{\cellcolor{white}Q1.5B} &  None & - &  $3.00_{\color{gray}{\,0.54}}$ & $23.09_{\color{gray}{\,0.67}}$ & $13.67_{\color{gray}{\,0.00}}$ & $60.95_{\color{gray}{\,0.00}}$ \\
\rowcolor{Gray}\cellcolor{white} &   \lorartc & 0.54M & $1.07_{\color{gray}{\,0.40}}$ & $7.94_{\color{gray}{\,0.01}}$ & $13.70_{\color{gray}{\,0.10}}$ & $60.78_{\color{gray}{\,0.17}}$ \\
\rowcolor{Gray}\cellcolor{white} & LoFIT-RL & 0.8596M & $0.37_{\color{gray}{\,0.06}}$ & $4.36_{\color{gray}{\,0.00}}$ & $14.12_{\color{gray}{\,0.07}}$ & $59.74_{\color{gray}{\,0.14}}$ \\
\arrayrulecolor{gray!50} \cmidrule{2-7}

  &  \prompt & - & $4.07_{\color{gray}{\,0.38}}$ & $21.02_{\color{gray}{\,1.44}}$ & $13.65_{\color{gray}{\,0.00}}$ & $60.96_{\color{gray}{\,0.00}}$ \\
  &  \caa (mid) & 0.0015M &  $2.86_{\color{gray}{\,0.53}}$ & $23.33_{\color{gray}{\,1.25}}$ & $13.69_{\color{gray}{\,0.02}}$ & $60.47_{\color{gray}{\,0.18}}$ \\
  &  \caa & 0.043M &  $1.15_{\color{gray}{\,0.37}}$ & $5.77_{\color{gray}{\,2.14}}$ & $\red{19.30}_{\color{gray}{\,2.76}}$ & $\red{37.67}_{\color{gray}{\,6.95}}$ \\
 &  \reft & 0.39M &  $2.57_{\color{gray}{\,0.60}}$ & $18.17_{\color{gray}{\,3.04}}$ & $15.58_{\color{gray}{\,0.52}}$ & $58.84_{\color{gray}{\,0.23}}$ \\

 &  \iti & 0.043M &  $1.87_{\color{gray}{\,0.21}}$ & $18.16_{\color{gray}{\,0.62}}$ & $12.39_{\color{gray}{\,0.09}}$ & $60.88_{\color{gray}{\,0.08}}$ \\
 &  \linearact & 0.086M &  $1.50_{\color{gray}{\,0.35}}$ & $13.88_{\color{gray}{\,1.72}}$ & $13.88_{\color{gray}{\,0.16}}$ & $60.09_{\color{gray}{\,0.25}}$ \\
 &  \method & 0.086M &  $\textbf{1.07}_{\color{gray}{\,0.46}}$ & $\textbf{12.70}_{\color{gray}{\,0.74}}$ & $14.10_{\color{gray}{\,0.07}}$ & $59.97_{\color{gray}{\,0.16}}$ \\
\arrayrulecolor{black} \midrule

\rowcolor{Gray} \multirow{9}{*}{\cellcolor{white}G2-2B} & None & - &  $4.00_{\color{gray}{\,0.45}}$ & $13.39_{\color{gray}{\,1.42}}$ & $14.79_{\color{gray}{\,0.00}}$ & $53.03_{\color{gray}{\,0.00}}$ \\
\rowcolor{Gray}\cellcolor{white} & \lorartc & 0.8M & $0.83_{\color{gray}{\,0.25}}$ & $3.47_{\color{gray}{\,0.01}}$ & $15.38_{\color{gray}{\,0.17}}$ & $52.56_{\color{gray}{\,0.11}}$ \\
\rowcolor{Gray}\cellcolor{white} & LoFIT-RL & 0.1065M & $0.40_{\color{gray}{\,0.20}}$ & $1.76_{\color{gray}{\,0.00}}$ & $15.43_{\color{gray}{\,0.08}}$ & $52.17_{\color{gray}{\,0.17}}$ \\
\arrayrulecolor{gray!50} \cmidrule{2-7}
 
  &  \prompt & - &  $4.60_{\color{gray}{\,0.36}}$ & $12.32_{\color{gray}{\,0.67}}$ & $14.81_{\color{gray}{\,0.00}}$ & $53.18_{\color{gray}{\,0.00}}$ \\
  &  \caa (mid) & 0.0023M &  $4.93_{\color{gray}{\,0.42}}$ & $14.04_{\color{gray}{\,0.52}}$ & $14.88_{\color{gray}{\,0.02}}$ & $51.49_{\color{gray}{\,0.51}}$ \\
  &  \caa & 0.06M &  $0.80_{\color{gray}{\,0.00}}$ & $2.44_{\color{gray}{\,1.99}}$ & $\red{23.52}_{\color{gray}{\,2.67}}$ & $\red{26.86}_{\color{gray}{\,0.08}}$ \\
 &  \reft & 0.54M &  $2.85_{\color{gray}{\,0.49}}$ & $11.15_{\color{gray}{\,1.91}}$ & $\red{19.93}_{\color{gray}{\,0.30}}$ & $48.99_{\color{gray}{\,1.34}}$ \\

&  \iti & 0.06M &  $1.17_{\color{gray}{\,0.60}}$ & $7.15_{\color{gray}{\,0.92}}$ & $14.00_{\color{gray}{\,0.11}}$ & $52.78_{\color{gray}{\,0.23}}$ \\
&  \linearact & 0.12M &  $1.60_{\color{gray}{\,0.32}}$ & $7.76_{\color{gray}{\,0.39}}$ & $14.78_{\color{gray}{\,0.12}}$ & $52.43_{\color{gray}{\,0.57}}$ \\
&  \method & 0.24M &  $\textbf{0.73}_{\color{gray}{\,0.10}}$ & $\textbf{4.02}_{\color{gray}{\,0.68}}$ & $15.46_{\color{gray}{\,0.21}}$ & $52.22_{\color{gray}{\,0.40}}$ \\
 
 \arrayrulecolor{black} \midrule
\rowcolor{Gray} \multirow{9}{*}{\cellcolor{white}D7B} & None & - &  $4.30_{\color{gray}{\,0.70}}$ & $18.62_{\color{gray}{\,0.51}}$ & $8.49_{\color{gray}{\,0.00}}$ & $48.31_{\color{gray}{\,0.00}}$ \\
\rowcolor{Gray}\cellcolor{white} & \lorartc & 1.97M & $1.97_{\color{gray}{\,0.38}}$ & $5.07_{\color{gray}{\,0.00}}$ & $8.67_{\color{gray}{\,0.05}}$ & $47.76_{\color{gray}{\,0.34}}$ \\
\rowcolor{Gray}\cellcolor{white} & LoFIT-RL & 0.2458 & $0.53_{\color{gray}{\,0.15}}$ & $1.63_{\color{gray}{\,0.00}}$ & $9.31_{\color{gray}{\,0.05}}$ & $46.78_{\color{gray}{\,0.14}}$ \\
\arrayrulecolor{gray!50} \cmidrule{2-7}

  &  \prompt & - &  $4.20_{\color{gray}{\,0.70}}$ & $15.69_{\color{gray}{\,0.82}}$ & $8.51_{\color{gray}{\,0.00}}$ & $48.23_{\color{gray}{\,0.00}}$ \\
  &  \caa (mid) & 0.0043M &  $4.72_{\color{gray}{\,0.54}}$ & $19.07_{\color{gray}{\,0.98}}$ & $8.73_{\color{gray}{\,0.17}}$ & $44.76_{\color{gray}{\,1.98}}$ \\
  &  \caa & 0.13M &  $0.07_{\color{gray}{\,0.15}}$ & $0.33_{\color{gray}{\,0.65}}$ & \red{$>1000$} & $\red{23.14}_{\color{gray}{\,0.34}}$ \\
 &  \reft & 1.11M &  $2.25_{\color{gray}{\,1.04}}$ & $10.39_{\color{gray}{\,6.01}}$ & $\red{51.58}_{\color{gray}{\,40.2}}$ & $\red{35.56}_{\color{gray}{\,11.1}}$ \\

 &  \iti & 0.13M &  $2.83_{\color{gray}{\,0.40}}$ & $15.18_{\color{gray}{\,2.00}}$ & $7.71_{\color{gray}{\,0.07}}$ & $48.47_{\color{gray}{\,0.25}}$ \\
 &  \linearact & 0.25M &  $\textbf{2.23}_{\color{gray}{\,0.69}}$ & $\textbf{11.08}_{\color{gray}{\,0.76}}$ & $8.67_{\color{gray}{\,0.03}}$ & $47.71_{\color{gray}{\,0.27}}$ \\
 &  \method & 0.25M &  $2.30_{\color{gray}{\,0.14}}$ & $12.09_{\color{gray}{\,0.83}}$ & $8.38_{\color{gray}{\,0.05}}$ & $48.13_{\color{gray}{\,0.07}}$ \\ 
 
\arrayrulecolor{black} \midrule
\rowcolor{Gray} \multirow{9}{*}{\cellcolor{white}Q7B} & None & - &  $3.92_{\color{gray}{\,0.59}}$ & $25.16_{\color{gray}{\,0.92}}$ & $10.67_{\color{gray}{\,0.00}}$ & $74.26_{\color{gray}{\,0.00}}$ \\
\rowcolor{Gray}\cellcolor{white} & \lorartc & 1.26M & $1.30_{\color{gray}{\,0.44}}$ & $6.59_{\color{gray}{\,0.01}}$ & $10.68_{\color{gray}{\,0.06}}$ & $74.08_{\color{gray}{\,0.15}}$ \\
\rowcolor{Gray}\cellcolor{white} & LoFIT-RL & 0.10M & $1.10_{\color{gray}{\,0.38}}$ & $7.11_{\color{gray}{\,0.30}}$ & $10.91_{\color{gray}{\,0.16}}$ & $73.87_{\color{gray}{\,0.17}}$ \\
\arrayrulecolor{gray!50} \cmidrule{2-7}

  &  \prompt & - &  $6.80_{\color{gray}{\,0.00}}$ & $21.22_{\color{gray}{\,0.21}}$ & $10.65_{\color{gray}{\,0.00}}$ & $74.23_{\color{gray}{\,0.00}}$ \\
  &  \caa (mid) & 0.0036M & $4.00_{\color{gray}{\,0.45}}$ & $22.32_{\color{gray}{\,1.13}}$ & $10.66_{\color{gray}{\,0.03}}$ & $73.45_{\color{gray}{\,0.14}}$ \\
  &  \caa & 0.10M &  $1.20_{\color{gray}{\,0.00}}$ & $9.25_{\color{gray}{\,3.07}}$ & $12.83_{\color{gray}{\,0.00}}$ & $\red{48.58}_{\color{gray}{\,0.00}}$ \\
 &  \reft & 0.90M &  $3.33_{\color{gray}{\,0.96}}$ & $20.38_{\color{gray}{\,2.37}}$ & $13.80_{\color{gray}{\,1.20}}$ & $70.43_{\color{gray}{\,0.60}}$ \\

 &  \iti & 0.10M &  $2.63_{\color{gray}{\,0.44}}$ & $19.98_{\color{gray}{\,1.24}}$ & $9.63_{\color{gray}{\,0.03}}$ & $74.08_{\color{gray}{\,0.05}}$ \\
 &  \linearact & 0.20M &  $2.72_{\color{gray}{\,0.46}}$ & $21.64_{\color{gray}{\,2.00}}$ & $11.42_{\color{gray}{\,0.34}}$ & $72.18_{\color{gray}{\,0.16}}$ \\
 &  \method & 0.20M &  $\textbf{1.95}_{\color{gray}{\,0.48}}$ & $\textbf{14.95}_{\color{gray}{\,0.92}}$ & $10.91_{\color{gray}{\,0.35}}$ & $73.67_{\color{gray}{\,0.05}}$ \\
\arrayrulecolor{black}\bottomrule
\end{tabular}}
\vskip 1mm
\caption{Toxicity mitigation on the RTP and TET datasets using four different models, Q1.5B: \qwenonefiveb, G2-2B: \gemmatwob, D7B: \deepseeksevenb and Q7B: \qwensevenb. \rebuttal{Strongly degraded utility is marked in \red{red}.} We report results at low ($N=32$ sentences to estimate the interventions) data regime. For each method, we use the best intervention layers according to an ablation study. See Table \ref{tab:toxicity_large} in Appendix for an ablation with larger training size. Results for \method improve significantly on \iti and \linearact with similar impact on quality metrics. The quality of these interventions is often on par with the strong baselines \lorartc and \lofitrl, despite this approach being far more involved (both in terms of parameter size, access to ground truth labeling oracle \RTC, and compute).}
    \label{tab:toxicity_full}
\end{table}

In \Cref{tab:toxicity_large} we show results analogous to those in \Cref{tab:toxicity_full} but in a higher data regime, \ie using 1024 sentences per set. 

Additionally, in \Cref{tab:toxicity_layer}, we show how the different activation steering methods perform in the setting of \Cref{sec:tox}, when intervening different layer types (namely \texttt{.*post\_.*\_layernorm} and \texttt{.*o\_proj} of the models' Huggingface implementation). We report results for the low data regime, showing that \method is much more robust to the layer choice. Indeed, for \texttt{.*post\_.*\_layernorm} and models \deepseeksevenb and \qwenonefiveb, \iti and \linearact induce a toxicity slightly higher than the original one (marked in {\color{red}red}). 

\begin{table}[htb!]
    \centering
    \resizebox{0.85\columnwidth}{!}{%
\begin{tabular}{lllrrrr}
\toprule
Model & method & \# par (M) & \toxclsrtp $(\downarrow)$ & \toxclstet $(\downarrow)$ & \pplwik $(\downarrow)$ & MMLU $(\uparrow)$ \\
\arrayrulecolor{black} \midrule        
\rowcolor{Gray}\multirow{9}{*}{\cellcolor{white}\gemmatwob} &  None & - &  $4.00_{\color{gray}{\,0.45}}$ & $13.39_{\color{gray}{\,1.42}}$ & $14.79_{\color{gray}{\,0.00}}$ & $53.03_{\color{gray}{\,0.00}}$ \\
\rowcolor{Gray}\cellcolor{white} & \lorartc & 0.8M & $0.50_{\color{gray}{\,0.44}}$ & $1.68_{\color{gray}{\,0.01}}$ & $15.78_{\color{gray}{\,0.19}}$ & $52.45_{\color{gray}{\,0.54}}$ \\
\arrayrulecolor{gray!50} \cmidrule{2-7}

& \prompt & - & $4.60_{\color{gray}{\,0.36}}$ & $12.32_{\color{gray}{\,0.67}}$ & $14.81_{\color{gray}{\,0.00}}$ & $53.18_{\color{gray}{\,0.00}}$ \\
& \caa (mid) & 0.0023M & $4.23_{\color{gray}{\,0.72}}$ & $12.41_{\color{gray}{\,0.68}}$ & $14.83_{\color{gray}{\,0.00}}$ & $52.32_{\color{gray}{\,0.08}}$ \\
& \caa & 0.06M & $0.70_{\color{gray}{\,0.14}}$ & $3.17_{\color{gray}{\,0.74}}$ & $16.38_{\color{gray}{\,0.01}}$ & $\red{46.44}_{\color{gray}{\,0.02}}$ \\
& \reft & 0.54M & $3.73_{\color{gray}{\,0.95}}$ & $14.04_{\color{gray}{\,1.61}}$ & $15.40_{\color{gray}{\,0.18}}$ & $51.32_{\color{gray}{\,0.25}}$ \\

 &  \iti & 0.06 &  $0.30_{\color{gray}{\,0.26}}$ & $2.68_{\color{gray}{\,0.43}}$ & $14.65_{\color{gray}{\,0.06}}$ & $52.04_{\color{gray}{\,0.17}}$ \\
 &  \linearact & 0.12 &  $1.07_{\color{gray}{\,0.52}}$ & $6.08_{\color{gray}{\,0.67}}$ & $14.85_{\color{gray}{\,0.04}}$ & $52.36_{\color{gray}{\,0.11}}$ \\
 &  \method & 0.24 &  $0.95_{\color{gray}{\,0.26}}$ & $3.46_{\color{gray}{\,0.44}}$ & $15.82_{\color{gray}{\,0.02}}$ & $51.28_{\color{gray}{\,0.08}}$ \\

\arrayrulecolor{black} \midrule
\rowcolor{Gray}\multirow{9}{*}{\cellcolor{white}\deepseeksevenb} &  None & - &  $4.30_{\color{gray}{\,0.70}}$ & $18.62_{\color{gray}{\,0.51}}$ & $8.49_{\color{gray}{\,0.00}}$ & $48.31_{\color{gray}{\,0.00}}$ \\
\rowcolor{Gray}\cellcolor{white} & \lorartc & 1.97M & $0.90_{\color{gray}{\,0.30}}$ & $1.95_{\color{gray}{\,0.00}}$ & $8.80_{\color{gray}{\,0.03}}$ & $47.28_{\color{gray}{\,0.58}}$ \\
\arrayrulecolor{gray!50} \cmidrule{2-7}

& \prompt & - & $4.20_{\color{gray}{\,0.57}}$ & $15.88_{\color{gray}{\,0.90}}$ & $8.51_{\color{gray}{\,0.00}}$ & $48.23_{\color{gray}{\,0.00}}$ \\
& \caa (mid) & 0.0043M & $4.63_{\color{gray}{\,0.25}}$ & $21.38_{\color{gray}{\,1.29}}$ & $8.72_{\color{gray}{\,0.15}}$ & $46.33_{\color{gray}{\,0.48}}$ \\
& \caa & 0.13M & $0.10_{\color{gray}{\,0.14}}$ & $0.08_{\color{gray}{\,0.16}}$ & \red{$>1000$} & $\red{23.69}_{\color{gray}{\,0.81}}$ \\
& \reft & 1.11M & $4.83_{\color{gray}{\,0.78}}$ & $17.76_{\color{gray}{\,0.90}}$ & $\red{12.34}_{\color{gray}{\,0.65}}$ & $\red{33.22}_{\color{gray}{\,4.55}}$ \\

 &  \iti & 0.13 &  $1.77_{\color{gray}{\,0.40}}$ & $10.70_{\color{gray}{\,1.14}}$ & $7.79_{\color{gray}{\,0.02}}$ & $48.20_{\color{gray}{\,0.15}}$ \\
 &  \linearact & 0.25 &  $1.42_{\color{gray}{\,0.43}}$ & $9.39_{\color{gray}{\,0.34}}$ & $8.77_{\color{gray}{\,0.01}}$ & $47.74_{\color{gray}{\,0.17}}$ \\
 &  \method & 0.25 &  $1.70_{\color{gray}{\,0.14}}$ & $9.49_{\color{gray}{\,0.52}}$ & $8.53_{\color{gray}{\,0.04}}$ & $48.01_{\color{gray}{\,0.04}}$ \\
 
\arrayrulecolor{black} \midrule
\rowcolor{Gray}\multirow{9}{*}{\cellcolor{white}\qwenonefiveb} &  None & - &  $3.00_{\color{gray}{\,0.54}}$ & $23.09_{\color{gray}{\,0.67}}$ & $13.67_{\color{gray}{\,0.00}}$ & $60.95_{\color{gray}{\,0.00}}$ \\
\rowcolor{Gray}\cellcolor{white} & \lorartc & 0.54M & $0.67_{\color{gray}{\,0.40}}$ & $3.93_{\color{gray}{\,0.01}}$ & $13.91_{\color{gray}{\,0.09}}$ & $60.70_{\color{gray}{\,0.19}}$ \\
\arrayrulecolor{gray!50} \cmidrule{2-7}

& \prompt & - & $4.07_{\color{gray}{\,0.38}}$ & $21.02_{\color{gray}{\,1.44}}$ & $13.65_{\color{gray}{\,0.00}}$ & $60.96_{\color{gray}{\,0.00}}$ \\
& \caa (mid) & 0.0015M & $2.87_{\color{gray}{\,0.72}}$ & $23.68_{\color{gray}{\,1.04}}$ & $13.67_{\color{gray}{\,0.02}}$ & $60.70_{\color{gray}{\,0.10}}$ \\
& \caa & 0.043M & $0.90_{\color{gray}{\,0.24}}$ & $6.69_{\color{gray}{\,2.07}}$ & $15.18_{\color{gray}{\,1.02}}$ & $\red{53.35}_{\color{gray}{\,3.94}}$ \\
& \reft & 0.39M & $2.75_{\color{gray}{\,0.29}}$ & $14.33_{\color{gray}{\,3.39}}$ & $\red{35.48}_{\color{gray}{\,20.8}}$ & $\red{52.63}_{\color{gray}{\,3.75}}$ \\

 &  \iti & 0.043 &  $1.60_{\color{gray}{\,0.10}}$ & $15.50_{\color{gray}{\,0.81}}$ & $12.53_{\color{gray}{\,0.04}}$ & $60.73_{\color{gray}{\,0.21}}$ \\
 &  \linearact & 0.086 &  $0.95_{\color{gray}{\,0.38}}$ & $11.61_{\color{gray}{\,1.43}}$ & $14.06_{\color{gray}{\,0.03}}$ & $59.82_{\color{gray}{\,0.22}}$ \\
 &  \method & 0.086 &  $0.90_{\color{gray}{\,0.26}}$ & $12.56_{\color{gray}{\,0.70}}$ & $14.20_{\color{gray}{\,0.04}}$ & $59.21_{\color{gray}{\,0.16}}$ \\
 
\arrayrulecolor{black} \midrule
\rowcolor{Gray}\multirow{9}{*}{\cellcolor{white}\qwensevenb} &  None & - &  $3.92_{\color{gray}{\,0.59}}$ & $25.16_{\color{gray}{\,0.92}}$ & $10.67_{\color{gray}{\,0.00}}$ & $74.26_{\color{gray}{\,0.00}}$ \\
\rowcolor{Gray}\cellcolor{white} & \lorartc & 1.26M & $1.50_{\color{gray}{\,0.36}}$ & $5.28_{\color{gray}{\,0.00}}$ & $11.03_{\color{gray}{\,0.05}}$ & $73.91_{\color{gray}{\,0.10}}$ \\
\arrayrulecolor{gray!50} \cmidrule{2-7}

& \prompt & - & $6.40_{\color{gray}{\,0.40}}$ & $21.22_{\color{gray}{\,0.21}}$ & $10.65_{\color{gray}{\,0.00}}$ & $74.23_{\color{gray}{\,0.00}}$ \\
& \caa (mid) &  0.0036M & $3.88_{\color{gray}{\,0.21}}$ & $22.87_{\color{gray}{\,0.54}}$ & $10.64_{\color{gray}{\,0.00}}$ & $73.86_{\color{gray}{\,0.04}}$ \\
& \caa & 0.10M & $2.00_{\color{gray}{\,0.00}}$ & $11.24_{\color{gray}{\,0.88}}$ & $11.18_{\color{gray}{\,0.00}}$ & $68.59_{\color{gray}{\,0.00}}$ \\
& \reft & 0.90M & $3.65_{\color{gray}{\,1.32}}$ & $22.70_{\color{gray}{\,3.39}}$ & $\red{17.42}_{\color{gray}{\,3.21}}$ & $\red{60.35}_{\color{gray}{\,11.8}}$ \\

 &  \iti & 0.10 &  $2.33_{\color{gray}{\,0.76}}$ & $18.18_{\color{gray}{\,2.00}}$ & $9.66_{\color{gray}{\,0.04}}$ & $74.19_{\color{gray}{\,0.10}}$ \\
 &  \linearact & 0.20 &  $1.65_{\color{gray}{\,0.26}}$ & $13.60_{\color{gray}{\,0.99}}$ & $10.80_{\color{gray}{\,0.02}}$ & $73.60_{\color{gray}{\,0.07}}$ \\
 &  \method & 0.20 &  $1.52_{\color{gray}{\,0.33}}$ & $13.92_{\color{gray}{\,0.54}}$ & $10.89_{\color{gray}{\,0.14}}$ & $73.37_{\color{gray}{\,0.07}}$ \\ 
 \arrayrulecolor{black}\bottomrule
\end{tabular}
    }
    \vskip 1mm
    \caption{Toxicity mitigation on the RTP and TET datasets. We report results at high ($N=1024$ sentences to estimate the interventions) data regime. We used 10k optimization steps, with minibatches of size $n=32$. In the high-data regime, \linearact achieves an outstanding 0.68 \toxclsrtp for \gemmatwob. However, this method struggles at reducing toxicity for the other models. Conversely, \method achieves similar (\gemmatwob) or better (other models) RTP toxicity mitigation than in the low data setup, and with better MMLU than \linearact for all models. Similary, \method outperforms all other methods on TET toxicity mitigation by a large margin.
}
    \label{tab:toxicity_large}
\end{table}

\begin{table}[htb!]
    \centering
    \resizebox{1.0\columnwidth}{!}{%
\begin{tabular}{llllrrrrr}
\toprule
Model & Method & Layer & Data & $\lambda$ & \toxclsrtp $(\downarrow)$ & \toxclstet $(\downarrow)$ & \pplwik $(\downarrow)$ & MMLU $(\uparrow)$ \\
\midrule
\multirow{4}{*}{\gemmatwob} &  None & \texttt{.*post\_.*\_layernorm} & - & - & $4.00_{\color{gray}{\,0.45}}$ & $13.39_{\color{gray}{\,1.42}}$ & $14.79_{\color{gray}{\,0.00}}$ & $53.03_{\color{gray}{\,0.00}}$ \\
 &  \iti & \texttt{.*post\_.*\_layernorm} & 32 & 1.0 & $2.38_{\color{gray}{\,0.91}}$ & $10.00_{\color{gray}{\,0.57}}$ & $13.89_{\color{gray}{\,0.13}}$ & $52.92_{\color{gray}{\,0.14}}$ \\
 &  \linearact & \texttt{.*post\_.*\_layernorm} & 32 & 1.0 & $1.35_{\color{gray}{\,0.17}}$ & $7.32_{\color{gray}{\,1.16}}$ & $15.08_{\color{gray}{\,0.13}}$ & $51.52_{\color{gray}{\,0.32}}$ \\
 &  \method & \texttt{.*post\_.*\_layernorm} & 32 & 1.0 & $\textbf{0.73}_{\color{gray}{\,0.10}}$ & $\textbf{4.02}_{\color{gray}{\,0.68}}$ & $15.46_{\color{gray}{\,0.21}}$ & $52.22_{\color{gray}{\,0.40}}$ \\

\midrule
\multirow{4}{*}{\deepseeksevenb} &  None & \texttt{.*post\_.*\_layernorm} & - & - & $4.30_{\color{gray}{\,0.70}}$ & $18.62_{\color{gray}{\,0.51}}$ & $8.49_{\color{gray}{\,0.00}}$ & $48.31_{\color{gray}{\,0.00}}$ \\
 &  \iti & \texttt{.*post\_.*\_layernorm} & 32 & 1.0 & $\color{red}{7.23}_{\color{gray}{\,0.76}}$ & $\color{red}{28.13}_{\color{gray}{\,2.45}}$ & $8.85_{\color{gray}{\,0.40}}$ & $45.82_{\color{gray}{\,1.15}}$ \\
 &  \linearact & \texttt{.*post\_.*\_layernorm} & 32 & 1.0 & $\color{red}{5.62}_{\color{gray}{\,0.25}}$ & $\color{red}{24.29}_{\color{gray}{\,1.46}}$ & $9.37_{\color{gray}{\,0.20}}$ & $45.95_{\color{gray}{\,0.04}}$ \\
 &  \method & \texttt{.*post\_.*\_layernorm} & 32 & 1.0 & $\textbf{2.30}_{\color{gray}{\,0.14}}$ & $\textbf{12.09}_{\color{gray}{\,0.83}}$ & $8.38_{\color{gray}{\,0.05}}$ & $48.13_{\color{gray}{\,0.07}}$ \\

\midrule
\multirow{4}{*}{\qwenonefiveb} &  None & \texttt{.*post\_.*\_layernorm} & - & - & $3.00_{\color{gray}{\,0.54}}$ & $23.09_{\color{gray}{\,0.67}}$ & $13.67_{\color{gray}{\,0.00}}$ & $60.95_{\color{gray}{\,0.00}}$ \\
 &  \iti & \texttt{.*post\_.*\_layernorm} & 32 & 1.0 & $2.62_{\color{gray}{\,0.30}}$ & $19.35_{\color{gray}{\,1.52}}$ & $13.23_{\color{gray}{\,0.17}}$ & $60.37_{\color{gray}{\,0.24}}$ \\
 &  \linearact & \texttt{.*post\_.*\_layernorm} & 32 & 1.0 & $2.75_{\color{gray}{\,0.68}}$ & $\color{red}{25.51}_{\color{gray}{\,1.79}}$ & $16.33_{\color{gray}{\,0.85}}$ & $57.66_{\color{gray}{\,0.56}}$ \\
 &  \method & \texttt{.*post\_.*\_layernorm} & 32 & 1.0 & $\textbf{1.07}_{\color{gray}{\,0.46}}$ & $\textbf{12.70}_{\color{gray}{\,0.74}}$ & $14.10_{\color{gray}{\,0.07}}$ & $59.97_{\color{gray}{\,0.16}}$ \\

\midrule
\multirow{4}{*}{\qwensevenb} &  None & \texttt{.*post\_.*\_layernorm} & - & - & $3.92_{\color{gray}{\,0.59}}$ & $25.16_{\color{gray}{\,0.92}}$ & $10.67_{\color{gray}{\,0.00}}$ & $74.26_{\color{gray}{\,0.00}}$ \\
 &  \iti & \texttt{.*post\_.*\_layernorm} & 32 & 1.0 & $2.88_{\color{gray}{\,0.60}}$ & $19.41_{\color{gray}{\,1.11}}$ & $9.69_{\color{gray}{\,0.07}}$ & $74.13_{\color{gray}{\,0.05}}$ \\
 &  \linearact & \texttt{.*post\_.*\_layernorm} & 32 & 1.0 & $2.77_{\color{gray}{\,0.39}}$ & $20.57_{\color{gray}{\,1.36}}$ & $11.64_{\color{gray}{\,0.24}}$ & $72.21_{\color{gray}{\,0.08}}$ \\
 &  \method & \texttt{.*post\_.*\_layernorm} & 32 & 1.0 & $\textbf{1.88}_{\color{gray}{\,0.19}}$ & $\textbf{15.39}_{\color{gray}{\,0.60}}$ & $10.83_{\color{gray}{\,0.25}}$ & $73.56_{\color{gray}{\,0.07}}$ \\

 \midrule
 \midrule
\multirow{4}{*}{\gemmatwob} &  None & \texttt{.*o\_proj} & - & - & $4.00_{\color{gray}{\,0.45}}$ & $13.39_{\color{gray}{\,1.42}}$ & $14.79_{\color{gray}{\,0.00}}$ & $53.03_{\color{gray}{\,0.00}}$ \\
 &  \iti & \texttt{.*o\_proj} & 32 & 0.5 & $\textbf{1.17}_{\color{gray}{\,0.60}}$ & $\textbf{7.15}_{\color{gray}{\,0.92}}$ & $14.00_{\color{gray}{\,0.11}}$ & $52.78_{\color{gray}{\,0.23}}$ \\
 &  \linearact & \texttt{.*o\_proj} & 32 & 1.0 & $1.60_{\color{gray}{\,0.32}}$ & $7.76_{\color{gray}{\,0.39}}$ & $14.78_{\color{gray}{\,0.12}}$ & $52.43_{\color{gray}{\,0.57}}$ \\
 &  \method & \texttt{.*o\_proj} & 32 & 1.0 & $2.10_{\color{gray}{\,0.34}}$ & $8.78_{\color{gray}{\,0.56}}$ & $14.72_{\color{gray}{\,0.16}}$ & $53.27_{\color{gray}{\,0.41}}$ \\

\midrule
\multirow{4}{*}{\deepseeksevenb} &  None & \texttt{.*o\_proj} & - & - & $4.30_{\color{gray}{\,0.70}}$ & $18.62_{\color{gray}{\,0.51}}$ & $8.49_{\color{gray}{\,0.00}}$ & $48.31_{\color{gray}{\,0.00}}$ \\
 &  \iti & \texttt{.*o\_proj} & 32 & 0.5 & $2.83_{\color{gray}{\,0.40}}$ & $15.18_{\color{gray}{\,2.00}}$ & $7.71_{\color{gray}{\,0.07}}$ & $48.47_{\color{gray}{\,0.25}}$ \\
 &  \linearact & \texttt{.*o\_proj} & 32 & 1.0 & $\textbf{2.23}_{\color{gray}{\,0.69}}$ & $11.08_{\color{gray}{\,0.76}}$ & $8.67_{\color{gray}{\,0.03}}$ & $47.71_{\color{gray}{\,0.27}}$ \\
 &  \method & \texttt{.*o\_proj} & 32 & 1.0 & $2.30_{\color{gray}{\,0.48}}$ & $\textbf{8.46}_{\color{gray}{\,0.54}}$ & $8.61_{\color{gray}{\,0.06}}$ & $46.35_{\color{gray}{\,0.37}}$ \\

\midrule
\multirow{4}{*}{\qwenonefiveb} &  None & \texttt{.*o\_proj} & - & - & $3.00_{\color{gray}{\,0.54}}$ & $23.09_{\color{gray}{\,0.67}}$ & $13.67_{\color{gray}{\,0.00}}$ & $60.95_{\color{gray}{\,0.00}}$ \\
 &  \iti & \texttt{.*o\_proj} & 32 & 0.5 & $1.87_{\color{gray}{\,0.21}}$ & $18.16_{\color{gray}{\,0.62}}$ & $12.39_{\color{gray}{\,0.09}}$ & $60.88_{\color{gray}{\,0.08}}$ \\
 &  \linearact & \texttt{.*o\_proj} & 32 & 1.0 & $\textbf{1.50}_{\color{gray}{\,0.35}}$ & $13.88_{\color{gray}{\,1.72}}$ & $13.88_{\color{gray}{\,0.16}}$ & $60.09_{\color{gray}{\,0.25}}$ \\
 &  \method & \texttt{.*o\_proj} & 32 & 1.0 & $\textbf{1.50}_{\color{gray}{\,0.29}}$ & $\textbf{12.03}_{\color{gray}{\,0.71}}$ & $14.04_{\color{gray}{\,0.10}}$ & $59.53_{\color{gray}{\,0.17}}$ \\

\midrule
\multirow{4}{*}{\qwensevenb} &  None & \texttt{.*o\_proj} & - & - & $3.92_{\color{gray}{\,0.59}}$ & $25.16_{\color{gray}{\,0.92}}$ & $10.67_{\color{gray}{\,0.00}}$ & $74.26_{\color{gray}{\,0.00}}$ \\
 &  \iti & \texttt{.*o\_proj} & 32 & 0.5 & $2.97_{\color{gray}{\,0.21}}$ & $20.68_{\color{gray}{\,2.21}}$ & $9.56_{\color{gray}{\,0.02}}$ & $74.20_{\color{gray}{\,0.07}}$ \\
 &  \linearact & \texttt{.*o\_proj} & 32 & 1.0 & $2.25_{\color{gray}{\,0.13}}$ & $16.04_{\color{gray}{\,0.95}}$ & $10.77_{\color{gray}{\,0.09}}$ & $73.56_{\color{gray}{\,0.08}}$ \\
 &  \method & \texttt{.*o\_proj} & 32 & 1.0 & $\textbf{2.00}_{\color{gray}{\,0.18}}$ & $\textbf{13.58}_{\color{gray}{\,0.76}}$ & $12.52_{\color{gray}{\,0.07}}$ & $71.34_{\color{gray}{\,0.10}}$ \\

\bottomrule
\end{tabular}
    }
\caption{\textbf{\method is more robust to the layer choice.} Toxicity mitigation on the RTP and TET datasets, intervening on \texttt{.*post\_.*\_layernorm} and  \texttt{.*o\_proj} layers. We report results at low (32 sentences) regime, showing in \textbf{bold} the best toxicity result per model. When intervening on \texttt{.*post\_.*\_layernorm}, both \iti and \linearact show poorer performance, specially for \deepseeksevenb and \qwenonefiveb, where the toxicity goes even above the original one (in {\color{red}red}). For \texttt{.*o\_proj}, \iti and \linearact perform better. Overall, \method shows strong robustness to the layer choice.}
    \label{tab:toxicity_layer}
\end{table}

\FloatBarrier
\section{Order of Magnitude of Interventions Parameters}
To inform the scale of regularization terms, we plot descriptive statistics of the values of $\allw$ and $\allb$, layer by layer.

\begin{figure*}[htbp]
    \centering
    \includegraphics[width=\textwidth]{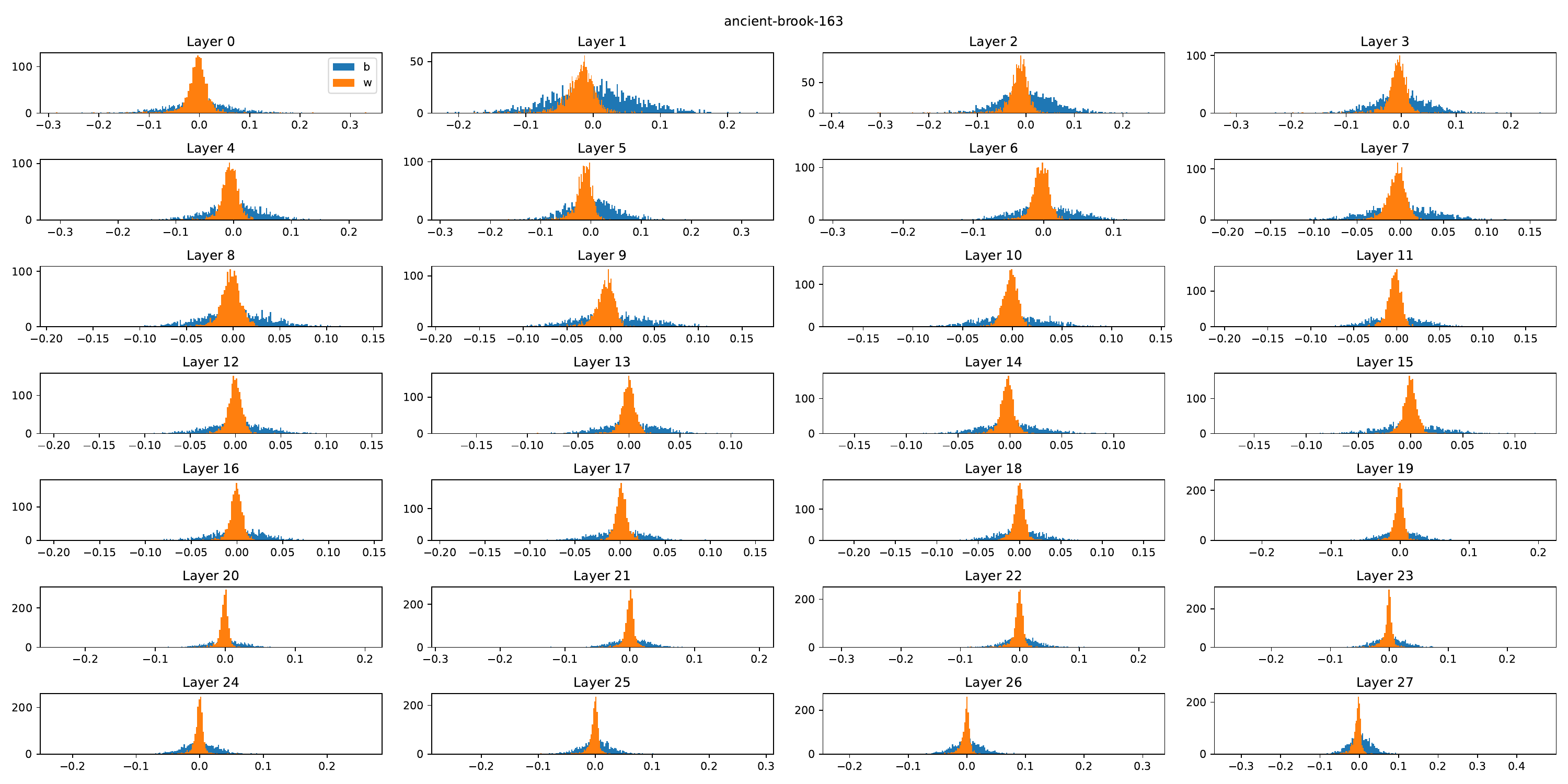}
    \caption{Distribution, layer by layer, of recentered scale parameters $\allw$ and $\allb$ biases, for a converged run of \method, intervening on the 28 intervened layers of \gemmatwob.}
    \label{fig:distrib}
\end{figure*}    

\paragraph{Sparsity and Refitting.} When $\reg \gg0$, several coordinates of ${\ROmega{\ell}}$ and ${\Rb{\ell}}$ (parameters of $\RT{\ell}$) will collapse to $1$ or $0$, respectively. While this is desired, non-zero parameters typically suffer from shrinking, where the regularization terms $\mathcal{R}_{1}, \mathcal{R}_{G} $ dampen the effect of $\mathcal{C}$. A typical solution to this phenomenon is to perform $m_{\text{post}}$ training steps updating only those non-collapsed parameters, a practice known as re-fitting in regression \citep{belloni2013least,chzhen2019lasso}. We have not observed improvements when refitting parameters, and therefore do not use it. We observe that the entries of $\ROmega{\ell}-\mathbf{1}$ and $\Rb{\ell}$ have similar scales (see \cref{fig:distrib}) and choose to use the same regularization strength.

\FloatBarrier
\section{Effect of Sparsity on Toxicity Mitigation (extended results)}
\label{app:tox_sparsity}

We complement the results shown in \Cref{fig:tox_sparsity_data32}, this time measuring the effect of sparsity on toxicity mitigation when the transport maps are optimized with 1024 (source and target) sentences. The results and trends discussed in \Cref{fig:tox_sparsity_data32} also hold in this setup with more data.

\begin{figure*}[htbp]
    \centering
    \includegraphics[trim={0 0 0 0},clip,width=\textwidth]{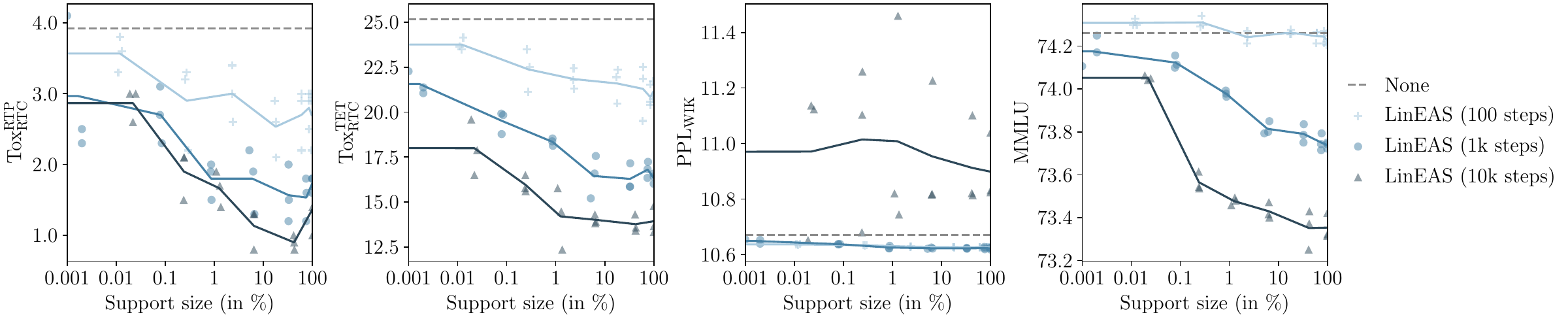}
    \caption{\textbf{Sparsity improves utility while mitigating toxicity, also in high data regime}. Toxicity results on \qwensevenb using 1024 sentences, at different levels of sparsity $\reg$ that result in different support sizes (x axis). With a support of 1\%-5\% we maintain similar toxicity (left, center-left) while \pplwik decreases (center-right) and MMLU increases (right). Note that too long optimizations (30k steps) might harm utility, due to overfitting. Similarly, short optimizations (\eg 300 steps) and strong sparsity leads to low conditioning (mild toxicity mitigation).}
    \label{fig:tox_sparsity_data1024}
\end{figure*}

\FloatBarrier
\section{Effect of Sparsity on T2I Generation}
\label{app:t2i-sparsity}
We ran a sweep over sparsity coefficients on DMD2~\citep{yin2024dmd2} and report it on~\cref{tab:t2i-sparsity}. We find that T2I models are \textit{more} sensitive than LMs to the sparsity penalty, almost saturating to either full support or no support outside the range $[0.4, 0.8]$. We hypothesize the UNet is more sensitive than the transformer because its activation maps are less redundant due to the changes in dimensionality as described by~\citet{veit2016residual} and \citet{jastrzebski2018residual}.

\begin{table}[h!]
\centering
\caption{Effect of the sparsity coefficient on DMD2. Performance metrics (support, IMGScore, and CLIPScore) at varying levels of sparsity. IMGScore generally increases with higher sparsity, while CLIPScore shows a slight increase.}
\label{tab:t2i-sparsity}
\begin{tabular}{cccc}
\toprule
\textbf{sparsity} & \textbf{support(\%)} & \textbf{IMGScore(\%)$\uparrow$} & \textbf{CLIPScore(\%)$\downarrow$} \\
\midrule
0.0 & 100$\pm$0.0 & 71.4$\pm$5.5 & 13.1$\pm$3.2 \\
0.4 & 92.7$\pm$5.0 & 84.7$\pm$6.0 & 14.5$\pm$2.9 \\
0.5 & 76.7$\pm$14.0 & 89.5$\pm$6.3 & 15.0$\pm$2.9 \\
0.6 & 51.6$\pm$22.1 & 94.8$\pm$3.6 & 15.5$\pm$2.8 \\
0.7 & 13.8$\pm$11.4 & 98.9$\pm$1.1 & 16.0$\pm$2.8 \\
0.8 & 3.3$\pm$3.2 & 99.6$\pm$0.4 & 16.1$\pm$2.8 \\
1.0 & 0.0$\pm$0.0 & 100$\pm$0.0 & 16.1$\pm$2.9 \\
\bottomrule
\end{tabular}
\end{table}

\FloatBarrier
\section{RLHF Implementation Details}
\label{app:rlhf}
We perform parameter-efficient adaptation of our baseline models with Huggingface's implementation of LoRA in their \href{https://huggingface.co/docs/peft/en/package_reference/lora}{PEFT} library and Huggingface's implementation of the PPO reinforcement learning algorithm in their \href{https://huggingface.co/docs/trl/v0.11.4/en/ppo_trainer}{TRL} library. For each sample size in~\Cref{tab:toxicity_main}, we performed an hyperparameter search and chose the hyperparameters that yielded best validation toxicity scores at a perplexity close to \method. 
Following~\citet{ouyang2022training}, we fine-tune the models using proximal policy optimization (PPO)~\citep{schulman2017proximal} on the same $N=32$ data and use \RTC~(Roberta toxicity classifier)\, as our reward model. We instantiate the reward model with the Roberta toxicity classifier from \citet{logacheva-etal-2022-paradetox} used for evaluation (\RTC{} in~\Cref{tab:toxicity_main}); we use the base model without LoRA weights as the reference model and the LoRA model as the policy; we add an off-the-shelf value head from TRL to the policy to estimate the value function. We follow the original LoRA implementation and only fine-tune \texttt{\{k,q,v,o\}\_proj} layers while keeping the MLPs frozen. The summary of hyperparameters can be found in~\Cref{tab:PPO_search}.

\begin{table}[htbp]
    \centering
    \begin{tabular}{l|ccc}
        \toprule
        Hyperparameter & Values & Best 32 Samples & Best 1024 Samples \\
        \midrule 
        \texttt{global\_epochs} & $\{10,15,20\}$ & $10$ & $15$ \\
        \texttt{ppo\_epochs} &  $\{1, 10, 20, 50\}$ & $20$ & $20$ \\
        \texttt{learning\_rate} & $\{10^{-4}, 5*10^{-5}, 10^{-5}\}$ & $10^{-5}$ & $10^{-5}$ \\
        \texttt{batch\_size} & $\{32, 64, 128\}$ & $32$ & $128$ \\
        \texttt{mini\_batch\_size} & $\{16, 32, 64\}$ & $32$ & $64$ \\
        \texttt{lora\_rank} &  $\{2, 4, 8\}$ & $2$ & $2$ \\
        \bottomrule
    \end{tabular}
    \vskip 1mm
    \caption{List of hyper-parameters used to train PPO for 32 and 1024 samples. Unless otherwise specified, we use \href{https://huggingface.co/docs/trl/v0.11.4/en/ppo_trainer}{TRL}'s defaults. We could not use a mini-batch size greater than 64 due to memory constraints. We did not use gradient accumulation. We found the \texttt{learning\_rate} and \texttt{global\_epochs} to be the most important hyper-parameters. Low learning rate for few epochs leads to underfitting while high learning rate for many epochs tends to overfit.}
    \label{tab:PPO_search}
\end{table}

\FloatBarrier
\section{Composition of Interventions}
\label{app:composition}
\begin{figure*}[htb]
    \centering

    \begin{subfigure}[b]{1.0\textwidth}
        \centering
        \includegraphics[width=\textwidth]{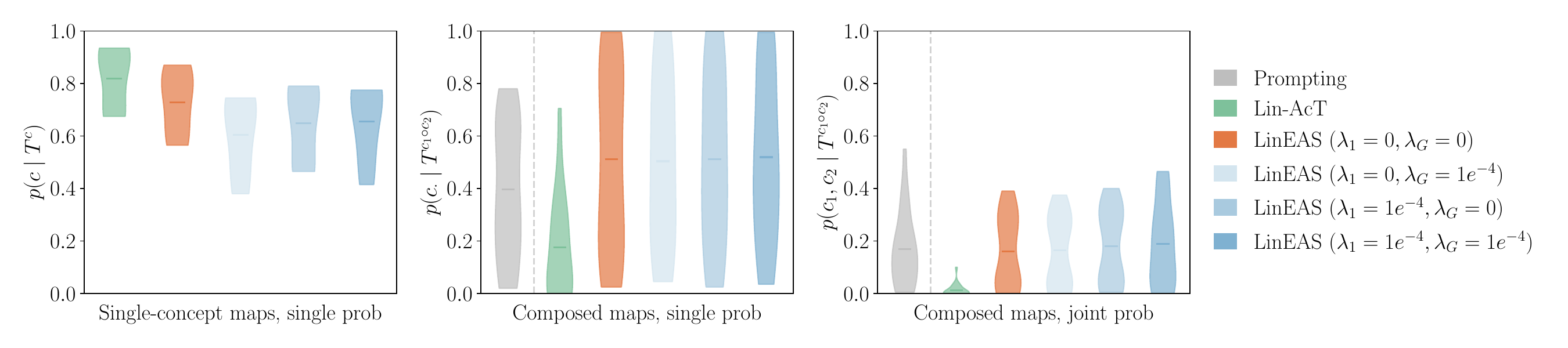}
        \label{fig:concept_merging}
    \end{subfigure}
    \vskip -6mm
    \caption{{\color{blue}} \textbf{\method obtains composable maps.} We benchmark Lin-ACT and \method for various regularization strengths $\gamma$ on a compositional task. We train each method to model distribution shifts towards a certain concept $c$ taking in a list of 5 concepts. \textbf{(Top-left)} We measure, using LLM as a judge, whether sentences formulated \textit{using that trained shift} do contain that concept. We obtain for each concept a probability, that we then aggregate using violin plots. \textbf{(Top-middle)} We pick two concepts $c_1$ and $c_2$ randomly, train separately their maps, and compose them. We then measure the probability of finding $c_1$, and then $c_2$, in the generated content, using the composition of both maps (in both orders). \textbf{(Top-right)} We now measure the probability of finding simultaneously \textit{both} concepts in the generations. As can be seen, the composition of \method maps (learnt separately for 2 concepts) yields an average $15\times$ (right) higher probability of including both 2 concepts in the same generation, with respect to \linearact, and a $3.1\times$ increase in generating at least one of the concepts (center). Interestingly, \linearact obtains slightly better probability when using only single-concept maps (left), but these fail at composition. Additionally, note that sparsity in \method is beneficial for compositionality, increasing the joint concept presence from 0.16 to 0.19 when using \textit{group lasso}. 
    } %
    \label{fig:compose_sparsity}
\end{figure*}

We evaluate in this section the ability to compose two \method maps pre-trained independently on two concepts. Our goal is to assess whether they can be composed, at the level of each activation, to induce \textit{both} concepts. Achieving concept composition in activation steering is an open goal,%
and our hypothesis is that sparse and end-to-end \method maps affect minimally the model, facilitating composisition.

\textbf{Setup.} For each of the concepts \textit{day, night, elephant, football} and \textit{fishing}, we generate $N=50$ diverse sentences using \gemmatwosevenb that contain that concept, to form five target $(q_i)$. Additionally, we ask \gemmatwosevenb to generate 50 diverse sentences about generic situations, forming a single source distribution $p$. We then learn five steering maps, from $p$ to each of the $q_i$ distributions, using both \linearact and \method. 
For \method we test all combinations of $\lambda_1=0, 1e^{-4}$ and $\lambda_G=0, 1e^{-4}$, to assess the impact of sparsity. Equipped with these five concept maps, we compose them for each pair of concepts $c_1, c_2$ as follows: $\RT{\ell}^{c_1\circ c_2}(z) := \RT{\ell}^{c_2}\big(  \RT{\ell}^{c_1}(z)\big)$. Note that $\RT{\ell}^{c_1\circ c_2} \neq \RT{\ell}^{c_2\circ c_1}$. 
Following \citet{rodriguez2024controllinglanguagediffusionmodels}, we intervene on \gemmatwob by generating 200 sentences that follow the prompt \textit{Once upon a time} and we measure the presence of the concepts in the generations in a \textit{LLM-as-a-judge manner}, querying \qwensevenbinstr. 
More precisely, we query about: the presence of a concept $c$ in each generated sentence, when using the map trained with concept $c$; the presence of a concept $c$, when using the map trained with that concept \textit{and} any other; the presence of both \textit{two} concepts $c_1, c_2$ in the same sentence, yielding $p(c_1, c_2)$, using either $\RT{\ell}^{c_1\circ c_2}$ or $\RT{\ell}^{c_2\circ c_1}$.%

\textbf{Results, comparing interventions. } \Cref{fig:compose_sparsity}.(top) plots the three probabilities described above. In particular, the right plot shows $p(c_1, c_2\mid T^{c_1\circ c_2})$, \ie the probability of observing both concepts in the same generated sentence, using the composed map. \linearact is able to generate concepts using single-concept maps (left plot) with average probability of 0.82 \textit{vs.} 0.73 using \method without sparsity. We also observe that increasing the  sparsity (larger $\lambda$s) slightly diminishes the presence of concepts when using single-concept maps. However, we observe a drastically different picture when using \textit{combined} maps: (middle) that probability goes from a \linearact average of $0.17$ to around $0.52$ with \method ($3.1\times$ increase). Most importantly, the joint presence of concepts probability (right) goes from $0.013$ for \linearact to $0.19$ ($15\times$ increase) for \method with \textit{group lasso} regularization (both $\lambda_1,\lambda_G$ used). See \Cref{app:composition} for generation examples. These results show that \method learns maps that are easier to compose than those from \linearact. Indeed, composition of \linearact maps is very brittle. While \method achieves much stronger compositionality, our results show that there is still room for improvement on this important problem. We provide qualitative examples in~\Cref{fig:fig_intro}                     .

\rebuttal{\textbf{Results, prompting as baseline.} We prompt the LLM to complete a generation with two concepts using \textit{"Continue the following text, make sure concepts [{c1}] and [{c2}] appear in the continuation: Once upon a time"}. We observe that the mean probability of:
\begin{itemize}
    \item generating either of the concepts (middle plot) is 39\% with prompting / 52\% for \method.
    \item generating both concepts (right plot) is 17\% for prompting / 19\% for \method.
\end{itemize}
In light of these results, we conclude that prompting is a strong baseline, since it has direct access to the concepts in textual form at generation time. However, \method achieves better compositions without such direct access.}

\subsection{Composition of Interventions: Qualitative Results}

We show qualitative results using composed maps as detailed in \Cref{app:composition}. In \Cref{tab:examples-fishing-elephant} and \Cref{tab:examples-night-football} we include generations that were marked as containing both concepts using LLM-as-a-judge (\qwensevenbinstr). We compare \linearact and \method (with $\reg=1e^{-4}$).

\begin{table}[htb]
\caption{
Generations inducing both concepts \textit{Fishing}, \textit{Elephant}. Only the provided 2 generations for \linearact were marked as containing both concepts using LLM-as-a-judge, out of 200. Conversely, 25 sentences generated by \method contain both concepts. Note the stark difference in quality, \method achieves high quality generations with composition. 
}
\label{tab:examples-fishing-elephant}
\renewcommand{\arraystretch}{1.5} %
\begin{tabular}{llp{0.8\linewidth}}
\toprule
Composition & Method & Generation \\
\midrule
Fishing $\circ$ Elephant & \linearact  & \once, the $<$strong$>$$<$em$>$Tuffa as the bull elephant's foot as she made a small <strong>putty as the water as she nosed the bank of the river as the water, as the sun as the $<$strong$>$swaying on the water as she stood in the shallows, the $<$strong$>$as the water [...]  \\
Fishing $\circ$ Elephant & \linearact  & \once, a large, gray, as it was the bull as it snorted in the water as it struggled to break the line of the bank as it thrashed in the water, a small stream of water, as it made a final sprint, as the water, as it snorted noisily, the water as the river [...]  \\

\midrule
Fishing $\circ$ Elephant & \method  & \once, a man was fishing in the river when he saw a beautiful, pink baby elephant walking towards him. The elephant was thirsty and the man offered him a drink of water. The elephant was grateful and splashed some water on the man's face.  \\
Fishing $\circ$ Elephant & \method  & \once, a huge elephant's tusk broke the water, his trunk splashing in the shallow river. "Hup!" he called, his tail swished against the muddy bank. I watched from the shore, my fishing rod dangling in the water.   \\
Fishing $\circ$ Elephant & \method  & \once, the elephant’s trunk broke the water, his massive body rising and disappearing. The jungle rumbled in the distance, their long tusks scraping against the mud. It was a young calf, its small, wet back.
The fisherman sat on the bank, his net swinging lazily in the shallow water. A small fish  \\

\bottomrule
\end{tabular}
\end{table}

\begin{table}[htb]
\caption{
Generations inducing both concepts \textit{Night}, \textit{Football}.
}
\label{tab:examples-night-football}
\renewcommand{\arraystretch}{1.5} %
\begin{tabular}{llp{0.8\linewidth}}
\toprule
Composition & Method & Generation \\
\midrule
Night $\circ$ Football & \linearact  & \once, a little girl's dream came true. It was the first night of the 2013-14 season and the young forward had just scored her first goal for the first team. The ball had nestled in the net in the 15th minute of the game, and the 16-year-old couldn't believe her luck.  \\
Night $\circ$ Football & \linearact  & \once, there was a man who was so determined to win the game, he went out to play in the rain.
It was a cold, wet night, and the rain was pouring down, but the man didn't let it dampen his spirits. He was out to win the game, and he knew that he had to make a late, late goal to seal the victory.  \\
Night $\circ$ Football & \linearact  & \once, I was a happy camper.
I was in the middle of a long drive through the woods, the sound of the wind whistling through the trees. The air was cool and the sun was setting, casting a warm glow over the forest.
As I drove, I couldn't help but feel a sense of excitement. I had just won the race to the finish line, the ball bouncing off the net, and the ball was heading for the goal.
  \\
\midrule
Night $\circ$ Football & \method  & \once, in the night, the lights of the field illuminated the players. The crowd roared in excitement, their voices echoing off the stands. It was the final whistle, and the opposing team celebrated, their coach shouting with joy.\\
Night $\circ$ Football & \method  & \once, the \emph{“Let’s go Rangers!”} could be heard through the dark, cold night. The home crowd cheered and the ball flew past the goal, sending a shower of confetti into the air. \\
Night $\circ$ Football & \method  & \once, the sun set on the field of grass. The opposing teams were locked in a fierce battle, the crowd roaring with excitement. It was the opening goal, and the referee blew his whistle, signaling the end of the game. \\
\bottomrule
\end{tabular}
\end{table}

\FloatBarrier
\section{Group Sparsity Trade-offs}
\label{app:interpretability}
\begin{figure*}[htb]
    \centering
    \begin{subfigure}[b]{0.99\textwidth}
        \includegraphics[trim={0 0 0 0},clip,width=\textwidth]{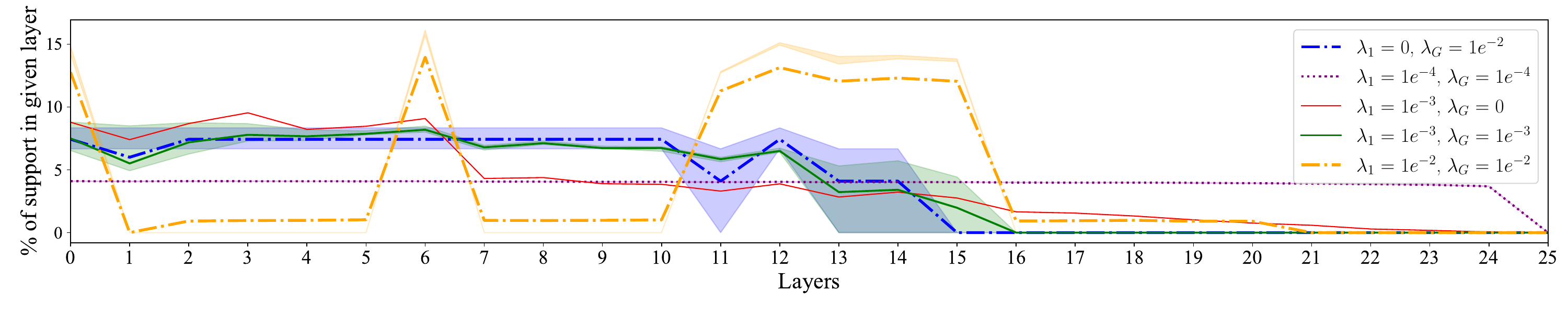}
        \caption{\method Distribution of support across layers (biases). Post Feed-Forward Layer Norms.}
        \label{}
    \end{subfigure}
    \hfill
    \begin{subfigure}[b]{0.99\textwidth}
        \includegraphics[trim={0 0 0 0},clip,width=\textwidth]{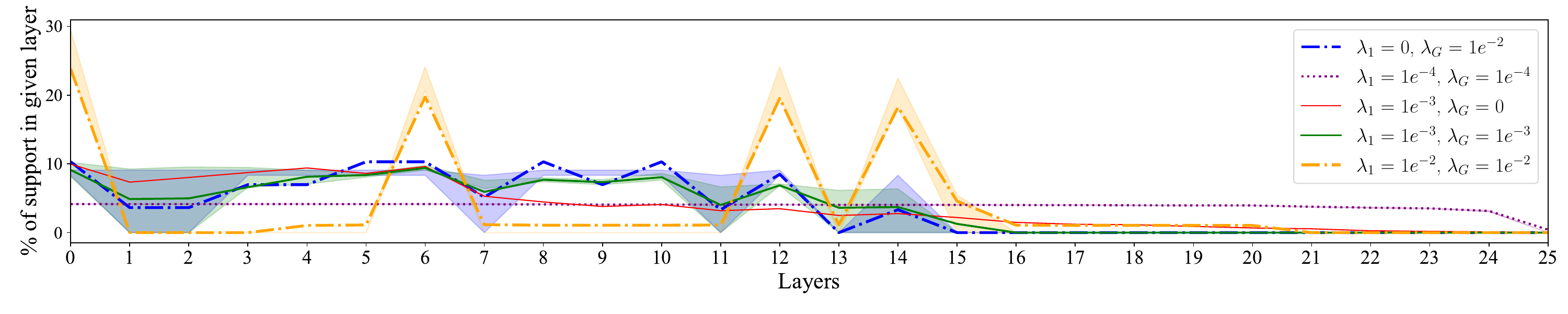}
        \caption{\method Distribution of support across layers (biases). Post Attention Layer Norms.}
        \label{}
    \end{subfigure}
    \hfill
    \begin{subfigure}[b]{0.99\textwidth}
        \includegraphics[trim={0 0 0 0},clip,width=\textwidth]{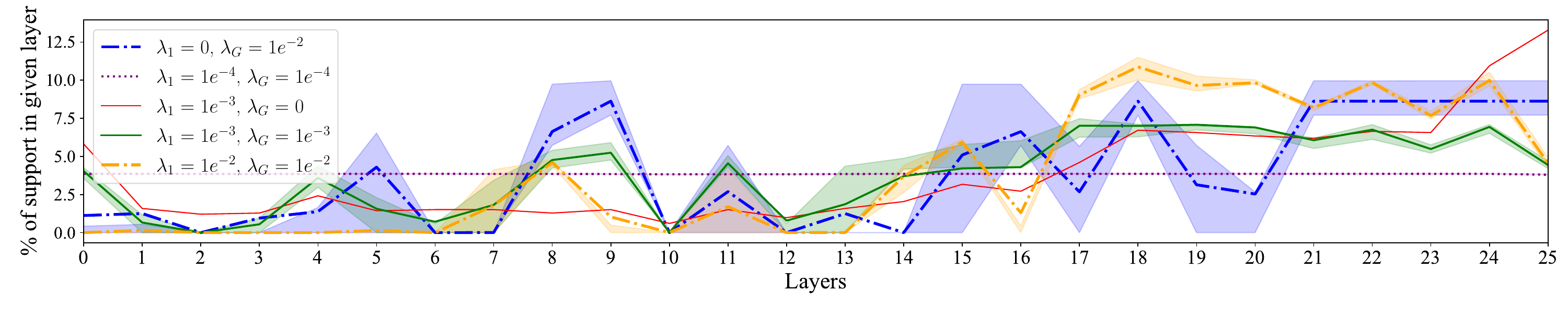}
        \caption{\method Distribution of support across layers (weights). Post Feed-Forward Layer Norms.}
        \label{}
    \end{subfigure}
    \hfill
    \begin{subfigure}[b]{0.99\textwidth}
        \includegraphics[trim={0 0 0 0},clip,width=\textwidth]{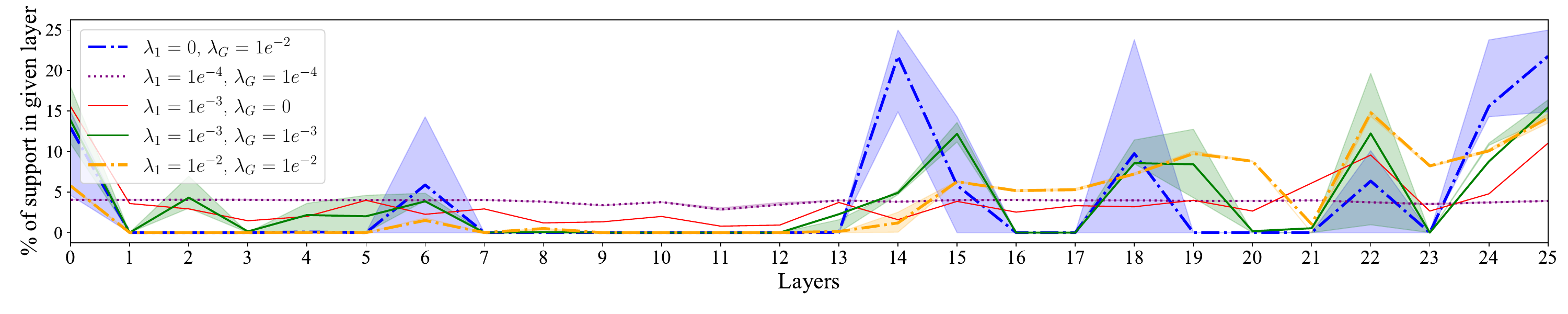}
        \caption{\method Distribution of support across layers (weights). Post Attention Layer Norms.}
        \label{}
    \end{subfigure}
    \caption{\textbf{\method Distribution of support across layers.} Each subfigure shows the percentage of intervened units (out of the total support) across layers for different regularization strengths, averaged over 5 concepts (with 50\% quantile range). %
    }
    \label{fig:interpretability_app}
\end{figure*}

We consider the interventions learned with Gemma2-2B for the concepts \textit{day}, \textit{night}, \textit{elephant}, \textit{football}, and \textit{fishing}, but set both $\lambda_1>0$ and $\lambda_G>0$, enforcing a sparse group regularizer.
Figure \ref{fig:interpretability_app} shows how the proportion of intervened units, out of their entire support, is distributed across layers. Since we are plotting proportions of the support, differences between models in terms of the absolute number of impacted neurons are not reflected. Rather, the plot allows to grasp how regularization, and notably group regularization, impacts layer selection. An important finding of this experiment is that we do observe, for high group regularization regimes, that very specific layers are consistently selected to induce diverse concepts. Additionally, we posit that this grouping of neurons at certain layers impacts positively efficient transfer of concepts, notably when investigating compositionality.

\FloatBarrier
\section{Similarity of \method Interventions with Human Judgment}
\label{app:men_similarity}

\rebuttal{We draw a set of 50 concepts, from the MEN dataset \citep{men_dataset}, a resource of 3,000 word pairs annotated with human similarity judgments (see also \citet{fedzechkina2025analyze} for a study on LLM interpretability building on the same resource). With these 50 concepts, we can recover 20 word pairs annotated for their similarity in the MEN dataset.
We train \method interventions for each concept on \gemmatwob. We then ask: do we recover similar interventions for similar concepts, and does sparsity help? We can answer positively to both: highly similar concepts have highly similar interventions; and enforcing sparsity through our scheme improves that correlation.}

\begin{figure}[htb]
    \centering
    \begin{subfigure}[b]{0.49\textwidth}
           \includegraphics[width=\linewidth]{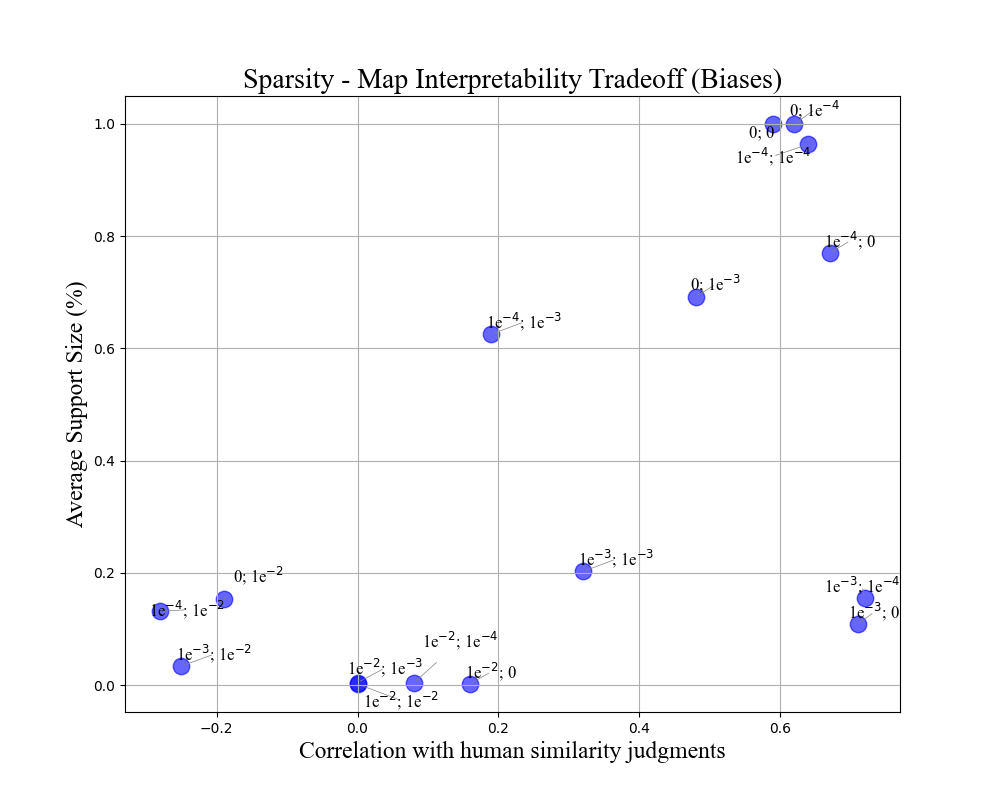} 
    \end{subfigure}
    \begin{subfigure}[b]{0.49\textwidth}
        \includegraphics[width=\linewidth]{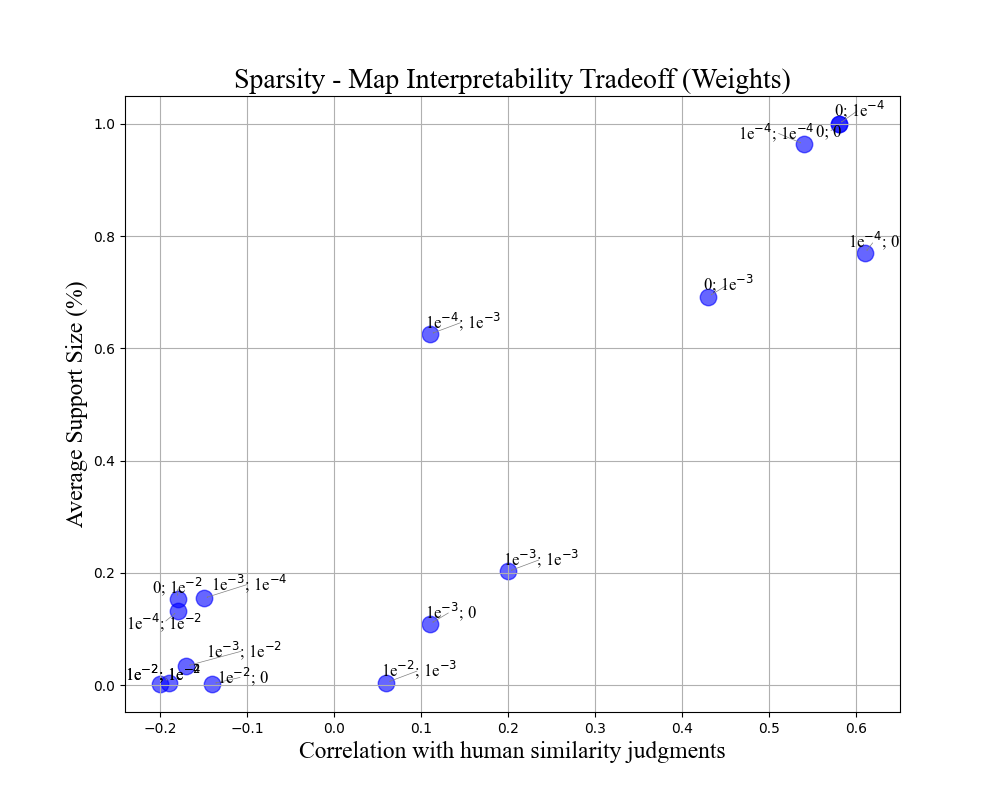}
    \end{subfigure}
   \caption{Scatter plots showing the correlation of interventions with human similarity judgment. We show the correlation with bias similarity in the left plot and with weight similarity on the right plot.}
    \label{fig:scatter_men}
\end{figure}

\rebuttal{To measure this, we compute the average sparse support of interventions (shown as y-axis in Figure \ref{fig:scatter_men}). We compute the similarity between the intervention vectors for each word pair, focusing on biases first: $s_{int}^b = \{\text{sim}(\vb_{c1}, \vb_{c2})\}_{\forall c1 \neq c2}$ (on the left in Figure \ref{fig:scatter_men}). Finally, we consider the human similarity judgments $s_{hum}$ reported in the MEN dataset for each word pair, and compute their correlation with the intervention similarity, $\text{corr}(s_{int}^b, s_{hum})$. We do the same for the weights, in this case through $s_{int}^w$ (on the right in Figure \ref{fig:scatter_men} and ). The numbers in the scatterplots in \Cref{fig:scatter_men} correspond to the \method regularization parameters $\lambda_1, \lambda_G$.}

\rebuttal{Note that sparsity helps improve correlation beyond the non-sparse version of \method (noted as 0;0, with a support of 100\% in \Cref{fig:scatter_men}). Overall, \method shows a strong correlation with human judgment.}

\begin{figure}[htb]
    \centering
    \begin{subfigure}[b]{0.39\textwidth}
           \includegraphics[width=\linewidth]{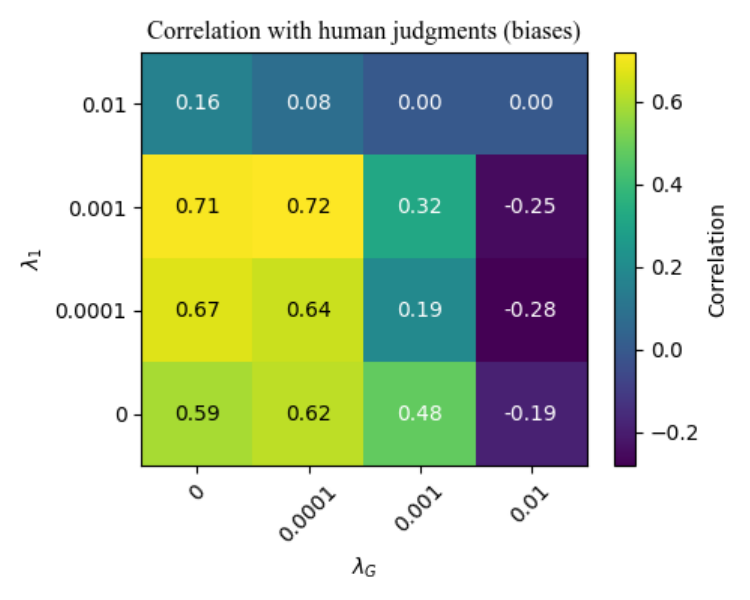} 
    \end{subfigure}
    \begin{subfigure}[b]{0.39\textwidth}
        \includegraphics[width=\linewidth]{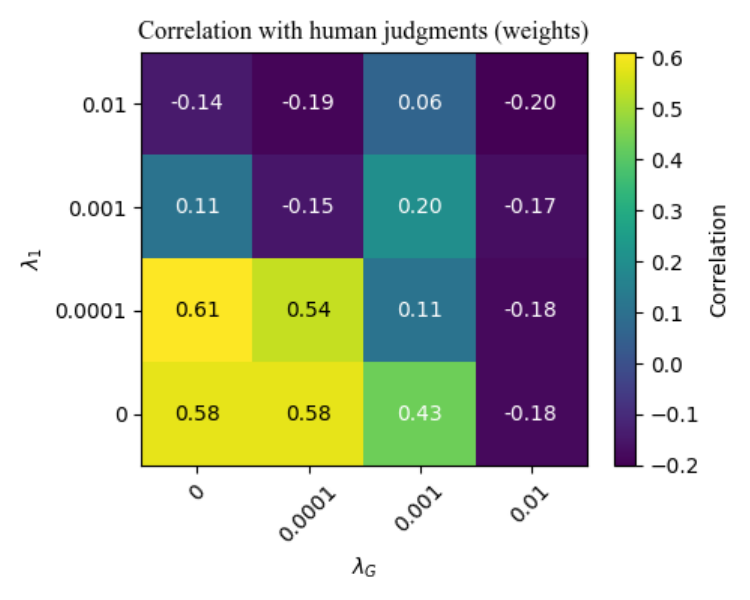}
    \end{subfigure}
    \caption{Heatmaps aggregating scatterplot results above but with a different view. We observe that highest correlations are obtained with a suitable regularization strength, for biases $\lambda_1=10^{-3}, \lambda_G=10^{-4}$, for weights $\lambda_1=10^{-4}, \lambda_G=0$.}
    \label{fig:matrices_men}
\end{figure}

\rebuttal{In \Cref{fig:matrices_men} we show the same results in form of matrix, showing that sparsity is indeed helpful to improve correlation with human alignment.}
\FloatBarrier

\FloatBarrier

\section{Additional Experiment: Inducing Truthfulness in LLMs}
\label{tqa}

In this section, we complement and corroborate our insights from the experiments on toxicity mitigation in \cref{sec:tox} with additional experiments on inducing truthfulness in LLMs, using the TruthfulQA benchmark \cite{lin2021truthfulqa}. 
In particular, we investigate how well \method achieves to induce truthfulness on this benchmark in comparison to {\linearact}, its strongest activation steering competitor from \cref{sec:tox}.

For \method, we apply the intervention again to the post layernorm layers, while for \linearact, we apply them to all layernorm layers as this was reported as optimal for \linearact for TruthfulQA experiments in \citet{rodriguez2024controllinglanguagediffusionmodels}.
We use 2-fold cross-validation on the $817$ questions of the multiple choice part of the benchmark and learn the intervention on the concatenation of training fold questions concatenated with either incorrect (source) or correct (target) multiple-choice answer options.  
We report both MC1 and MC2 of TruthfulQA, and monitor overfitting on the TruthfulQA task by also evaluating MMLU 5-shot accuracy \citep{hendryckstest2021}.

The results can be found in \cref{tab:tqa}. 
We see that both methods can successfully induce truthfulness when presented with enough samples to learn the interventions, increasing the accuracy by up to almost $5 \%$ ($7 \%$) on MC1 (MC2) in the high sample regime when 1024 samples are available. 
Overall the highest increases can be achieved with \linearact, but only for the high sample regime. 
In the low sample regime, where only 32 samples are available, \linearact tends to fail catastrophically: either it gets lower accuracies on MC1 and MC2 than even the unintervened model (\qwensevenb, \qwenonefiveb), or it fails completely on MMLU (\gemmatwob). 
\method on the other hand does well also in this low sample regime, and achieves second best overall performance on the Qwen models with only 32 samples available to learn interventions.

\begin{table}[htb!]
    \centering
    
\begin{tabular}{lllrrr}
        \toprule
        Model & Samples & Method & MC1 Acc. (\%) $(\uparrow)$ & MC2 Acc. (\%)$(\uparrow)$ & MMLU $(\uparrow)$ \\
        
        \midrule

        \multirow{5}{*}{\qwensevenb} 
        &   -  & None   & $37.82_{\color{gray}{\,0.00}}$ & $52.14_{\color{gray}{\,0.00}}$ & $74.26_{\color{gray}{\,0.00}}$  \\
        \cdashline{2-6}
        & \multirow{2}{*}{32} & \linearact  & $32.17_{\color{gray}{\,0.66}}$ & $47.69_{\color{gray}{\,0.87}}$ & $59.22_{\color{gray}{\,1.74}}$  \\
        & & \method  & $\underline{40.10}_{\color{gray}{\,0.37}}$ & $\underline{55.74}_{\color{gray}{\,0.31}}$ & $\textbf{73.88}_{\color{gray}{\,0.07}}$ \\
        \cdashline{2-6}
        & \multirow{2}{*}{1024} & \linearact  & $\textbf{42.59}_{\color{gray}{\,0.50}}$ & $\textbf{58.92}_{\color{gray}{\,0.82}}$ & $73.79_{\color{gray}{\,0.14}}$ \\
        & & \method  & $39.56_{\color{gray}{\,0.13}}$ & $55.20_{\color{gray}{\,0.15}}$ & $\underline{73.88}_{\color{gray}{\,0.02}}$ \\

        \midrule
        
        \multirow{5}{*}{\qwenonefiveb} 
        &   -  & None   & $30.23_{\color{gray}{\,0.00}}$ & $43.70_{\color{gray}{\,0.00}}$ & $60.95_{\color{gray}{\,0.00}}$  \\
        \cdashline{2-6}
        & \multirow{2}{*}{32} & \linearact  & $27.22_{\color{gray}{\,1.38}}$ & $42.64_{\color{gray}{\,3.17}}$ & $32.10_{\color{gray}{\,1.95}}$ \\
        & & \method  & $\underline{32.26}_{\color{gray}{\,0.65}}$ & $\underline{46.07}_{\color{gray}{\,0.63}}$ & $\underline{60.34}_{\color{gray}{\,0.12}}$ \\
        \cdashline{2-6}
        & \multirow{2}{*}{1024} & \linearact  & $\textbf{32.90}_{\color{gray}{\,0.36}}$ & $\textbf{47.17}_{\color{gray}{\,0.61}}$ & $60.17_{\color{gray}{\,0.18}}$ \\
        & & \method  & $31.77_{\color{gray}{\,0.11}}$ & $45.31_{\color{gray}{\,0.20}}$ & $\textbf{60.41}_{\color{gray}{\,0.01}}$ \\
        
        \midrule
        
        \multirow{5}{*}{\gemmatwob} 
        &   -  & None   & $21.18_{\color{gray}{\,0.00}}$ & $33.05_{\color{gray}{\,0.00}}$ & $53.03_{\color{gray}{\,0.00}}$  \\
        \cdashline{2-6}
        & \multirow{2}{*}{32} & \linearact  & $\underline{24.94}_{\color{gray}{\,1.15}}$ & $\textbf{40.95}_{\color{gray}{\,1.29}}$ & $27.59_{\color{gray}{\,1.09}}$ \\
        & & \method  & $24.21_{\color{gray}{\,0.85}}$ & $38.09_{\color{gray}{\,1.14}}$ & $\underline{52.08}_{\color{gray}{\,0.16}}$ \\
        \cdashline{2-6}
        & \multirow{2}{*}{1024} & \linearact  & $\textbf{25.65}_{\color{gray}{\,0.53}}$ & $\underline{39.73}_{\color{gray}{\,0.49}}$ & $51.40_{\color{gray}{\,0.17}}$ \\
        & & \method  & $23.82_{\color{gray}{\,0.19}}$ & $37.80_{\color{gray}{\,0.53}}$ & $\textbf{52.37}_{\color{gray}{\,0.05}}$ \\
        
        \bottomrule
    \end{tabular}

    \vskip 1mm
    \caption{Results on TruthfulQA. We report results at low data (32 samples to estimate the interventions) and high (1024 samples) regimes. Results are averaged over five random seeds.}
    \label{tab:tqa}
\end{table}

\clearpage
\FloatBarrier
\section{Additional details and results on text to image generation}
\label{app:t2i}

\FloatBarrier
\subsection{Additional qualitative results with LLM-generated prompts}

\begin{figure}
    \centering
    \includegraphics[width=\linewidth]{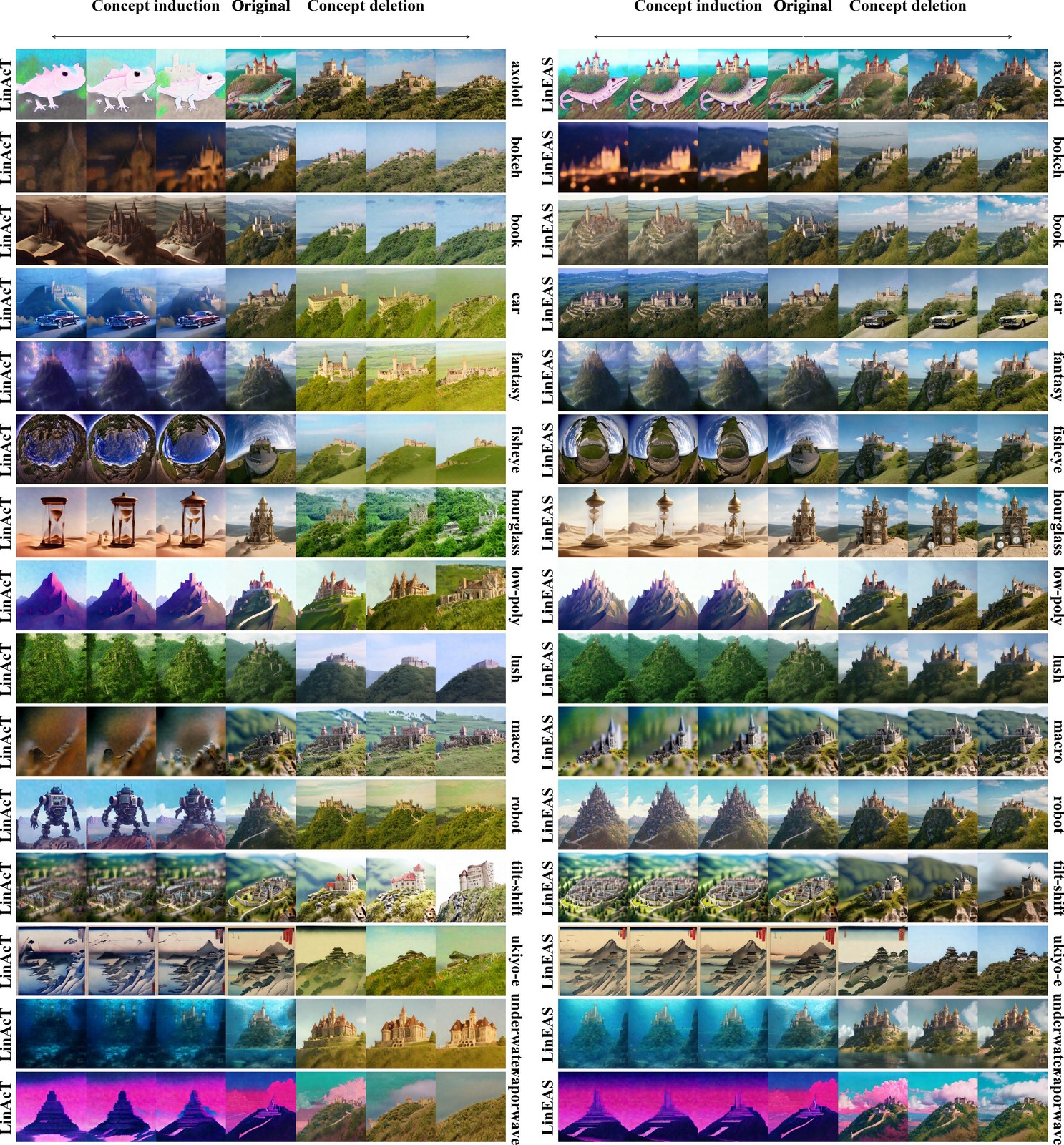}
    \caption{Images generated with \linearact (left panel) and \method (right panel) with the prompt \textit{A grand castle sits atop a hill overlooking a valley. [concept tag]} Each row contains a different conditioning concept to be mitigated by the steering method and each column a different intervention strength ($\lambda$). Each column contains a generation for $\lambda=1.0, 0.8, 0.6, 0, 0.6,0.8,1.0$ respectively and the columns to the left of ``original'' contain generations using $\RT{\ell}^{-1}(z)$.}
    \label{fig:app-qualitative-1}
\end{figure}

\begin{figure}
    \centering
    \includegraphics[width=\linewidth]{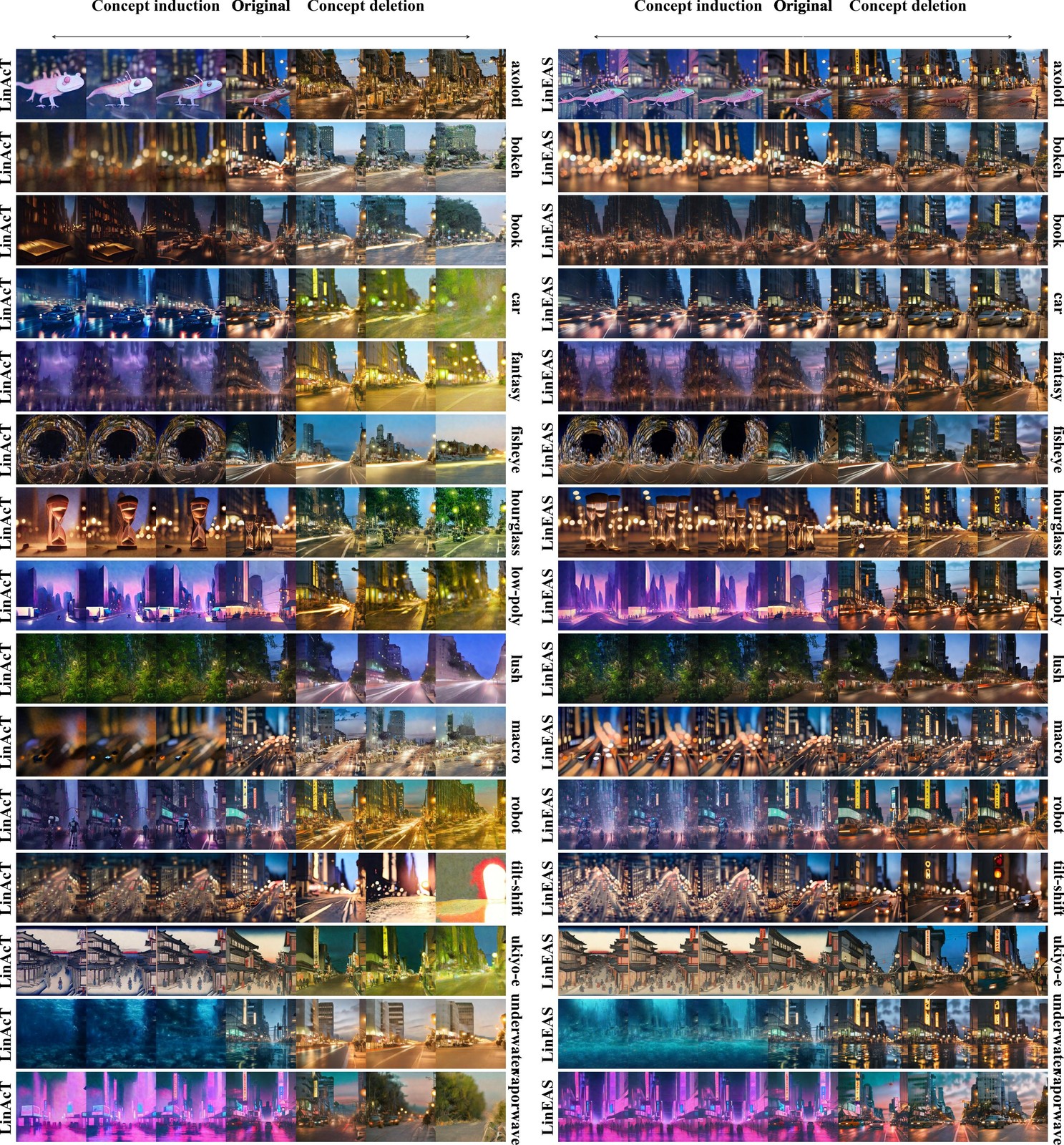}
    \caption{Images generated with \linearact (left panel) and \method (right panel) with the prompt \textit{A bustling city street at twilight, lights blurring. [concept tag]} Each row contains a different conditioning concept to be mitigated by the steering method and each column a different intervention strength ($\lambda$). Each column contains a generation for $\lambda=[1.0 ,..., 0, ..., 1.0]$ respectively and the columns to the left of ``original'' contain generations using $\RT{\ell}^{-1}(z)$.}
    \label{fig:app-qualitative-2}
\end{figure}

\begin{figure}
    \centering
    \includegraphics[width=\linewidth]{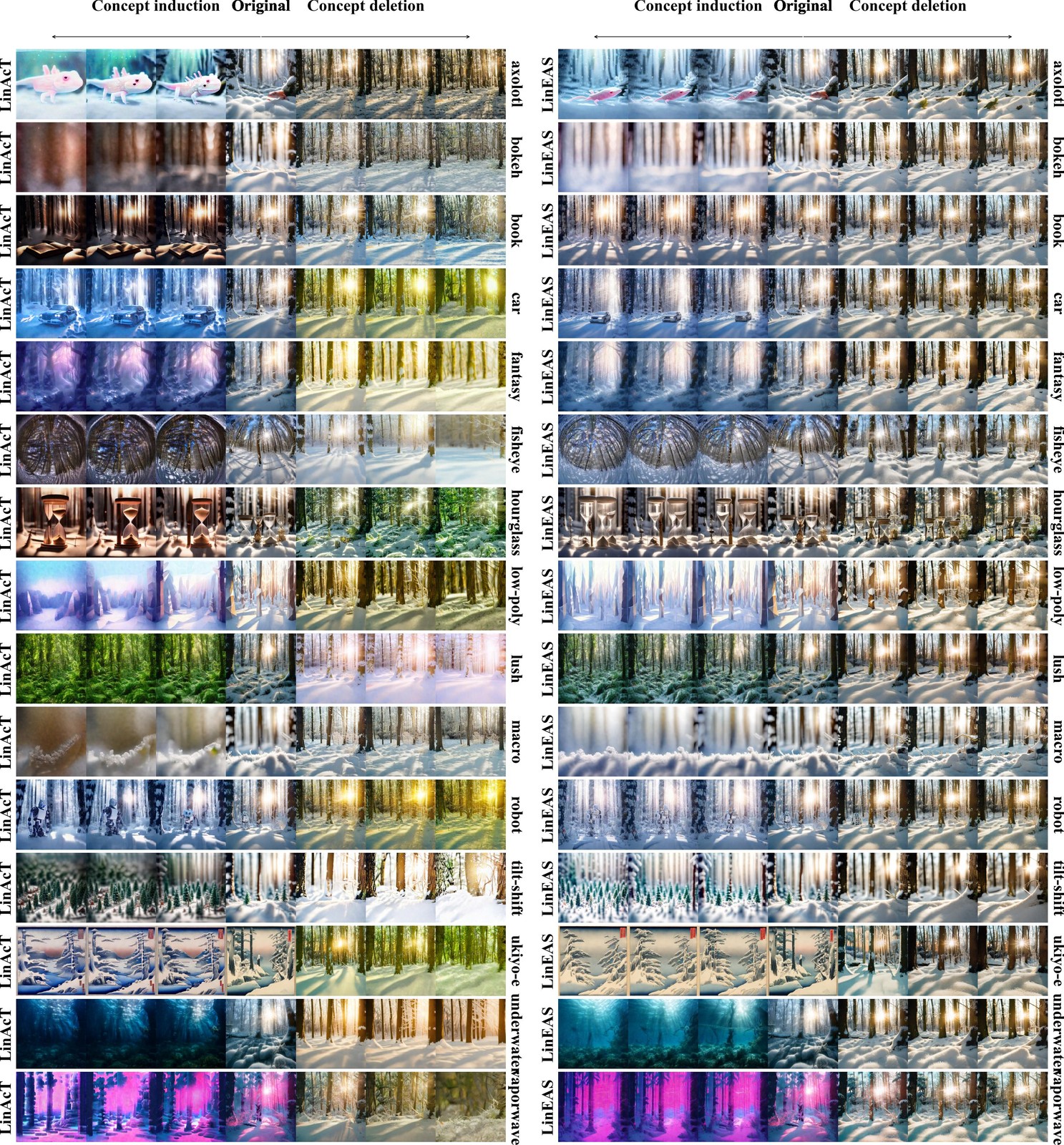}
    \caption{Images generated with \linearact (left panel) and \method (right panel) with the prompt \textit{A snow-covered forest with sunlight filtering through the trees. [concept tag]} Each row contains a different conditioning concept to be mitigated by the steering method  and each column a different intervention strength ($\lambda$). Each column contains a generation for $\lambda=[1.0 ,..., 0, ..., 1.0]$ respectively and the columns to the left of ``original'' contain generations using $\RT{\ell}^{-1}(z)$.}
    \label{fig:app-qualitative-3}
\end{figure}

\begin{figure}
    \centering
    \includegraphics[width=\linewidth]{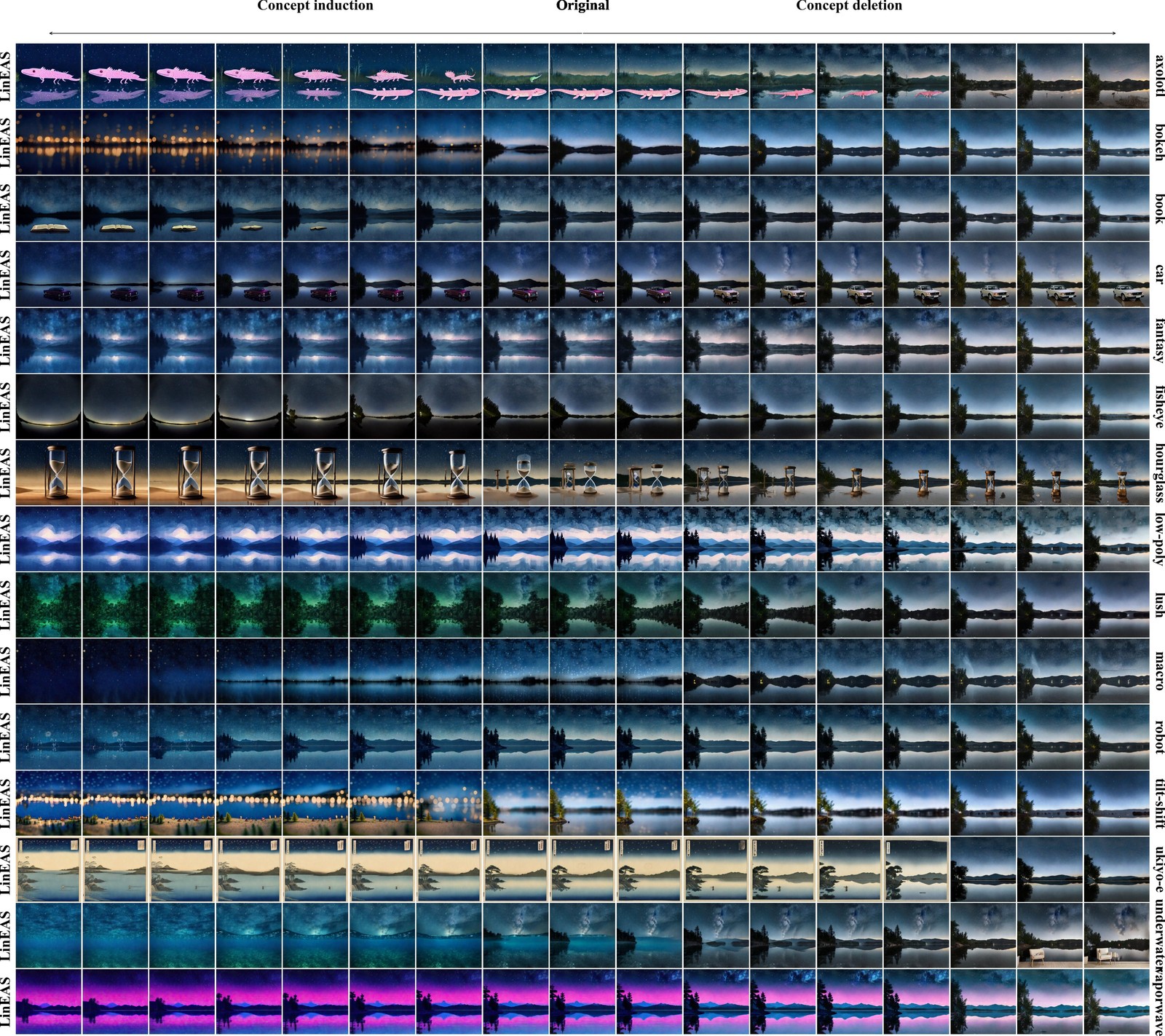}
    \caption{Images generated with \method with the prompt \textit{A starry night sky over a calm lake. [concept tag]} Each column contains images generated with a different steering strength for mitigating the concept corresponding to the row: $\lambda=[1.0 ,..., 0, ..., 1.0]$. Images to the left of ``original'' were produced using the inverse steering map $\RT{\ell}^{-1}(z)$. }
    \label{fig:app-qualitative-4}
\end{figure}

\begin{figure}
    \centering
    \includegraphics[width=\linewidth]{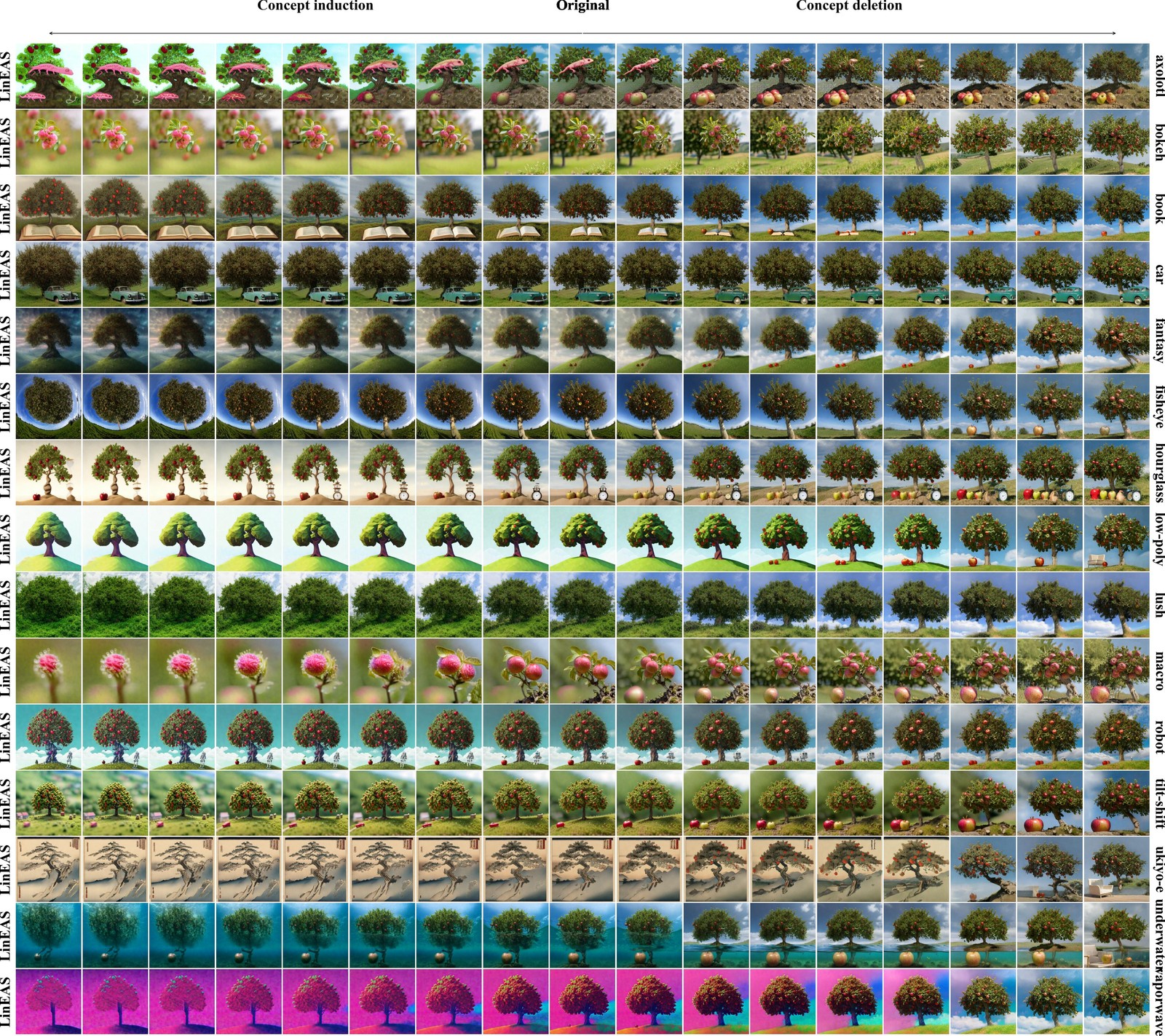}
    \caption{Images generated with \method with the prompt \textit{A fruiting apple tree on top of a hill. [concept tag]} Each column contains images generated with a different steering strength for mitigating the concept corresponding to the row: $\lambda=[1.0 ,..., 0, ..., 1.0]$. Images to the left of ``original'' were produced using the inverse steering map $\RT{\ell}^{-1}(z)$. }
    \label{fig:app-qualitative-5}
\end{figure}

\begin{figure}
    \centering
    \includegraphics[width=\linewidth]{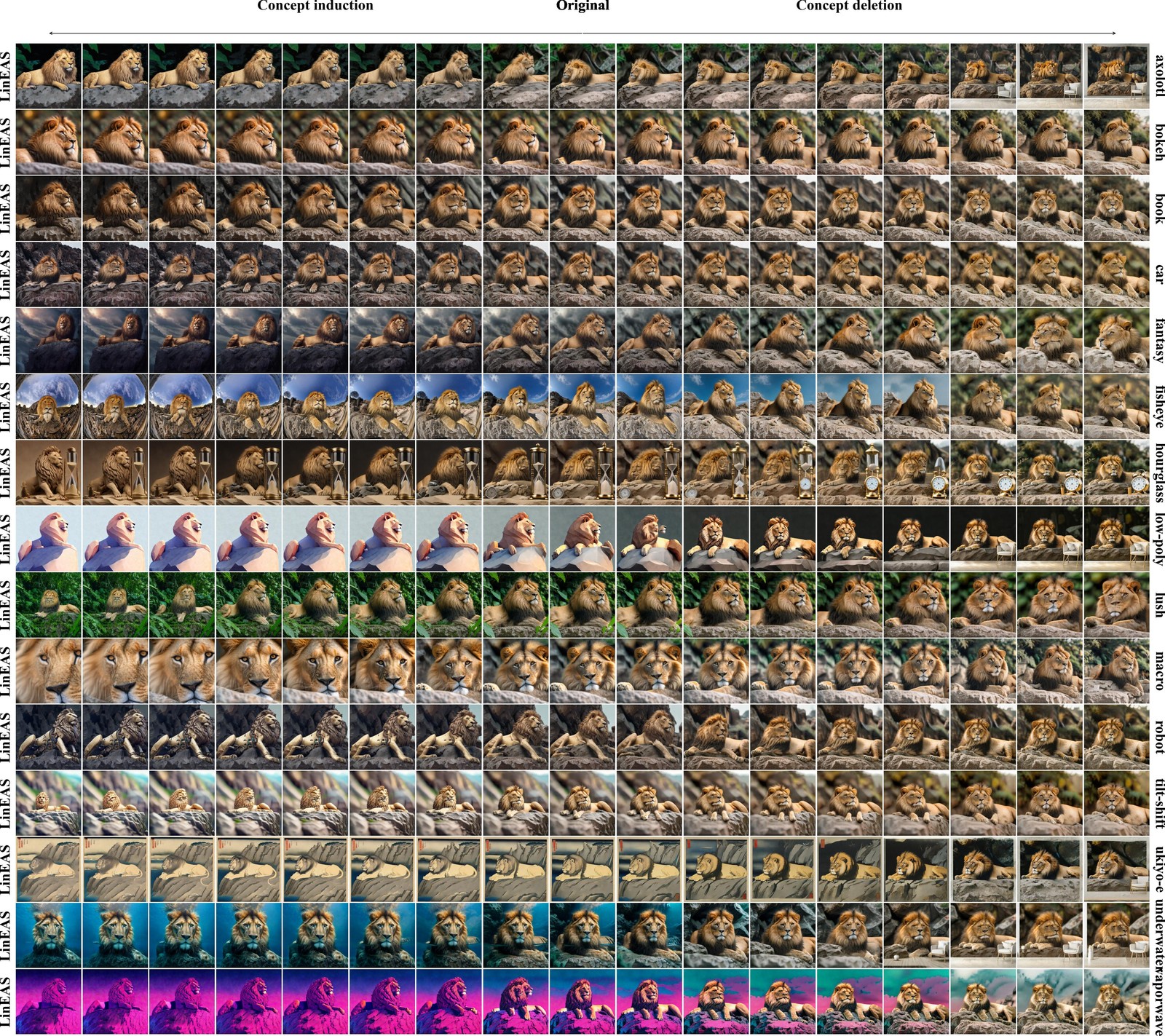}
    \caption{Images generated with \method with the prompt \textit{A majestic lion rests on a rocky outcrop. [concept tag]} Each column contains images generated with a different steering strength for mitigating the concept corresponding to the row: $\lambda=[1.0 ,..., 0, ..., 1.0]$. Images to the left of ``original'' were produced using the inverse steering map $\RT{\ell}^{-1}(z)$. }
    
    \label{fig:app-qualitative-6}
\end{figure}

We extend here the qualitative results mitigating and inducing styles on the 15 concepts described in~\cref{app:t2i-prompts}. \Cref{fig:app-qualitative-1,fig:app-qualitative-2,fig:app-qualitative-3} compare \linearact, and \method. Both the diversity and (human) perceptual quality of the generations is higher with \method. We observe stark differences between both methods' generations. \Cref{fig:app-qualitative-4,fig:app-qualitative-5,fig:app-qualitative-6} show additional generations with \method with more granular strengths.

Note that all the generations start from the same prompt, with the concept of interest appended in the form of textual tags. It is interesting to see that \method recovers an image conforms with the prompt \textit{without the concept} at $\lambda=1$. Also, observe how \method's generations are much more gradual than \linearact.

We also comment on the surprising results obtained when inverting the \method linear maps. We observe how the concept increases, and \method shows much higher quality and coherence under this regime. This points to the fact that \method is better exploiting the underlying structure in activation space.


\FloatBarrier
\clearpage
\subsection{Detailed Quantitative Results}
\label{app:t2i-quantitative-detail}
Here we report detailed per-concept ImgScores and ClipScores for \iti, \linearact, and \method both for the forward (($\RT{\ell}(z)$) and inverse (($\RT{\ell}^{-1}(z)$) application of the steering (\Cref{tab:t2i-app-detail}). Remarkably, we find that inverting the steering operation tends to induce the concepts that these methods mitigate when they are not inverted.

\begin{table}[h]
\centering
\resizebox{0.7\textwidth}{!}{%
\begin{tabular}{llll|ll}
\toprule
& & \multicolumn{2}{c|}{$\RT{\ell}(z)$} & \multicolumn{2}{c}{$\RT{\ell}^{-1}(z)$} \\
Task & Intervention & $\uparrow$ImgScore & $\uparrow$ClipScore & $\uparrow$ImgScore & $\downarrow$ClipScore.\\
\midrule
\multirow{3}{*}{Axolotl} & \iti & $0.13_{\color{gray}{\,0.11}}$ & $\textbf{0.17}_{\color{gray}{\,0.02}}$ & $0.26_{\color{gray}{\,0.17}}$ & $0.27_{\color{gray}{\,0.02}}$ \\
& \linearact & $0.35_{\color{gray}{\,0.19}}$ & $0.15_{\color{gray}{\,0.02}}$ & $0.28_{\color{gray}{\,0.18}}$ & $0.29_{\color{gray}{\,0.03}}$ \\
& \method & $\textbf{0.56}_{\color{gray}{\,0.23}}$ & $0.16_{\color{gray}{\,0.04}}$ & $\textbf{0.60}_{\color{gray}{\,0.19}}$ & $\textbf{0.23}_{\color{gray}{\,0.05}}$ \\
\midrule
\multirow{3}{*}{Bokeh} & \iti & $0.38_{\color{gray}{\,0.19}}$ & $\textbf{0.18}_{\color{gray}{\,0.02}}$ & $0.33_{\color{gray}{\,0.18}}$ & $\textbf{0.24}_{\color{gray}{\,0.02}}$ \\
& \linearact & $0.55_{\color{gray}{\,0.18}}$ & $\textbf{0.18}_{\color{gray}{\,0.02}}$ & $0.31_{\color{gray}{\,0.19}}$ & $\textbf{0.24}_{\color{gray}{\,0.02}}$ \\
& \method & $\textbf{0.75}_{\color{gray}{\,0.14}}$ & $\textbf{0.18}_{\color{gray}{\,0.02}}$ & $\textbf{0.62}_{\color{gray}{\,0.22}}$ & $\textbf{0.24}_{\color{gray}{\,0.02}}$ \\
\midrule
\multirow{3}{*}{Book} & \iti & $0.37_{\color{gray}{\,0.20}}$ & $\textbf{0.19}_{\color{gray}{\,0.01}}$ & $0.45_{\color{gray}{\,0.19}}$ & $0.21_{\color{gray}{\,0.02}}$ \\
& \linearact & $0.62_{\color{gray}{\,0.16}}$ & $0.18_{\color{gray}{\,0.01}}$ & $0.49_{\color{gray}{\,0.19}}$ & $0.22_{\color{gray}{\,0.01}}$ \\
& \method & $\textbf{0.80}_{\color{gray}{\,0.10}}$ & $0.17_{\color{gray}{\,0.01}}$ & $\textbf{0.78}_{\color{gray}{\,0.14}}$ & $\textbf{0.19}_{\color{gray}{\,0.02}}$ \\
\midrule
\multirow{3}{*}{Car} & \iti & $0.18_{\color{gray}{\,0.17}}$ & $\textbf{0.19}_{\color{gray}{\,0.02}}$ & $0.43_{\color{gray}{\,0.23}}$ & $0.24_{\color{gray}{\,0.02}}$ \\
& \linearact & $0.37_{\color{gray}{\,0.21}}$ & $0.18_{\color{gray}{\,0.02}}$ & $0.42_{\color{gray}{\,0.23}}$ & $0.23_{\color{gray}{\,0.02}}$ \\
& \method & $\textbf{0.78}_{\color{gray}{\,0.13}}$ & $0.18_{\color{gray}{\,0.03}}$ & $\textbf{0.81}_{\color{gray}{\,0.12}}$ & $\textbf{0.19}_{\color{gray}{\,0.03}}$ \\
\midrule
\multirow{3}{*}{Fantasy} & \iti & $0.12_{\color{gray}{\,0.14}}$ & $\textbf{0.20}_{\color{gray}{\,0.01}}$ & $0.35_{\color{gray}{\,0.19}}$ & $\textbf{0.21}_{\color{gray}{\,0.02}}$ \\
& \linearact & $0.49_{\color{gray}{\,0.19}}$ & $\textbf{0.20}_{\color{gray}{\,0.02}}$ & $0.52_{\color{gray}{\,0.17}}$ & $0.23_{\color{gray}{\,0.01}}$ \\
& \method & $\textbf{0.72}_{\color{gray}{\,0.16}}$ & $0.18_{\color{gray}{\,0.02}}$ & $\textbf{0.73}_{\color{gray}{\,0.15}}$ & $\textbf{0.21}_{\color{gray}{\,0.02}}$ \\
\midrule
\multirow{3}{*}{Fisheye} & \iti & $0.40_{\color{gray}{\,0.16}}$ & $\textbf{0.20}_{\color{gray}{\,0.02}}$ & $0.21_{\color{gray}{\,0.13}}$ & ${0.25}_{\color{gray}{\,0.01}}$ \\
& \linearact & $0.49_{\color{gray}{\,0.18}}$ & $\textbf{0.20}_{\color{gray}{\,0.02}}$ & $0.26_{\color{gray}{\,0.15}}$ & $0.25_{\color{gray}{\,0.02}}$ \\
& \method & $\textbf{0.68}_{\color{gray}{\,0.14}}$ & $\textbf{0.20}_{\color{gray}{\,0.02}}$ & $\textbf{0.51}_{\color{gray}{\,0.16}}$ & $\textbf{0.24}_{\color{gray}{\,0.02}}$ \\
\midrule
\multirow{3}{*}{Hourglass} & \iti & $0.16_{\color{gray}{\,0.18}}$ & $\textbf{0.20}_{\color{gray}{\,0.02}}$ & $0.57_{\color{gray}{\,0.26}}$ & $0.30_{\color{gray}{\,0.01}}$ \\
& \linearact & $0.29_{\color{gray}{\,0.23}}$ & $0.17_{\color{gray}{\,0.02}}$ & $0.60_{\color{gray}{\,0.27}}$ & $0.29_{\color{gray}{\,0.01}}$ \\
& \method & $\textbf{0.54}_{\color{gray}{\,0.23}}$ & $\textbf{0.20}_{\color{gray}{\,0.04}}$ & $\textbf{0.74}_{\color{gray}{\,0.19}}$ & $\textbf{0.27}_{\color{gray}{\,0.03}}$ \\
\midrule
\multirow{3}{*}{Low-poly} & \iti & $0.12_{\color{gray}{\,0.12}}$ & $\textbf{0.19}_{\color{gray}{\,0.02}}$ & $0.50_{\color{gray}{\,0.17}}$ & $\textbf{0.23}_{\color{gray}{\,0.02}}$ \\
& \linearact & $0.39_{\color{gray}{\,0.16}}$ & $\textbf{0.19}_{\color{gray}{\,0.02}}$ & $0.57_{\color{gray}{\,0.14}}$ & $0.24_{\color{gray}{\,0.02}}$ \\
& \method & $\textbf{0.57}_{\color{gray}{\,0.19}}$ & $0.18_{\color{gray}{\,0.01}}$ & $\textbf{0.67}_{\color{gray}{\,0.14}}$ & $0.25_{\color{gray}{\,0.02}}$ \\
\midrule
\multirow{3}{*}{Lush} & \iti & $0.29_{\color{gray}{\,0.21}}$ & $0.20_{\color{gray}{\,0.02}}$ & $0.44_{\color{gray}{\,0.17}}$ & $0.26_{\color{gray}{\,0.01}}$ \\
& \linearact & $0.50_{\color{gray}{\,0.20}}$ & $\textbf{0.21}_{\color{gray}{\,0.02}}$ & $0.47_{\color{gray}{\,0.17}}$ & $0.25_{\color{gray}{\,0.01}}$ \\
& \method & $\textbf{0.71}_{\color{gray}{\,0.15}}$ & $0.18_{\color{gray}{\,0.02}}$ & $\textbf{0.75}_{\color{gray}{\,0.12}}$ & $\textbf{0.23}_{\color{gray}{\,0.02}}$ \\
\midrule
\multirow{3}{*}{Macro} & \iti & $0.31_{\color{gray}{\,0.16}}$ & $\textbf{0.20}_{\color{gray}{\,0.01}}$ & $0.18_{\color{gray}{\,0.15}}$ & $\textbf{0.22}_{\color{gray}{\,0.02}}$ \\
& \linearact & $0.52_{\color{gray}{\,0.18}}$ & $0.18_{\color{gray}{\,0.02}}$ & $0.16_{\color{gray}{\,0.13}}$ & $\textbf{0.22}_{\color{gray}{\,0.02}}$ \\
& \method & $\textbf{0.68}_{\color{gray}{\,0.17}}$ & $0.18_{\color{gray}{\,0.02}}$ & $\textbf{0.54}_{\color{gray}{\,0.21}}$ & $0.23_{\color{gray}{\,0.02}}$ \\
\midrule
\multirow{3}{*}{Robot} & \iti & $0.36_{\color{gray}{\,0.20}}$ & $\textbf{0.20}_{\color{gray}{\,0.02}}$ & $0.32_{\color{gray}{\,0.17}}$ & $0.23_{\color{gray}{\,0.02}}$ \\
& \linearact & $0.56_{\color{gray}{\,0.17}}$ & $0.18_{\color{gray}{\,0.02}}$ & $0.36_{\color{gray}{\,0.19}}$ & $0.25_{\color{gray}{\,0.02}}$ \\
& \method & $\textbf{0.74}_{\color{gray}{\,0.15}}$ & $0.18_{\color{gray}{\,0.02}}$ & $\textbf{0.71}_{\color{gray}{\,0.15}}$ & $\textbf{0.21}_{\color{gray}{\,0.02}}$ \\
\midrule
\multirow{3}{*}{Tilt-shift} & \iti & $0.11_{\color{gray}{\,0.11}}$ & $0.19_{\color{gray}{\,0.02}}$ & $0.44_{\color{gray}{\,0.20}}$ & $0.29_{\color{gray}{\,0.01}}$ \\
& \linearact & $0.46_{\color{gray}{\,0.20}}$ & $0.21_{\color{gray}{\,0.02}}$ & $0.48_{\color{gray}{\,0.21}}$ & $0.29_{\color{gray}{\,0.01}}$ \\
& \method & $\textbf{0.65}_{\color{gray}{\,0.16}}$ & $\textbf{0.23}_{\color{gray}{\,0.03}}$ & $\textbf{0.65}_{\color{gray}{\,0.15}}$ & $\textbf{0.28}_{\color{gray}{\,0.02}}$ \\
\midrule
\multirow{3}{*}{Ukiyo-e} & \iti & $0.20_{\color{gray}{\,0.12}}$ & $\textbf{0.21}_{\color{gray}{\,0.02}}$ & $0.47_{\color{gray}{\,0.16}}$ & $\textbf{0.28}_{\color{gray}{\,0.02}}$ \\
& \linearact & $0.29_{\color{gray}{\,0.17}}$ & $0.20_{\color{gray}{\,0.02}}$ & $0.64_{\color{gray}{\,0.15}}$ & $0.29_{\color{gray}{\,0.01}}$ \\
& \method & $\textbf{0.48}_{\color{gray}{\,0.18}}$ & $0.18_{\color{gray}{\,0.03}}$ & $\textbf{0.73}_{\color{gray}{\,0.12}}$ & $\textbf{0.28}_{\color{gray}{\,0.01}}$ \\
\midrule
\multirow{3}{*}{Underwater} & \iti & $0.33_{\color{gray}{\,0.19}}$ & $\textbf{0.21}_{\color{gray}{\,0.02}}$ & $0.43_{\color{gray}{\,0.19}}$ & $\textbf{0.26}_{\color{gray}{\,0.01}}$ \\
& \linearact & $0.52_{\color{gray}{\,0.21}}$ & $0.18_{\color{gray}{\,0.02}}$ & $0.41_{\color{gray}{\,0.21}}$ & $\textbf{0.26}_{\color{gray}{\,0.02}}$ \\
& \method & $\textbf{0.59}_{\color{gray}{\,0.21}}$ & $0.17_{\color{gray}{\,0.02}}$ & $\textbf{0.56}_{\color{gray}{\,0.20}}$ & $\textbf{0.26}_{\color{gray}{\,0.01}}$ \\
\midrule
\multirow{3}{*}{Vaporwave} & \iti & $0.12_{\color{gray}{\,0.17}}$ & $\textbf{0.18}_{\color{gray}{\,0.01}}$ & $0.35_{\color{gray}{\,0.17}}$ & $\textbf{0.26}_{\color{gray}{\,0.01}}$ \\
& \linearact & $0.33_{\color{gray}{\,0.18}}$ & $0.15_{\color{gray}{\,0.02}}$ & $0.56_{\color{gray}{\,0.14}}$ & $0.27_{\color{gray}{\,0.01}}$ \\
& \method & $\textbf{0.65}_{\color{gray}{\,0.17}}$ & $0.16_{\color{gray}{\,0.02}}$ & $\textbf{0.66}_{\color{gray}{\,0.16}}$ & $\textbf{0.26}_{\color{gray}{\,0.02}}$ \\
\bottomrule
\end{tabular}}
\vskip 1mm
\caption{ImgScore and ClipScore on mitigation for all concepts. Columns 3 and 4 contain results with the original steering direction ($\RT{\ell}(z)$), \ie removing or mitigating the concepts. Columns 5 and 6 contain results applying the inverse steering operation ($\RT{\ell}^{-1}(z)$), which effectively \textit{induces} the concepts.  \method consistently achieves a higher ImgScore for similar ClipScore values.}
\label{tab:t2i-app-detail}
\end{table}

\FloatBarrier
\clearpage
\section{Image Prompts Dataset}
\label{app:t2i-prompts}
In \Cref{sec:results_diffusion}, we evaluate multiple activation steering methods on DMD2~\citep{yin2024dmd2}. To probe different aspects of T2I generation, we query an open-source LLM to generate a new dataset of prompts covering 3 different conditioning categories and 5 concepts per category. We also include a neutral category, which is used as the source in concept addition and target for removal. In \Cref{tab:t2i-dataset}, we show a sample of the dataset, and we include the full dataset in the supplementary material.

\begin{table}[h!]
\centering
\resizebox{\linewidth}{!}{\begin{tabular}{llp{14cm}}
\toprule
\textbf{Supercategory} & \textbf{Concept} & \textbf{Example Prompts} \\
\midrule
\multirow{4}{*}{Neutral} &  & A beaver in its natural habitat. \\
& & A playful dolphin jumping out of water. \\
& & A train passing through mountains.\\
& & A sturdy table with drawers.\\
\midrule
\multirow{5}{*}{Style} & Vaporwave & Vaporwave neon cityscape at dusk \\
 & & Retro-futuristic arcade machine in a vaporwave setting \\
\cmidrule{2-3}
 & Lush & In a lush, overgrown jungle, two young men sitting on a bench and a lady standing next to them. \\
 & & Seagulls in flight with a person feeding one, a lighthouse in the distance, surrounded by a lush, overgrown jungle. \\
\cmidrule{2-3}
 & Low-poly & A mountain range at sunrise rendered in low poly style with crisp, angular facets. \\
 & & A futuristic cityscape with neon accents and low poly geometry that creates a digital vibe. \\
\cmidrule{2-3}
 & Ukiyo-e & A majestic view of Mount Fuji, cherry blossoms in full bloom, woodblock print style, Ukiyo-e, Edo period aesthetics \\
 & & A stormy sea with giant waves crashing, a lone boat struggling against the current, traditional Japanese woodblock print, Ukiyo-e \\
\cmidrule{2-3}
 & Fantasy & A majestic dragon flying over a glowing crystal mountain range, under a purple sky, fantasy art \\
 & & A knight in shining armor standing before a towering, ancient forest, mist swirling around, high fantasy \\
\midrule
\multirow{5}{*}{Objects} & Robot & A sleek silver robot waves hello in a futuristic city. \\
 & & A tiny robot with glowing eyes explores a dark cave. \\
\cmidrule{2-3}
 & Axolotl & A pale pink leucistic axolotl with feathery external gills, smiling serenely in a clear aquarium. \\
 & & A wild-type axolotl, dark and speckled, camouflaged amongst aquatic plants in its natural habitat. \\
\cmidrule{2-3}
 & Book & A leather-bound antique book, its gold-leaf title faded, resting on a dusty mahogany desk. \\
 & & A stack of colorful children's picture books, vibrant illustrations peeking from the edges. \\
\cmidrule{2-3}
 & Car & A gleaming cherry-red classic 1950s convertible, chrome shining, cruising down a sun-drenched coastal highway, ocean on one side. \\
 & & A rugged, mud-splattered off-road 4x4 vehicle navigating a steep, rocky mountain trail, dust kicking up from its tires. \\
\cmidrule{2-3}
 & Hourglass & An antique brass hourglass, its fine golden sand steadily flowing from the top bulb to the bottom, against a dark, moody background. \\
 & & A minimalist, modern glass hourglass with vibrant blue sand, casting a sharp shadow on a white surface. \\
\midrule
\multirow{5}{*}{Perspectives} & Macro & Macro shot of a ladybug on a vibrant green leaf, its tiny black spots in sharp focus, dewdrop clinging nearby. \\
 & & Extreme macro shot of a honeybee's multifaceted eye, revealing intricate hexagonal patterns. \\
\cmidrule{2-3}
 & Fisheye & Fisheye lens perspective of a bustling street market, vibrant stalls and crowds curving dramatically around a central point. \\
 & & Extreme fisheye shot from the center of a packed concert crowd, hands raised, stage lights creating circular flares. \\
\cmidrule{2-3}
 & Bokeh & Portrait of a smiling woman, her face in sharp focus, against a background of beautifully blurred city lights creating circular bokeh. \\
 & & A single red rose in perfect focus, its delicate petals detailed, with a creamy green bokeh background of garden foliage. \\
\cmidrule{2-3}
 & Underwater & Underwater perspective of a vibrant coral reef teeming with colorful tropical fish, sunlight filtering through clear turquoise water. \\
 & & Sunken pirate shipwreck resting on the sandy ocean floor, schools of fish swimming through its decaying hull, underwater view. \\
\cmidrule{2-3}
 & Tilt-shift & Tilt-shift perspective of a bustling city intersection, cars and pedestrians appearing like tiny toys, vibrant colors. \\
 & & A miniature-effect tilt-shift shot of a freight train winding through a verdant, rolling landscape. \\
\bottomrule
\end{tabular}}
\caption{Sample of the dataset used for conditioning T2I models. The dataset is divided in 4 different categories: (1) neutral prompts used as source for concept addition and target for removal, (2) style prompts, (3) objects, and (4) perspectives.}
\label{tab:t2i-dataset}
\end{table}

\applefootnote{ \textcolor{textgray}{\sffamily Apple and the Apple logo are trademarks of Apple Inc., registered in the U.S. and other countries and regions.}}

\end{document}